\PassOptionsToPackage{dvipsnames}{xcolor}

\documentclass[letterpaper]{article}

\usepackage[T1]{fontenc}

\usepackage{geometry}
\usepackage{setspace}
\usepackage[sort&compress,numbers]{natbib}

\usepackage{graphicx}
\usepackage{float}

\usepackage{booktabs}
\usepackage{threeparttable}
\usepackage{xspace}
\usepackage{longtable}
\usepackage[normalem]{ulem}   %
\usepackage{amsmath}
\usepackage{amssymb}
\usepackage{xcolor}
\usepackage{hyperref}
\usepackage{subcaption}
\usepackage{multirow}
\usepackage{adjustbox}
\usepackage{pdflscape}

\usepackage{tikz}
\usetikzlibrary{
  arrows.meta,
  positioning,
  fit,
  backgrounds,
  shapes.geometric,
  calc,
  decorations.pathreplacing
}

\graphicspath{{./}{images/}}

\newcommand{\rev}[1]{#1}
\newcommand{\old}[1]{}
\newenvironment{revblock}{}{}
\newcommand{\revfloat}[1]{#1}          %

\newcommand{\ModelName}{STAR-VAE\xspace}

\DeclareRobustCommand{\circled}[1]{%
  \mbox{%
    \tikz[baseline=(c.base)]{%
      \node[circle, draw=black!70, fill=white, inner sep=0.8pt,
            font=\tiny\bfseries, text=black!80,
            minimum size=0.38cm] (c) {\vphantom{0}#1};}}}

\usepackage{authblk}

\author[1,*]{Bum Chul Kwon}
\author[3]{Ben Shapira}
\author[3]{Moshiko Raboh}
\author[2]{Shreyans Sethi}
\author[2]{Shruti Murarka}
\author[2]{Joseph A. Morrone}
\author[2]{Leili Zhang}
\author[2]{Wendy Cornell}
\author[2]{Pablo Meyer}
\author[2]{Jianying Hu}
\author[2,*]{Parthasarathy Suryanarayanan}

\affil[1]{IBM Research, Cambridge, MA 02142, USA}
\affil[2]{IBM Research, Yorktown Heights, NY 10598, USA}
\affil[3]{IBM Research, Haifa, Israel}

\title{\ModelName: A Scalable Latent-Variable Transformer for Controllable
  Molecular Generation}

\date{%
  *Corresponding Authors: \\
  Bum Chul Kwon \url{bumchul.kwon@us.ibm.com} \\
  Parthasarathy Suryanarayanan \url{psuryan@us.ibm.com}%
}

\begin{document}

\maketitle

\begin{abstract}
\begin{sloppypar}
\rev{Many molecular Transformers lack probabilistic latent variables for posterior inference and latent interpolation. We introduce \ModelName, a \textbf{S}ELFIES-encoded, \textbf{T}ransformer-based, \textbf{A}uto\textbf{R}egressive \textbf{V}ariational \textbf{A}uto\textbf{E}ncoder combining a bidirectional encoder with an autoregressive decoder pretrained on 79 million PubChem molecules. A property signal jointly conditions the prior, posterior, and decoder, while LoRA adapters support fine-tuning on small datasets without modifying the backbone. \ModelName\ achieves 100\% validity and near-perfect novelty under unconditional MOSES sampling, the lowest KL divergence on five of ten GuacaMol descriptors, Spearman $\rho\!=\!0.62$ at $98\%$ validity for synthetic-accessibility conditioning, and directional docking-score control for three Tartarus protein targets. Across four ChEMBL targets, seed-based posterior sampling recovers target-associated held-out scaffolds while label-conditioned sampling produces structurally diverse outputs. Code is available at \url{https://github.com/BiomedSciAI/STAR-VAE}.}
\end{sloppypar}
 \end{abstract}

\section*{Keywords}
\ModelName; variational autoencoder; molecular generation; SELFIES; Transformer;
conditional generation; LoRA; latent variable model; drug discovery

\section{Introduction}
\label{sec:introduction}
The space of synthesizable small molecules is estimated to exceed $10^{33}$ compounds~\cite{polishchuk2013estimation}, far beyond what can be exhaustively enumerated or tested experimentally. Generative modeling offers a principled route through this space, enabling data-driven exploration focused on molecules with desirable properties~\cite{gao2020synthesizability, di2009drug}. Deep generative models have consequently become central to de novo drug design, with a wide range of architectures now proposed for molecular generation~\cite{cheng2021review, tang2024survey}. Useful molecular generative models must learn broad chemical distributions, support label-guided generation, and remain computationally tractable.

Variational autoencoders (VAEs) were among the first deep generative models applied to molecules and remain influential for their ability to produce smooth latent spaces amenable to interpolation and optimization. Gómez-Bombarelli et al.~\cite{gomez2018automatic} showed that a convolutional encoder and gated recurrent unit (GRU) decoder operating on SMILES strings could navigate chemical space via gradient-based latent optimization. Graph-based approaches~\cite{simonovsky2018graphvae,ma2021gf,jin2018junction} moved to direct graph generation, and conditional VAEs~\cite{lim2018molecular,kang2018conditional,yang2023cmgn} extended the paradigm to label-guided synthesis. Despite this breadth, early molecular VAEs relied on recurrent or graph-based architectures and were trained on relatively small datasets, leaving open the question of whether the VAE paradigm can scale to modern data regimes.

Subsequent work shifted toward large Transformer architectures. Models such as MolGPT~\cite{bagal2021molgpt} and MolGen~\cite{fang2023domain} generate SMILES~\cite{weininger1989smiles} or SELFIES one token at a time. They achieve high validity and diversity but do not use an explicit variational latent space for posterior inference or latent interpolation. Chemformer~\cite{irwin2022chemformer}, SELF-BART~\cite{priyadarsini2024self}, and SELFIES-TED~\cite{priyadarsini2025selfiested} use encoder--decoder architectures with denoising pretraining. They support several molecular prediction and generation tasks, but they also lack an explicit variational latent distribution. Adversarial, flow, and diffusion methods~\cite{li2023spotgan,li2024tengan,kuznetsov2021molgrow,jo2022score,huang2023conditional,lee2023exploring,park2026diffusion,norouzi2026mt} provide other approaches to molecular generation. Target-aware graph models~\cite{kang2026pocket} address goal-oriented generation~\cite{du2024machine}. Large training sets and large decoders have improved validity and diversity, while limited labeled data remain a challenge for conditional generation.

\old{We revisit the latent-variable paradigm and show that a modernized VAE, scaled appropriately and paired with principled conditioning, is competitive with recent large-scale generators.} \rev{We revisit the latent-variable paradigm and ask whether a modernized VAE, scaled to 79 million pretraining molecules and paired with principled conditioning, can reproduce properties of the MOSES and GuacaMol benchmark distributions and support controlled generation.} \old{We present \ModelName\ (\textbf{S}\textsc{elfies-encoded}, \textbf{T}\textsc{ransformer-based}, \textbf{A}\textsc{uto}\textbf{R}\textsc{egressive} \textbf{V}ariational \textbf{A}uto\textbf{E}ncoder), a conditional VAE combining a bidirectional Transformer encoder with an autoregressive Transformer decoder, pretrained on 79 million drug-like molecules from PubChem in SELFIES~\cite{krenn_self_referencing_2020}.}\rev{We present \ModelName\ (\textbf{S}\textsc{elfies-encoded}, \textbf{T}\textsc{ransformer-based}, \textbf{A}\textsc{uto}\textbf{R}\textsc{egressive} \textbf{V}ariational \textbf{A}uto\textbf{E}ncoder), which integrates a bidirectional Transformer encoder and an autoregressive Transformer decoder with a variational latent space. The model is pretrained on 79 million drug-like PubChem molecules represented as SELFIES~\cite{krenn_self_referencing_2020}. This design supports posterior inference, latent-space sampling, and property-conditioned generation in one model.} A single scalar label conditions the latent prior, inference network, and decoder jointly; the current implementation conditions on one property at a time. Label sources differ by experiment: SA score and docking score are computed directly with RDKit and QVina respectively, while the rediscovery experiments use MMELON-derived pseudo-labels. LoRA modules in both encoder and decoder enable label-conditioned fine-tuning on small datasets with modest compute.

\old{Our contributions are: (i) a Transformer-based latent-variable encoder-decoder pretrained at scale on SELFIES, showing that the VAE paradigm scales to modern data regimes; (ii) a conditional latent-variable formulation in which a single property signal shapes the prior, posterior, and decoder \emph{jointly}; and (iii) conditional fine-tuning via LoRA in both encoder and decoder, enabling adaptation with limited labeled data. On the GuacaMol~\cite{brown2019guacamol} and MOSES~\cite{polykovskiy2020molecular} unconditional benchmarks, \ModelName\ achieves strong scores on validity, uniqueness, and novelty, and matches or exceeds the four retrained GuacaMol baselines on five of ten physicochemical descriptors. On the Tartarus protein-ligand design tasks~\cite{nigam_2023_tartarus}, the conditional model produces a positive alignment between requested target and realized docking score for all three targets, generating many high-scoring diverse candidates rather than a single optimized outlier. Rediscovery experiments across four ChEMBL~\cite{mendez2019chembl} target datasets show that the pretrained VAE under seed-based sampling recovers more target-associated held-out scaffolds, while label-only CVAE sampling under a broad label distribution produces more structurally varied outputs. These are different operating regimes rather than a single uniformly better mode.}
\rev{This work contributes a Transformer-based variational model pretrained at scale on SELFIES and a conditional formulation in which one property signal shapes the prior, approximate posterior, and decoder. LoRA supports adaptation to smaller labeled datasets. On MOSES~\cite{polykovskiy2020molecular} and GuacaMol~\cite{brown2019guacamol}, \ModelName\ shows high validity and novelty and obtains the lowest KL divergence for five of ten molecular properties among the four retrained GuacaMol baselines. Quantitative conditioning experiments show control of synthetic accessibility and docking scores for all three Tartarus protein targets~\cite{nigam_2023_tartarus}. The SA-score comparison with REINVENT~4, MolGPT, and DrugDiff characterizes trade-offs among different conditioning approaches. Compared with MolGPT under the shared continuous-conditioning protocol, \ModelName\ produces a higher fraction of valid molecules and more unique scaffolds, while MolGPT tracks requested SA scores more closely among its valid outputs. A five-factor ablation study with three independent fine-tuning runs per selected configuration shows that measurable control requires a conditioning path to the decoder and that differences among conditioning-capable variants fall within run-to-run variation. Finally, experiments on four ChEMBL targets~\cite{mendez2019chembl} show complementary generation modes: seed-based VAE sampling recovers more target-associated held-out scaffolds, while label-conditioned CVAE sampling produces more varied structures.}
 
\section{Methods}
\label{sec:methods}

\subsection{Data}
\label{sec:data}
The objective of the pretraining dataset is to learn a task-agnostic molecular representation through large-scale self-supervision.
We use PubChem~\cite{kim_pubchem_2022}, a public chemical database curated by the National Center for Biotechnology Information (NCBI) that aggregates data on small molecules from over 870 contributing sources, including substances, bioassays, targets, and patents. From this resource, we curate a drug-like subset of approximately 79 million molecules. The curation process involves extracting the largest molecular fragment, removing duplicates, and applying standard drug-likeness filters: MW $\leq$ 600~Da, $\leq$5 H-bond donors, $\leq$10 acceptors, $\leq$10 rotatable bonds. 
These filters yield a dataset covering broad chemical diversity while remaining within drug-relevant physicochemical ranges.

\subsection{Molecular Representation}
\label{sec:representation}

We represent molecules using the SELFIES (Self-Referencing Embedded Strings) representation~\cite{krenn_self_referencing_2020}. SELFIES substantially reduces syntax-related generation failures because valid SELFIES strings decode to molecular graphs; downstream conversion and RDKit sanitization can nevertheless reject some generated structures, so we report empirical RDKit validity as the operational metric throughout. 
Compared to SMILES, SELFIES removes most syntactic failure modes and is therefore well-suited for sequence-based generative modeling at scale.

\paragraph{Tokenization.}
Each molecule is converted into a SELFIES string and then split into discrete tokens drawn from a fixed vocabulary. 
Padding, start-of-sequence, end-of-sequence, and unknown tokens are added for sequence handling during training and generation.
The resulting vocabulary covers atoms, branches, rings, and stereochemical markers encountered in the dataset. For efficient batching and consistent positional encoding, we limit sequence length to a maximum of 71 tokens, which covers over 99\% of drug-like molecules in our training dataset. Shorter sequences are padded, while longer ones are truncated.

\paragraph{Positional encoding.}
Absolute positional encodings are used throughout; the fixed 71-token maximum makes rotary or relative schemes unnecessary.

\subsection{Model Architecture}
\label{sec:architecture}

\begin{figure}[H]
  \centering
  \begin{adjustbox}{max width=\linewidth,
                    max totalheight=0.35\textheight,trim={2.5cm 0 0 0},
                    keepaspectratio}

\colorlet{cPre}  {blue!65!black}
\colorlet{cFine} {green!55!black}
\colorlet{cGenV} {violet!75!black}
\colorlet{cGenC} {orange!70!black}
\colorlet{cLat}  {teal!60!black}
\colorlet{cFilm} {purple!60!black}
\colorlet{cLoRA} {red!60!black}
\colorlet{cCond} {orange!50!black}
\colorlet{cBg}   {gray!6}

\pgfdeclarelayer{behind}
\pgfdeclarelayer{middle}
\pgfdeclarelayer{aboveB}
\pgfsetlayers{behind,background,middle,aboveB,main}

\tikzset{
  blk/.style={
    draw=#1, rounded corners=2pt, fill=cBg,
    minimum width=3.2cm, minimum height=0.50cm,
    align=center, font=\scriptsize, inner sep=2pt
  },
  blk/.default=black,
  tfblk/.style={
    draw=#1, rounded corners=2pt, fill=#1!10,
    minimum width=3.2cm, minimum height=0.70cm,
    align=center, font=\scriptsize\bfseries, inner sep=2pt
  },
  tfblk/.default=black,
  lorabadge/.style={
    draw=cLoRA, rounded corners=2pt,
    fill=cLoRA!12, font=\tiny\bfseries, text=cLoRA,
    minimum width=0.80cm, minimum height=0.28cm, align=center
  },
  cbadge/.style={
    circle, draw=black!70, fill=white, font=\tiny\bfseries,
    minimum size=0.38cm, inner sep=0pt, text=black!80
  },
  sec/.style={font=\footnotesize\bfseries, align=center},
  arr/.style={-Latex, semithick, #1},
  arr/.default=black,
  darr/.style={-Latex, semithick, dashed, #1},
  darr/.default=black,
  stagebox/.style={draw=#1, rounded corners=4pt, fill=#1!4, inner sep=5pt},
  stagebox/.default=black,
}

\newlength{\HB}  \setlength{\HB}{11.0cm}
\newlength{\HCD} \setlength{\HCD}{6.5cm}  %
\newlength{\GR}  \setlength{\GR}{0.6cm}
\newlength{\HE}  \setlength{\HE}{2.8cm}

\def\xA{1.78cm}
\def\xB{9.08cm}
\def\xC{16.94cm}
\def\xD{23.30cm}

\def\gA{0.544cm}   %
\def\gB{0.354cm}   %

\begin{tikzpicture}

\coordinate (y0) at (0,0);

\node[sec, cPre] at (\xA, 0 |- y0) (hA)
  {\textbf{(A) Pre-training: VAE}};

\node[inner sep=0pt, minimum size=0pt]
  at ([yshift=-\HB]hA.north) (pinA){};

\node[blk=cPre,   below=\gA of hA] (inA)
  {Input SELFIES};
\node[tfblk=cPre, below=\gA of inA] (encA)
  {Transformer Encoder\\[-1pt]\tiny$12{\times}$ layers,\ $d{=}512$};
\node[blk=cLat,   below=\gA of encA, fill=cLat!10] (muA)
  {$\boldsymbol{\mu},\log\boldsymbol{\sigma}^{2}$\ (FC)};
\node[blk=cLat,   below=\gA of muA, fill=cLat!22,
      font=\scriptsize\bfseries] (zA)
  {$\mathbf{z}{\sim}q(\mathbf{z}|x),\ \mathbb{R}^{256}$};
\node[blk=cLat,   below=\gA of zA, fill=cLat!10] (l2hA)
  {Latent-to-hidden (FC)};
\node[tfblk=cPre, below=\gA of l2hA] (decA)
  {Transformer Decoder\\[-1pt]\tiny$12{\times}$ layers,\ $d{=}512$};
\node[blk=cPre,   below=\gA of decA] (outA)
  {Recon.\ SELFIES};
\node[blk=cPre!55,below=\gA of outA, fill=cPre!5,
      font=\tiny, minimum width=3.4cm] (lossA)
  {$\mathcal{L}{=}\mathcal{L}_\mathrm{recon}
   {+}\beta\,\mathrm{KL}[q\|\mathcal{N}(0,I)]$};

\begin{pgfonlayer}{background}
  \node[stagebox=cPre,
        fit=(hA)(inA)(encA)(muA)(zA)(l2hA)(decA)(outA)(lossA)(pinA)]
       (boxA){};
\end{pgfonlayer}

\draw[arr=cPre](inA)--(encA);
\draw[arr=cLat](encA)--(muA);
\draw[arr=cLat](muA)--(zA);
\draw[arr=cLat](zA)--(l2hA);
\draw[arr=cPre](l2hA)--(decA);
\draw[arr=cPre](decA)--(outA);
\draw[arr=cPre](outA)--(lossA);

\node[sec, cFine] at (\xB, 0 |- y0) (hB)
  {\textbf{(B) Fine-tuning: CVAE (VAE + LoRA + FiLM)}};

\node[inner sep=0pt, minimum size=0pt]
  at ([yshift=-\HB]hB.north) (pinB){};

\node[blk=cFine,  below=\gB of hB] (inB)   {Input SELFIES};
\node[tfblk=cFine,below=\gB of inB] (encB)
  {Transformer Encoder\\[-1pt]\tiny$12{\times}$ layers};
\node[blk=cFine,  below=\gB of encB, fill=cFine!8,
      minimum width=3.2cm] (lproj)
  {Mean pool $+$ label proj.\ ($y{\to}d$)};
\node[blk=cLat,   below=\gB of lproj, fill=cLat!10] (muB)
  {$\boldsymbol{\mu},\log\boldsymbol{\sigma}^{2}$\ (FC)};
\node[blk=cLat,   below=\gB of muB, fill=cLat!22,
      font=\scriptsize\bfseries] (zB)
  {$\mathbf{z}{\sim}q(\mathbf{z}|x,y),\ \mathbb{R}^{256}$};
\node[blk=cFilm,  below=\gB of zB, fill=cFilm!12,
      font=\scriptsize] (film)
  {\textbf{FiLM:}\ $\tilde{\mathbf{z}}{=}\gamma(y)\!\odot\!\mathbf{z}{+}\beta(y)$};
\node[blk=cLat,   below=\gB of film, fill=cLat!10] (l2hB)
  {Latent-to-hidden (FC)};
\node[tfblk=cFine,below=\gB of l2hB] (decB)
  {Transformer Decoder\\[-1pt]\tiny$12{\times}$ layers};
\node[blk=cFine,  below=\gB of decB] (outB)  {Recon.\ SELFIES};
\node[blk=cFine!55,below=\gB of outB, fill=cFine!5,
      font=\tiny, minimum width=3.4cm] (lossB)
  {$\mathcal{L}{=}\mathcal{L}_\mathrm{recon}
   {+}\mathrm{KL}[q(\mathbf{z}|x,y)\|p(\mathbf{z}|y)]$};

\node[lorabadge, left=0.14cm of encB] (lbenc) {LoRA};
\node[lorabadge, left=0.14cm of decB] (lbdec) {LoRA};

\node[blk=cCond, right=0.60cm of inB, fill=cCond!10,
      minimum width=1.2cm, minimum height=0.50cm,
      font=\scriptsize] (lblIn) {Score $y$};

\node[blk=cFilm, right=0.60cm of muB, fill=cFilm!8,
      minimum width=1.8cm, font=\tiny, align=center] (cprior)
  {Prior $p(\mathbf{z}|y)$\\[-1pt]$y{\to}\mu_p,\sigma_p^2$};

\begin{pgfonlayer}{background}
  \node[stagebox=cFine,
        fit=(hB)(inB)(lblIn)(lproj)(encB)(lbenc)(lbdec)
            (muB)(cprior)(zB)(film)(l2hB)(decB)(outB)(lossB)(pinB)]
       (boxB){};
\end{pgfonlayer}

\draw[arr=cFine](inB)--(encB);
\draw[arr=cFine](encB)--(lproj);
\draw[arr=cLat] (lproj)--(muB);
\draw[arr=cLat] (muB)--(zB);
\draw[arr=cFilm](zB)--(film);
\draw[arr=cLat] (film)--(l2hB);
\draw[arr=cFine](l2hB)--(decB);
\draw[arr=cFine](decB)--(outB);
\draw[arr=cFine](outB)--(lossB);

\begin{pgfonlayer}{aboveB}
\draw[darr=cCond]
  (lblIn.south) -- (lblIn.south |- lproj.east) -- (lproj.east);
\draw[darr=cCond]
  (lblIn.south) -- (lblIn.south |- film.east) -- (film.east);
\draw[darr=cCond]
  (lblIn.center) -- (cprior.center |- lblIn.center) -- (cprior.north);
\draw[darr=cFilm]
  (cprior.south) -- (cprior.south |- lossB.east)
  -- node[above, font=\tiny, cFilm, pos=0.7]{KL} (lossB.east);
\end{pgfonlayer}

\def\xCL{15.87cm}    %
\def\xCR{18.40cm}    %
\def\xCmid{17.135cm} %
\def\nwC{2.50cm}     %
\def\nwCz{1.90cm}    %

\node[sec, cGenV] at (\xCmid, 0 |- y0) (hC)
  {\textbf{(C) VAE Generation}};

\node[inner sep=0pt, minimum size=0pt]
  at ([yshift=-\HCD]hC.north) (pinC){};

\coordinate (seedCctr) at ([yshift=-1.00cm]hC.north);
\node[blk=cGenV, minimum width=\nwC, fill=cGenV!8, font=\scriptsize]
  at (\xCL, 0 |- seedCctr) (seedC)
  {Seed molecule};

\node[font=\tiny, text=cGenV!70, above=0.05cm of seedC, align=center] (lblSeedC)
  {\textit{seed-based generation}};

\coordinate (encCctr) at ([yshift=-2.30cm]hC.north);
\node[tfblk=cGenV, minimum width=\nwC, fill=cGenV!10]
  at (\xCL, 0 |- encCctr) (encC)
  {Pre-trained Encoder\\[-1pt]\tiny(frozen)};

\coordinate (zCseedctr) at ([yshift=-3.60cm]hC.north);
\node[blk=cLat, minimum width=\nwCz, fill=cLat!22, font=\scriptsize\bfseries]
  at (\xCL, 0 |- zCseedctr) (zCseed)
  {$\mathbf{z}{\sim}q(\mathbf{z}|x)$};

\coordinate (decCctr) at ([yshift=-4.90cm]hC.north);
\node[tfblk=cGenV, minimum width=\nwC, fill=cGenV!10]
  at (\xCmid, 0 |- decCctr) (decC)
  {Pre-trained Decoder\\[-1pt]\tiny(frozen, autoregressive)};

\coordinate (outCctr) at ([yshift=-6.20cm]hC.north);
\node[blk=cGenV, minimum width=\nwC]
  at (\xCmid, 0 |- outCctr) (outC)
  {Generated molecules};

\node[blk=cLat, minimum width=\nwCz, fill=cLat!22, font=\scriptsize\bfseries]
  at (\xCR, 0 |- zCseed.center) (zCrand)
  {$\mathbf{z}{\sim}\mathcal{N}(\mathbf{0},\mathbf{I})$};
\node[font=\tiny, text=cGenV!70, above=0.05cm of zCrand, align=center] (lblRandC)
  {\textit{unconditional generation}};

\node[inner sep=0pt, minimum size=0pt]
  at (\xCR, 0 |- lblRandC.north) (pinCRtop){};
\node[inner sep=0pt, minimum size=0pt]
  at ([xshift=0.30cm]zCrand.east |- outC.south) (pinCRbot){};

\begin{pgfonlayer}{background}
  \node[stagebox=cGenV, dashed,
        fit=(hC)(lblSeedC)(seedC)(encC)(zCseed)
            (lblRandC)(zCrand)(pinCRtop)(pinCRbot)
            (decC)(outC)(pinC)]
       (boxC){};
\end{pgfonlayer}

\draw[arr=cGenV](seedC)--(encC);
\draw[arr=cLat] (encC)--(zCseed);

\coordinate (mergeCctr) at ([yshift=-4.20cm]hC.north);
\coordinate (mergeC)    at (\xCmid, 0 |- mergeCctr);

\draw[cGenV, semithick] (zCseed.south) -- (\xCL, 0 |- mergeC) -- (mergeC);
\draw[cGenV, semithick] (zCrand.south) -- (\xCR, 0 |- mergeC) -- (mergeC);
\draw[arr=cGenV] (mergeC) -- (decC.north);
\draw[arr=cGenV] (decC) -- (outC);

\node[sec, cGenC] at (\xD, 0 |- y0) (hD)
  {\textbf{(D) CVAE Generation}};

\node[inner sep=0pt, minimum size=0pt]
  at ([yshift=-\HCD]hD.north) (pinD){};

\node[blk=cCond,  below=0.45cm of hD, fill=cCond!10,
      minimum width=1.1cm, font=\scriptsize] (tgtLbl)
  {Target $y^{*}$};
\node[font=\tiny, text=cGenC!70, above=0.05cm of tgtLbl, align=center] (lblCondD)
  {\textit{label-conditioned generation}};
\node[tfblk=cGenC,below=0.55cm of tgtLbl, fill=cGenC!10] (encD)
  {Conditional Prior\\[-1pt]\tiny$p_{\theta}(\mathbf{z}|y)$ network};
\node[blk=cLat,   below=0.40cm of encD, fill=cLat!22,
      font=\scriptsize\bfseries] (zD)
  {$\mathbf{z}{\sim}p_{\theta}(\mathbf{z}|y^{*})$};
\node[blk=cFilm,  below=0.40cm of zD, fill=cFilm!12,
      font=\scriptsize] (filmD)
  {\textbf{FiLM:}\ $\tilde{\mathbf{z}}{=}\gamma(y^{*})\mathbf{z}{+}\beta(y^{*})$};
\node[tfblk=cGenC,below=0.40cm of filmD, fill=cGenC!10] (decD)
  {CVAE Decoder\\[-1pt]\tiny(LoRA fine-tuned)};
\node[blk=cGenC,  below=0.40cm of decD] (outD)
  {Molecules optimized for $y^{*}$};

\node[inner sep=0pt, minimum size=0pt]
  at ([xshift=0.80cm]filmD.east) (pinDR){};

\begin{pgfonlayer}{background}
  \node[stagebox=cGenC, dashed,
        fit=(hD)(lblCondD)(tgtLbl)(encD)(zD)(filmD)(decD)(outD)(pinD)(pinDR)]
       (boxD){};
\end{pgfonlayer}

\draw[arr=cCond] (tgtLbl)--(encD);
\draw[arr=cLat]  (encD)--(zD);
\draw[arr=cGenC] (zD)--(filmD);
\draw[arr=cGenC] (filmD)--(decD);
\draw[arr=cGenC] (decD)--(outD);

\coordinate (bypassX) at ([xshift=0.60cm]filmD.east);
\draw[arr=cCond]
  (tgtLbl.east) -- (bypassX |- tgtLbl.east) --
  (bypassX |- filmD.east) -- (filmD.east);

\draw[-Latex, thick, cPre, shorten >=5pt, shorten <=5pt]
  (boxA.east) -- node[above, font=\tiny\bfseries, cPre, align=center]
    {pretrained\\[-2pt]weights} (boxB.west);

\coordinate (yAC) at ([yshift=0.35cm]lproj.center);
\begin{pgfonlayer}{behind}
\draw[-Latex, semithick, cFilm, dotted, shorten >=5pt, shorten <=5pt]
  (boxA.east |- yAC) --
  node[above, font=\tiny, cFilm, align=center]
    {pretrained\\[-2pt]ckpt}
  (boxC.west |- yAC);
\end{pgfonlayer}

\begin{pgfonlayer}{behind}
\draw[-Latex, semithick, cFine, dashed, shorten >=5pt, shorten <=5pt]
  (boxB.east |- inB.center) --
  node[above, font=\tiny, cFine, align=center]
    {fine-tuned\\[-2pt]ckpt}
  (boxD.west |- inB.center);
\end{pgfonlayer}

\coordinate (EtopY) at ([yshift=\dimexpr\HE+0.30cm\relax]pinA);

\coordinate (EtopMidX) at ($(hC.center)!0.5!(hD.center)$);
\coordinate (EtopMid)  at (EtopMidX |- EtopY);

\node[sec, cLoRA] at (EtopMid) (hE)
  {\textbf{(E) Stage-1 LoRA mechanism}};

\coordinate (EpinTLtop) at (boxC.west |- EtopY);
\coordinate (EpinBRtop) at (boxD.east |- EtopY);

\node[inner sep=0pt, minimum size=0pt] at ([yshift=-\HE]EpinTLtop) (pinEL){};
\node[inner sep=0pt, minimum size=0pt] at ([yshift=-\HE]EpinBRtop) (pinER){};

\node[blk=cLoRA, below=-.1cm of hE, fill=cLoRA!6,
      minimum width=1.6cm, minimum height=0.44cm,
      font=\scriptsize] (xin) {Input $x$};

\node[blk=gray!55, below=0.05cm of xin, fill=gray!10,
      minimum width=3.0cm] (wfrozen)
  {Frozen $\mathbf{W}_0$};

\node[blk=cLoRA, below left=0.36cm and 0.55cm of wfrozen,
      fill=cLoRA!12, minimum width=1.2cm,
      font=\scriptsize\bfseries] (la)
  {$\mathbf{A}$\\[-2pt]\tiny$r{\times}d_\mathrm{in}$};

\node[blk=cLoRA, below right=0.36cm and 0.55cm of wfrozen,
      fill=cLoRA!12, minimum width=1.2cm,
      font=\scriptsize\bfseries] (lb)
  {$\mathbf{B}$\\[-2pt]\tiny$d_\mathrm{out}{\times}r$};

\node[circle, draw=cLoRA, fill=cLoRA!15, minimum size=0.45cm,
      below=0.80cm of wfrozen,
      font=\bfseries\scriptsize, text=cLoRA] (sm) {$+$};

\node[blk=cLoRA, below=0.20cm of sm, fill=cLoRA!8,
      font=\scriptsize, minimum width=3.8cm] (hout)
  {$h = \mathbf{W}_0 x + \tfrac{\alpha}{r}\,\mathbf{B}\mathbf{A}\,x
   \quad(r{=}8,\ \alpha{=}32)$};

\draw[arr=gray!55](xin)--(wfrozen);
\draw[arr=gray!55](wfrozen.south)--(sm.north);
\begin{pgfonlayer}{middle}
\draw[darr=cLoRA]
  (xin.center) -- ++(0,-0.10cm) -| (la.north);
\end{pgfonlayer}
\draw[arr=cLoRA](la.east)
  -- node[above, font=\tiny, pos=0.5]{rank $r{=}8$}(lb.west);
\draw[arr=cLoRA](la.south) |- (sm.west);
\draw[arr=cLoRA](lb.south) |- (sm.east);
\draw[arr=cLoRA](sm)--(hout);

\begin{pgfonlayer}{background}
  \node[stagebox=cLoRA,
        fit=(hE)(xin)(wfrozen)(la)(lb)(sm)(hout)(pinEL)(pinER)]
       (boxE){};
\end{pgfonlayer}

\draw[-Latex, semithick, cLoRA, dotted, shorten >=5pt, shorten <=5pt]
  (boxE.west) -- (boxE.west -| boxB.east);

\node[cbadge, anchor=south west] at (encA.north east) {\textbf{1}};
\node[cbadge, anchor=south west] at (decA.north east) {\textbf{2}};
\node[cbadge, anchor=south west] at (zA.north east)   {\textbf{3}};
\node[cbadge, anchor=south west] at (lbenc.north east){\textbf{4}};
\node[cbadge, anchor=south west] at (lbdec.north east){\textbf{4}};
\node[cbadge, anchor=north east] at ([xshift=10pt]hE.north east) {\textbf{4}};

\pgfresetboundingbox
\useasboundingbox (boxA.north west) rectangle (boxD.south east |- pinEL);

\end{tikzpicture}
   \end{adjustbox}
  \caption{%
    \textbf{\ModelName\ architecture overview.}
    Circled numbers mark the four principal architectural components:
    \textbf{\circled{1}}~bidirectional Transformer encoder,
    \textbf{\circled{2}}~autoregressive Transformer decoder,
    \textbf{\circled{3}}~latent-variable bottleneck, and
    \textbf{\circled{4}}~LoRA-based adaptation.
    \textbf{(A)}~Pretraining: a Transformer-based VAE (12-layer encoder/decoder,
    $d{=}512$, latent $\mathbb{R}^{256}$) is trained on 79\,M SELFIES
    molecules with a reconstruction + KL-divergence objective.
    \textbf{(B)}~Fine-tuning: CVAE (VAE + LoRA + FiLM): the pretrained backbone is frozen and augmented with LoRA adapters ($r{=}8$, $\alpha{=}32$) on the attention output projections and feed-forward projections of the encoder and decoder layers. A FiLM module modulates the latent vector with the property score $y$; a learned conditional prior $p(\mathbf{z}|y)$ replaces the isotropic Gaussian prior in the KL term. Which of these components is trainable differs by experiment (Table~\ref{tab:experiment-configurations}).
    \textbf{(C)}~VAE seed-based generation: a seed molecule is encoded to obtain $\mathbf{z}{\sim}q(\mathbf{z}|x_{\text{seed}})$, which is then decoded autoregressively to produce structurally similar molecules.
    \textbf{(D)}~CVAE label-conditioned generation: the requested label $y^{*}$ is passed to the conditional-prior network, which parameterizes $p_{\theta}(\mathbf{z}|y^{*})$; a latent sampled from this prior is FiLM-modulated and decoded autoregressively. The encoder is not used in this protocol.
    \textbf{(E)}~LoRA adapter mechanism (stage~1): each frozen weight $\mathbf{W}_0$ is augmented
    by a trainable low-rank perturbation $\tfrac{\alpha}{r}\mathbf{B}\mathbf{A}$
    ($r{=}8$, scaling ${=}4$). The panel illustrates the LoRA-specific update;
    per-experiment trainable modules differ, and the reported SA-score and docking
    models use a second stage in which LoRA is frozen and only the conditioning
    pathway is updated (Table~\ref{tab:experiment-configurations}).%
  }
  \begin{flushleft}\small Alt text: Five-panel diagram of the STAR-VAE architecture. Panel A: pretraining; a bidirectional Transformer encoder compresses SELFIES molecules to a 256-dimensional latent vector decoded autoregressively. Panel B: conditional fine-tuning; Transformer states are mean-pooled, the projected property score is added to the pooled representation before the posterior heads, and decoder-side Feature-wise Linear Modulation conditions the latent. Panels C and D: VAE generation via seed encoding or Gaussian prior sampling; Conditional VAE generation in which the requested label y-star is passed to the conditional-prior network, which supplies the latent vector. Panel E: stage-1 Low-Rank Adaptation rank-8 decomposition. Per-experiment trainable modules are reported in Table~\ref{tab:experiment-configurations}.\end{flushleft}
  \label{fig:star-vae}
\end{figure}

\ModelName\ (\textbf{S}\textsc{elfies-encoded}, \textbf{T}\textsc{ransformer-based}, 
\textbf{A}\textsc{uto}\textbf{R}\textsc{egressive}  \textbf{VAE}) consists of a bidirectional Transformer encoder, a latent-variable bottleneck, and an autoregressive Transformer decoder (Figure~\ref{fig:star-vae}). Both encoder and decoder use absolute positional encoding, consistent with the fixed maximum SELFIES length of 71 tokens. This architecture employs 12 encoder and 12 decoder layers, 8 attention heads, a feed-forward width of 2048, dropout 0.1, and a latent dimensionality of 256. \rev{Model size varies with the components used in each experiment. The pretrained unconditional backbone contains 89.18\,M parameters. Adding the label projection, conditional-prior networks, and FiLM modulation increases the total to 89.77\,M parameters. Adding encoder and decoder LoRA adapters increases it to 91.05\,M parameters. Table~\ref{tab:experiment-configurations} gives the total and trainable counts for every experiment.}

\paragraph{Encoder.}
The encoder is a stack of Transformer layers operating on SELFIES token embeddings; contextualized hidden states are mean-pooled over sequence positions to produce the mean and log-variance of a Gaussian latent distribution. In the label-conditioned configurations, a learned linear projection of the scalar label is added to the pooled representation before the latent mean and log-variance heads, so that the encoder parameterizes $q_\phi(\mathbf{z}|x,y)$. LoRA adapters may additionally be applied within the encoder layers to adapt the backbone to the task (Table~\ref{tab:experiment-configurations}).

\paragraph{Latent bottleneck.}
The latent vector is sampled via the reparameterization trick, linearly projected to the decoder width, and repeated across memory positions to serve as decoder cross-attention memory.

\paragraph{Decoder.}
The autoregressive Transformer decoder generates SELFIES tokens one at a time, conditioned on previously generated tokens and on the latent-derived cross-attention memory. In label-conditioned configurations, the property value enters through FiLM modulation of the latent vector before it is projected to decoder memory: a small network maps the scalar label to per-dimension scale and shift parameters $(\gamma(y), \beta(y))$, and the modulated latent is $\gamma(y) \odot \mathbf{z} + \beta(y)$. LoRA adapters may be applied in the decoder layers to adapt the backbone while keeping the majority of parameters frozen.

\subsection{Pretraining Setup}

The model is trained with Adam ($\text{lr}=10^{-5}$, batch 256) on 8 GPUs for approximately 72 hours, with no learning-rate schedule. \rev{The KL coefficient follows a linear warmup from 0 to $\beta=1.0$ over the first 10{,}000 optimizer steps and is held at $\beta=1.0$ thereafter. The cyclical $\beta$ annealing and free-bits options described in the Ablation Studies apply to conditional fine-tuning, not to this pretraining run.}

\subsection{Conditional Generative Formalism}
After pretraining, we fine-tune the model on a label-guided objective where label $y$ encodes a property of interest (activity, ADMET endpoint, or multi-component objective). We formalize conditional generation via three distributions:

\begin{align}
    p_\theta(z \mid y) &= \text{prior over latents given property } y, \\
    q_\phi(z \mid x, y) &= \text{approximate posterior inferred from molecule $x$ and property $y$}, \\
    p_\theta(x \mid z, y) &= \text{autoregressive decoder generating molecule $x$ given $(z,y)$}.
\end{align}

The inference network is modeled as a Gaussian:
\begin{equation}
    q_\phi(z \mid x, y) = \mathcal{N}\!\left(\mu_\phi(x,y), \; \mathrm{diag}(\sigma_\phi^2(x,y)) \right),
\end{equation}
where $\mu_\phi$ and $\sigma_\phi$ are outputs of the Transformer encoder.  
The label-conditioned prior is also Gaussian:
\begin{equation}
    p_\theta(z \mid y) = \mathcal{N}\!\left(\mu_\theta(y), \; \mathrm{diag}(\sigma_\theta^2(y)) \right),
\end{equation}
with parameters predicted from the property embedding $y$.

The decoder defines the conditional likelihood:
\begin{equation}
    p_\theta(x \mid z, y) = \prod_{t=1}^{T} p_\theta\!\left(x_t \mid x_{<t}, z, y \right).
\end{equation}

Training maximizes the conditional evidence lower bound (ELBO):
\begin{equation}
    \mathcal{L}(x,y) = 
    \mathbb{E}_{q_\phi(z \mid x,y)} \big[ \log p_\theta(x \mid z, y) \big]
    - \beta \, \mathrm{KL}\!\left[ q_\phi(z \mid x,y) \,\|\, p_\theta(z \mid y) \right],
\end{equation}
where $\beta$ controls the strength of the KL regularization.

\paragraph{LoRA-based task adaptation.}
To adapt the pretrained backbone to each conditional task, we insert LoRA adapters into the Transformer encoder and decoder layers.
Adapters are inserted at the attention \emph{output} projections (\(W_O\) in encoder self-attention, and in both decoder self-attention and cross-attention) and at the two position-wise feed-forward projections of each layer.
The fused query/key/value projection of each attention block is left unmodified.
\rev{One implementation detail materially qualifies this description. The attention modules dispatch to a fused attention kernel that consumes the output-projection weight directly rather than calling the wrapped module, so the adapters inserted at $W_O$ are never evaluated in the forward pass. Their $B$ matrices remain at their zero initialization and receive no gradient. We verified this by gradient inspection and by checking the released checkpoints, in which all attention-output $B$ matrices are exactly zero while the feed-forward $B$ matrices are not. In effect, therefore, LoRA adapts the position-wise feed-forward projections only. These inert adapters account for 294{,}912 of the 1{,}277{,}952 LoRA parameters (23\%), so the reference ablation configuration trains approximately 1.64\,M effective parameters rather than the 1.94\,M reported as trainable. Parameter counts elsewhere in this paper are reported as instantiated, since that is what the checkpoints contain. A unit test accompanying the released code pins this behaviour. The attention-output LoRA implementation issue does not alter the reported results because those adapters remained at zero and were inactive. Effective feed-forward LoRA was trained in Stage 1, inherited and frozen in Stage 2, and remains active in the decoder during inference (Table~\ref{tab:experiment-configurations}).}
For each adapted linear map \(W_{\bullet}\),
\begin{equation}
    \tilde{W}_{\bullet} = W_{\bullet}^{(0)} + \tfrac{\alpha}{r}\, B_{\bullet} A_{\bullet},
\end{equation}
where \(W_{\bullet}^{(0)}\) are the frozen pretrained weights, \(A_{\bullet} \in \mathbb{R}^{r \times d_{\text{in}}}\), and \(B_{\bullet} \in \mathbb{R}^{d_{\text{out}} \times r}\) with \(r \ll d\).
LoRA adapts the task-specific weights; it does not itself introduce the input-dependent property value.
The property value enters through the label-conditioned prior, the label projection in the approximate posterior, and the decoder-side FiLM modulation described above.
\rev{The scaling factor $\alpha/r$ is fixed and embedded into the trained matrices in accordance with standard LoRA practice~\cite{hu2022lora}. In all experiments, $r = 8$ and $\alpha = 32$, resulting in a fixed scaling factor of 4.}
\old{The scalar \(\lambda\) controls the strength of conditioning at inference and can optionally absorb the LoRA scaling factor \(\alpha / r\).}

\subsection{\rev{Per-Experiment Configurations}}
\label{sec:experiment-configurations}

\rev{The conditional components described above are not all trainable, or even all present, in every experiment. Table~\ref{tab:experiment-configurations} therefore reports, for each reported model, which modules are trainable, whether the label enters the approximate posterior, whether the label-conditioned prior exists and is trained, whether it is sampled from at generation time, and the exact trainable and total parameter counts.}

\rev{Two aspects of this table warrant emphasis, because they qualify how the reported results should be interpreted. First, the SA-score and docking models were adapted in two stages. LoRA adapters, FiLM, and the latent-to-hidden projection were trained in the first stage. Those weights were then frozen, and only the conditioning pathway, including the conditional prior, was trained in the second stage. The reported models are the product of both stages, and the conditional-prior parameters are updated only in the second. Second, the ablation study samples latents from $\mathcal{N}(\mathbf{0},\mathbf{I})$ rather than from $p_\theta(\mathbf{z}|y)$ for all variants, so it characterizes the FiLM and LoRA conditioning pathways but does not isolate the contribution of the learned conditional prior.}

\begin{table}[H]
\centering
\caption{\rev{\textbf{Per-experiment model configuration, conditioning pathway, and parameter counts.}
Trainable and total parameter counts are derived from the freezing logic in the released code and cross-checked against training logs. ``Post.\ label'' indicates whether the scalar label is projected into the pooled encoder representation before the latent heads, i.e.\ whether the encoder parameterizes $q_\phi(\mathbf{z}|x,y)$ rather than $q_\phi(\mathbf{z}|x)$. ``Prior'' columns distinguish whether the label-conditioned prior $p_\theta(\mathbf{z}|y)$ exists, whether its parameters are updated, and whether it is sampled from at generation time. Where the prior exists but is frozen, it contributes to the KL term with randomly initialized parameters. The SA-score and docking models were adapted in two stages (see footnotes).}}
\label{tab:experiment-configurations}
\smallskip
\footnotesize
\setlength{\tabcolsep}{4pt}
\begin{adjustbox}{max width=\linewidth}
\begin{tabular}{@{}llccccccrr@{}}
\toprule
 & & \multicolumn{2}{c}{LoRA} & Post. & \multicolumn{3}{c}{Conditional prior} & \multicolumn{2}{c}{Parameters} \\
\cmidrule(lr){3-4}\cmidrule(lr){6-8}\cmidrule(lr){9-10}
Model / experiment & FiLM & Enc. & Dec. & label & exists & trained & sampled & Trainable & Total \\
\midrule
Pretrained VAE backbone            & --      & --      & --      & --      & --      & --      & --      & 89{,}175{,}368 & 89{,}175{,}368 \\
\midrule
\multicolumn{10}{@{}l}{\emph{SA-score and docking conditional models (two-stage adaptation)}} \\
Stage 1: LoRA adaptation$^{a}$     & $\checkmark$ & $\checkmark$ & $\checkmark$ & $\checkmark$ & $\checkmark$ & --      & --      & 1{,}936{,}384 & 91{,}047{,}752 \\
Stage 2: SA score (reported)$^{b}$ & $\checkmark$ & frozen  & frozen  & $\checkmark$  & $\checkmark$ & $\checkmark$ & $\checkmark$ & 726{,}016 & 91{,}047{,}752 \\
Stage 2: 1SYH / 4LDE / 6Y2F$^{b}$  & $\checkmark$ & frozen  & frozen  & $\checkmark$  & $\checkmark$ & $\checkmark$ & $\checkmark$ & 726{,}016 & 91{,}047{,}752 \\
\midrule
Rediscovery CVAE$^{c}$             & $\checkmark$ & --      & $\checkmark$ & --      & --      & --      & --      & 1{,}346{,}560 & 90{,}391{,}368 \\
\midrule
\multicolumn{10}{@{}l}{\emph{Ablations (SA-score setting; all generate from $\mathcal{N}(\mathbf{0},\mathbf{I})$, not from $p_\theta(\mathbf{z}|y)$)$^{d}$}} \\
REF (reference)                    & $\checkmark$ & $\checkmark$ & $\checkmark$ & $\checkmark$ & $\checkmark$ & --      & --      & 1{,}936{,}384 & 91{,}047{,}752 \\
A2 deterministic AE                & $\checkmark$ & $\checkmark$ & $\checkmark$ & $\checkmark$ & $\checkmark$ & --      & --      & 1{,}936{,}384 & 91{,}047{,}752 \\
A3 encoder-only conditioning       & --      & $\checkmark$ & --      & $\checkmark$ & $\checkmark$ & --      & --      & 721{,}408 & 89{,}832{,}776 \\
A3 decoder-only conditioning       & $\checkmark$ & --      & $\checkmark$ & --      & $\checkmark$ & --      & --      & 1{,}346{,}560 & 90{,}457{,}928 \\
A3 no conditioning$^{e}$           & --      & $\checkmark$ & $\checkmark$ & $\checkmark$ & --      & --      & --      & 1{,}409{,}536 & 90{,}454{,}344 \\
A4 full fine-tuning                & $\checkmark$ & --      & --      & $\checkmark$ & $\checkmark$ & $\checkmark$ & --      & 89{,}769{,}800 & 89{,}769{,}800 \\
A4 frozen backbone                 & $\checkmark$ & --      & --      & $\checkmark$ & $\checkmark$ & $\checkmark$ & --      & 988{,}672 & 89{,}769{,}800 \\
A5 fixed $\beta$ / linear $\beta$  & $\checkmark$ & $\checkmark$ & $\checkmark$ & $\checkmark$ & $\checkmark$ & --      & --      & 1{,}936{,}384 & 91{,}047{,}752 \\
\bottomrule
\end{tabular}
\end{adjustbox}

\begin{minipage}{\linewidth}\footnotesize
$^{a}$~Counts are instantiated counts. Attention-output LoRA adapters are bypassed by the fused attention kernel; 294{,}912 both-side LoRA parameters are therefore inert. Stage~1 trains the effective feed-forward LoRA, FiLM, and latent-to-hidden parameters, while the conditional prior and posterior label projection remain frozen.\\
$^{b}$~The reported SA-score and docking models add a second stage in which all Transformer and LoRA weights are frozen and only FiLM, the conditional-prior networks, latent-to-hidden projection, and label projection are updated. The posterior label projection is updated during Stage-2 training but is inactive during label-only generation because the encoder is not used.\\
$^{c}$~The rediscovery CVAE uses decoder-only LoRA, no conditional prior, and $\mathbf{z}\sim\mathcal{N}(\mathbf{0},\mathbf{I})$ followed by FiLM; counts follow the released configuration because the original checkpoints were unavailable.\\
$^{d}$~All ablations generate from $\mathcal{N}(\mathbf{0},\mathbf{I})$, so the study does not isolate conditional-prior sampling.\\
$^{e}$~The ``no conditioning'' configuration disables the prior and FiLM but still routes training through a frozen randomly initialized label projection; labels have no effect at generation. In rows marked ``--'' under FiLM, no modulation module is constructed.
\end{minipage}
\end{table}
 
\subsection{Generation Protocols}
\label{sec:generation-protocols}

\ModelName\ supports three generation protocols, distinguished by what information is available at inference time.

\paragraph{Protocol~I: Unconditional generation (VAE).}
The latent vector is sampled directly from the standard Gaussian prior: $\mathbf{z} \sim \mathcal{N}(\mathbf{0}, \mathbf{I})$.
No seed molecule and no property label are required.
The pretrained decoder then generates molecules autoregressively from $\mathbf{z}$.
This protocol is used for the MOSES and GuacaMol ``Random'' unconditional benchmarks.

\paragraph{Protocol~II: Seed-based generation (VAE).}
A seed molecule $x_{\text{seed}}$ is first encoded by the pretrained encoder to obtain a posterior sample $\mathbf{z} \sim q_\phi(\mathbf{z}|x_{\text{seed}})$.
The pretrained decoder then generates new molecules from $\mathbf{z}$, producing outputs that are structurally similar to the seed.
No property label is used.
This protocol is used for the GuacaMol ``Seed'' experiment and for the VAE arm of the rediscovery evaluation.

\paragraph{Protocol~III: Label-conditioned generation (CVAE).}
A target label $y^{*}$ (such as a target-binding activity level or an ADMET endpoint) is the sole input at inference time; no seed molecule is encoded.
The label is passed to the conditional-prior network, which parameterizes $p_\theta(\mathbf{z}|y^{*})$; a latent vector is sampled from this prior, FiLM-modulated, and decoded autoregressively.
The encoder is not used in this protocol, since it represents $q_\phi(\mathbf{z}|x,y)$ and requires a molecule.
This protocol is used for the Tartarus protein-ligand design tasks.
\rev{For the CVAE arm of the rediscovery evaluation, the conditional prior was not enabled in the fine-tuned model. Latents were instead drawn from the isotropic Gaussian $\mathcal{N}(\mathbf{0},\mathbf{I})$ and FiLM-modulated with the requested label before decoding. Table~\ref{tab:experiment-configurations} reports this configuration.}

\paragraph{\rev{Label-reference decoder guidance.}}
\rev{At inference, we can strengthen the effect of the requested property label by combining two decoder predictions. We first draw one latent vector from the conditional prior, $\mathbf{z} \sim p_\theta(\mathbf{z}|y^{*})$. We then apply FiLM to this same latent vector with the requested label $y^{*}$ and with the reference label $y_{\varnothing}=0$. At each decoding step, both branches use the same token prefix and produce logit vectors $\boldsymbol{\ell}_{y^{*}}$ and $\boldsymbol{\ell}_{\varnothing}$. We combine them as}
\begin{equation}
    \boldsymbol{\ell}_{\text{guided}} = \boldsymbol{\ell}_{\varnothing} + \alpha \, \bigl( \boldsymbol{\ell}_{y^{*}} - \boldsymbol{\ell}_{\varnothing} \bigr),
    \label{eq:cfg}
\end{equation}
\rev{where $\alpha \geq 1$ controls the guidance strength. At $\alpha=1$, the equation gives the prediction for the requested label. Larger values increase the difference between the requested-label and reference-label predictions. The decoder selects the token with the highest guided logit at each step.}

\rev{This procedure is inspired by classifier-free guidance~\cite{ho2022classifierfree}, but it is not the same method. The models were trained without condition dropout, and $y_{\varnothing}=0$ is a real endpoint of the property scale rather than a learned null label. Both branches also share the same latent vector and token prefix. The procedure therefore adjusts the decoder logits relative to a reference property value without adding latent sampling noise.}
 
\section{Results}
\label{sec:results}
\subsection{Unconditional Generation}
\begin{table}[H]
\centering
\caption{\revfloat{Comparison of models on the MOSES benchmark using seven key metrics. Modern baselines (SAFE-GPT, DiGress, FragFM, MolHIT, GVT) do not report all metrics. Columns that are uniformly absent for modern baselines have been omitted here (see Table~S1 in the Supporting Information for the full 15-column table). Baseline values for classic models are from the official MOSES repository~\cite{polykovskiy2020molecular}. Modern baseline values are from the cited publications. \ModelName\ runs are means $\pm$ SD over three independent evaluation seeds. \textbf{Bold} indicates the best value per column across all models with reported data. Column headers: Validity (Valid), Uniqueness (Unique@10k), Fr\'{e}chet ChemNet Distance (FCD/Test, lower is better), Nearest Neighbor Similarity (SNN/Test), Scaffold Similarity (Scaf/Test), drug-likeness filter pass rate (Filters), and Novelty.}}
\label{tab:model_comparison}
\footnotesize
\setlength{\tabcolsep}{3pt}
\begin{adjustbox}{max width=\linewidth}
\begin{tabular}{lccccccc}
\toprule
\textbf{Models} & \textbf{Valid (↑)} & \textbf{Unique@10k (↑)} & \textbf{FCD/Test (↓)} & \textbf{SNN/Test (↑)} & \textbf{Scaf/Test (↑)} & \textbf{Filters (↑)} & \textbf{Novelty (↑)} \\
\midrule
\textit{Classic baselines} & & & & & & & \\
HMM & 0.076 & 0.5671 & 24.4661 & 0.3876 & 0.2065 & 0.9024 & 0.9994 \\
NGram & 0.2376 & 0.9217 & 5.5069 & 0.5209 & 0.5302 & 0.9582 & 0.9694 \\
Combinatorial & \textbf{1.0} & 0.9909 & 4.2375 & 0.4514 & 0.4445 & 0.9557 & 0.9878 \\
CharRNN & 0.9748 & 0.9994 & \textbf{0.0732} & 0.6015 & 0.9242 & \textbf{0.9943} & 0.8419 \\
AAE & 0.9368 & 0.9973 & 0.5555 & 0.6081 & 0.9022 & 0.9960 & 0.7931 \\
VAE & 0.9767 & 0.9984 & 0.0990 & 0.6257 & \textbf{0.9386} & 0.9970 & 0.6949 \\
JTN-VAE & \textbf{1.0} & 0.9996 & 0.3954 & 0.5477 & 0.8964 & 0.9760 & 0.9143 \\
LatentGAN & 0.8966 & 0.9968 & 0.2968 & 0.5371 & 0.8867 & 0.9735 & 0.9498 \\
\midrule
\rev{\textit{Modern baselines}} & & & & & & & \\
\rev{SAFE-GPT~\cite{noutahi2024safe}} & \rev{0.998} & \rev{0.989} & \rev{0.72} & \rev{0.57} & \rev{0.063} & \rev{0.977} & \rev{0.437} \\
\rev{DiGress~\cite{vignac2023digress}} & \rev{0.857} & \rev{\textbf{1.000}} & \rev{1.19} & \rev{0.52} & \rev{0.148} & \rev{0.971} & \rev{0.950} \\
\rev{FragFM~\cite{kim2025fragfm}} & \rev{0.998} & \rev{\textbf{1.000}} & \rev{0.58} & \rev{0.56} & \rev{0.109} & \rev{0.991} & \rev{0.871} \\
\rev{MolHIT~\cite{jung2026molhit}} & \rev{0.991} & \rev{0.998} & \rev{1.03} & \rev{0.55} & \rev{0.144} & \rev{0.980} & \rev{0.914} \\
\rev{GVT~\cite{zheng2025gvt}} & \rev{0.994} & \rev{\textbf{1.000}} & \rev{0.16} & \rev{\textbf{0.64}} & \rev{0.034} & \rev{0.996} & \rev{0.614} \\
\midrule
\textit{Ours (\ModelName)} & & & & & & & \\
PubChem 80M (Random) & \textbf{1.0000}$\pm$0.0000 & \textbf{1.0000}$\pm$0.0000 & 14.1140$\pm$0.0607 & 0.3140$\pm$0.0002 & 0.2831$\pm$0.0095 & 0.4766$\pm$0.0014 & \textbf{0.9999}$\pm$0.0000 \\
PubChem 80M (Seed) & \textbf{1.0000}$\pm$0.0000 & 0.9998$\pm$0.0000 & 2.0062$\pm$0.0084 & 0.5090$\pm$0.0007 & 0.8987$\pm$0.0029 & 0.8313$\pm$0.0013 & 0.7583$\pm$0.0010 \\
\rev{MOSES 1.5M (Random)} & \rev{\textbf{1.0000}$\pm$0.0000} & \rev{0.9999$\pm$0.0001} & \rev{9.1489$\pm$0.0291} & \rev{0.3842$\pm$0.0003} & \rev{0.5622$\pm$0.0120} & \rev{0.6398$\pm$0.0029} & \rev{0.9939$\pm$0.0002} \\
\rev{MOSES 1.5M (Seed)} & \rev{\textbf{1.0000}$\pm$0.0000} & \rev{0.9996$\pm$0.0002} & \rev{6.6121$\pm$0.0335} & \rev{0.4263$\pm$0.0003} & \rev{0.7583$\pm$0.0057} & \rev{0.6931$\pm$0.0018} & \rev{0.9567$\pm$0.0004} \\
\bottomrule
\end{tabular}
\end{adjustbox}
\end{table}
 
\rev{To establish a task-specific baseline alongside the PubChem-80M model, we trained a smaller variant of \ModelName\ (denoted \ModelName{}-MOSES) directly on the official MOSES training set. The architecture uses $d_{\text{model}} = 256$, $n_{\text{head}} = 4$, 4 encoder layers, 4 decoder layers, a feedforward dimensionality of $1{,}024$, and a $128$-dimensional latent bottleneck, yielding approximately 7.7 million parameters (versus 89.18 million for the PubChem-80M model). Training used SMILES converted to SELFIES, with maximum sequence length 74 tokens, for 15 epochs with the Adam optimizer, learning rate $0.0002$, and batch size 256 on a single GPU. For evaluation, we generate exactly $30{,}000$ molecules following the official MOSES benchmark experiments~\cite{polykovskiy2020molecular}. Protocol I (Random) draws latent vectors from $\mathcal{N}(\mathbf{0}, \mathbf{I})$. Protocol II (Seed) encodes 30,000 randomly selected training molecules, adds a small perturbation around each posterior mean, and decodes one molecule per seed. The resulting molecules are evaluated offline against the official MOSES splits.}

\rev{On the MOSES~\cite{polykovskiy2020molecular} benchmark, we evaluate \ModelName\ under two pretraining datasets (PubChem 80M and MOSES 1.5M) and two generation protocols (Unconditional Random Prior sampling and Seed-Conditioned perturbation), with means and standard deviations over three independent evaluation seeds (Table~\ref{tab:model_comparison}).}

\begin{sloppypar}
\rev{Under the \textbf{Unconditional (Random)} setting ($\mathbf{z} \sim \mathcal{N}(\mathbf{0}, \mathbf{I})$), both models achieve perfect Validity ($1.0000 \pm 0.0000$), near-perfect Uniqueness@10K, and high Novelty, with IntDiv1 $= 0.8990 \pm 0.0001$ for PubChem, above all classic baselines. FCD/Test is higher ($14.1140 \pm 0.0607$ and $9.1489 \pm 0.0291$) and scaffold similarity is lower, reflecting unconstrained sampling from a prior trained on PubChem rather than on the MOSES corpus.}
\end{sloppypar}

\rev{Under the \textbf{Seed-Conditioned} protocol, 30,000 randomly selected training SMILES are encoded and their latent vectors perturbed with $\epsilon \sim \mathcal{N}(0, I)$ before decoding. \ModelName\ PubChem 80M (Seed) reduces FCD/Test from $14.1140$ to $2.0062 \pm 0.0084$, a substantial improvement over the Random prior but still above the classic baselines trained on the MOSES corpus (CharRNN: $0.0732$, AAE: $0.5555$). Nearest Neighbor Similarity increases from $0.3140$ to $0.5090 \pm 0.0007$. Fragment Similarity and Scaffold Similarity reach $0.9964 \pm 0.0001$ and $0.8987 \pm 0.0029$. Generation remains within the topological neighborhood of the seeds without mode collapse (Unique@10K $= 0.9998 \pm 0.0000$).}

\old{Comparing the two pre-training regimes under seed conditioning illustrates the value of large-scale pretraining. Despite never being trained on MOSES molecules, \ModelName PubChem 80M (Seed) substantially outperforms the MOSES 1.5M model trained directly on the target distribution.}\rev{Under seed conditioning, \ModelName\ PubChem 80M outperforms the MOSES 1.5M model on FCD/Test ($2.0062 \pm 0.0084$ vs.\ $6.6121 \pm 0.0335$), SNN/Test ($0.5090 \pm 0.0007$ vs.\ $0.4263 \pm 0.0003$), and filter pass rate ($0.8313 \pm 0.0013$ vs.\ $0.6931 \pm 0.0018$). This comparison should be read with two asymmetries in view, both of which favour the MOSES-1.5M arm. First, that model was trained on the official MOSES training split and used the official MOSES test split for validation and checkpoint selection, so the reported FCD/Test and SNN/Test are computed against molecules it selected checkpoints on. Second, the seed protocol draws its 30{,}000 seeds from the MOSES training split, which the MOSES-1.5M model saw directly during training. \ModelName\ PubChem 80M received no parameter updates on any MOSES split, but an overlap audit (Appendix~F, Table~S5) shows that 71.45\% of MOSES training molecules and 71.18\% of MOSES test molecules match the full curated PubChem source corpus, providing upper bounds on exposure to this chemistry. The defensible reading is therefore narrow. Corpus-scale pretraining yields a latent manifold that supports better local sampling around a supplied seed than a model trained on the benchmark distribution itself, even though the smaller model had direct access to both the seeds and the evaluation split. Neither arm demonstrates transfer to chemistry absent from pretraining.}

GuacaMol evaluates distributional similarity via KL divergence across ten RDKit physicochemical descriptors between generated and reference molecules; higher scores indicate closer alignment (Table~\ref{tab:mol_props}).
\rev{Four baselines (CharRNN~\cite{segler2018generating}, VAE~\cite{DBLP:journals/corr/KingmaW13}, AAE~\cite{DBLP:journals/corr/MakhzaniSJG15}, MolGPT~\cite{DBLP:journals/jcisd/BagalAVP22}) were retrained on the official GuacaMol split and each sampled for 10,000 molecules, providing full per-descriptor KL values for Table~\ref{tab:mol_props}.
Four additional baselines (DiGress~\cite{vignac2023digress} 0.929, MolHIT~\cite{jung2026molhit} 0.967, FragFM~\cite{kim2025fragfm} 0.974, GVT~\cite{zheng2025gvt} 0.996) are taken from their respective publications, which report only a 5-column summary (Validity, Uniqueness, Novelty, KL Divergence Score, FCD). Per-descriptor breakdowns are not available, so a row-by-row comparison with Table~\ref{tab:mol_props} is not possible for these models.
\ModelName~(Random) scores 0.9157 and \ModelName~(Seed) scores 0.9979 on the same overall KL Divergence Score scale.}

\begin{table}[H]
\caption{Molecular physicochemical distribution scores on the GuacaMol benchmark.
The first row (KL Divergence Score) is the overall GuacaMol score $\in [0,1]$
(\emph{higher is better\,$\uparrow$}), computed as the mean of $\exp(-\mathrm{KL})$ across
ten descriptors. The remaining rows report raw per-descriptor KL divergence
values (\emph{lower is better\,$\downarrow$}). Bold indicates the best value per row.
\ModelName~(Random) samples from the pretrained PubChem prior; \ModelName~(Seed)
samples from latent encodings of GuacaMol training molecules.}
\label{tab:mol_props}
\scriptsize
\centering
\begin{threeparttable}
\begin{tabular}{lccccccc}
\toprule
\multirow{2}{*}{\textbf{Metric}} & \multirow{2}{*}{\textbf{CharRNN}} &  \multirow{2}{*}{\textbf{VAE}} &  \multirow{2}{*}{\textbf{AAE}} &  \multirow{2}{*}{\textbf{MolGPT}} &   \multicolumn{2}{c}{\textbf{\ModelName~(ours)}} \\
& & & & & Random & Seed \\
\midrule
KL Divergence Score$^\dagger$ ($\uparrow$) & .8894 & .9173 & .9356 & .8817 & .9157 & \textbf{.9979} \\
\midrule
BertzCT                   & .3460 & .2617 & .1660 & .3926 & \textbf{.0274} & \textbf{.0012} \\
MolLogP                   & .0684 & .0723 & .0714 & .0744 & \textbf{.0105} & \textbf{.0026} \\
MolWt                     & .2337 & .1986 & .1358 & .2804 & \textbf{.0076} & \textbf{.0018} \\
TPSA                      & .1092 & .0725 & .0643 & .1051 & \textbf{.0052} & \textbf{.0015}\\
NumHAcceptors             & .1414 & .0852 & .0687 & .1549 & \textbf{.0055} & \textbf{.0008} \\
NumHDonors                & .0234 & .0203 & .0223 & .0149 & .0199 & \textbf{.0005} \\
NumRotatableBonds         & .0075 & .0141 & \textbf{.0067} & .0196 & .0083 & .0053 \\
NumAliphaticRings         & .0116 & .0069 & .0107 & .0117 & .0414 & \textbf{.0035}\\
NumAromaticRings          & .2047 & .1355 & .1027 & .1779 & .1846 & \textbf{.0007}\\
DiffInternalSimilarities  & .0797 & .0280 & .0298 & .0943 & .7993 & \textbf{.0031}\\
\bottomrule
\end{tabular}
\begin{tablenotes}
\small
\item[$\dagger$] The overall KL Divergence Score ($\uparrow$) is the mean of
$\exp(-\mathrm{KL}_i)$ across all ten descriptors, normalized to $[0,1]$.
Per-descriptor raw KL values ($\downarrow$): lower is better.
\end{tablenotes}
\end{threeparttable}
\end{table} 

\old{Under the GuacaMol evaluation protocol, \ModelName, pretrained on the full PubChem dataset in contrast to the benchmark's ChEMBL-derived training set, attains an overall KL-divergence score of 0.916 (Table~\ref{tab:mol_props}, ``Random'' column), indicating close alignment between the generated and reference distributions of key physicochemical descriptors.}
\old{The model achieves particularly low KL values for \texttt{BertzCT} (0.062), \texttt{MolLogP} (0.045), \texttt{MolWt} (0.048), and \texttt{TPSA} (0.056), outperforming all baseline methods on these metrics.}
\old{These results demonstrate that \ModelName\ reproduces the marginal property distributions of the reference dataset with high fidelity, with only minor deviations observed for aromatic ring counts.}
\rev{Under the GuacaMol evaluation protocol, \ModelName, pretrained on the full PubChem dataset in contrast to the benchmark's ChEMBL-derived training set, attains an overall KL-divergence score of 0.9157 (\autoref{tab:mol_props}, ``Random'' column). The model achieves the lowest KL divergence among the four retrained baselines on five descriptors: BertzCT (0.0274), MolLogP (0.0105), MolWt (0.0076), TPSA (0.0052), and NumHAcceptors (0.0055). Two descriptors show substantially higher KL divergence than all baselines: NumAromaticRings (0.1846, vs.\ AAE best of 0.1027) and DiffInternalSimilarities (0.7993, vs.\ VAE best of 0.0280). DiffInternalSimilarities measures the per-molecule spread of pairwise Tanimoto similarities to the rest of the generated set. The broader PubChem chemical space yields a wider spread than ChEMBL24, raising this KL divergence.}

The higher internal dissimilarity under GuacaMol reflects the distributional mismatch between the PubChem pretraining corpus and the ChEMBL24 reference set. To quantify this, we generated 10{,}000 molecules from latent encodings of randomly selected GuacaMol training-set molecules (the ``Seed'' column). This setting yields the lowest KL divergences across all descriptors. \rev{Because the seeds are drawn from the GuacaMol training set itself, this protocol measures reconstruction within a supplied domain rather than unconditional distribution matching, and its scores are not comparable to unconditional benchmark results. The overlap audit further shows that 87.04\% of GuacaMol training molecules match the full curated PubChem source corpus (Appendix~F), providing an upper bound on exposure to the reconstructed domain. The result establishes that seed-based posterior sampling can localize generation to a narrower property distribution without fine-tuning. It does not establish generalization to chemistry absent from pretraining.}

\subsection{Conditional Generation}
\label{sec:conditional-generation}

\old{Hereafter, ``CVAE'' (Conditional VAE) denotes the fine-tuned conditional model and ``VAE'' the pre-trained model sampling from the unconditional prior.}
\rev{Hereafter, ``CVAE'' denotes a fine-tuned conditional model and ``pretrained VAE'' denotes the base model. The sampling protocol (random prior or seed-based posterior) is stated separately. We evaluate conditioning on two directly computable scores: the RDKit SA score for synthetic accessibility and the QVina docking score for protein--ligand binding.}

\paragraph{SA score.}
\label{sec:docking-score-conditioned}

\begin{figure}[H]
    \centering
    \begin{subfigure}[t]{0.48\linewidth}
        \centering
        \includegraphics[alt={UMAP scatterplot of the STAR-VAE (SELFIES-encoded Transformer-based AutoRegressive VAE) latent space colored by synthetic accessibility score bin; the gradient runs from low SA score (easier to synthesize) through intermediate bins to high SA score (harder to synthesize), with smooth overlap between adjacent bins.}, width=\linewidth]{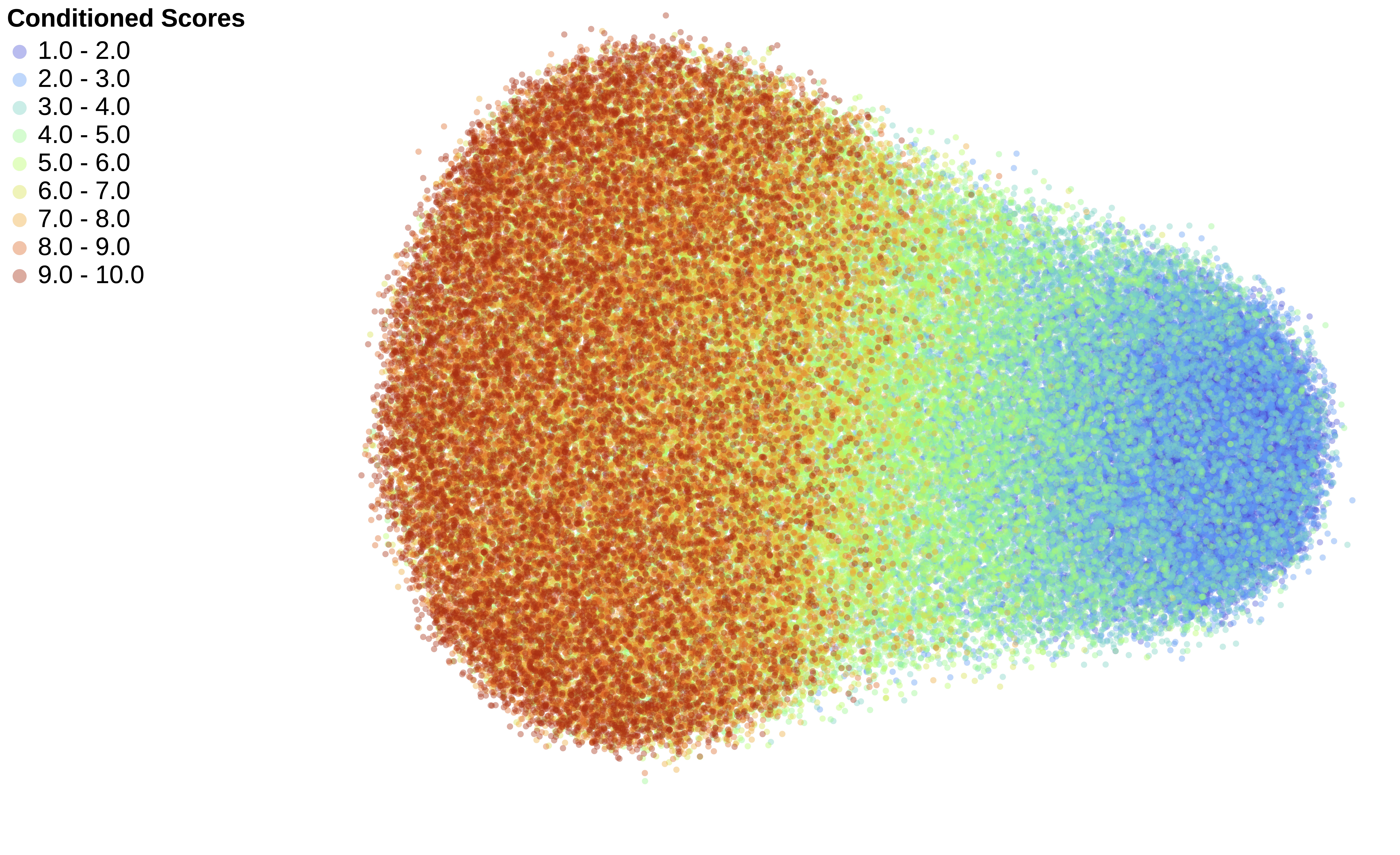}
        \caption{SA score}
        \label{fig:latent-umap-sascore}
    \end{subfigure}
    \hfill
    \begin{subfigure}[t]{0.48\linewidth}
        \centering
        \includegraphics[alt={UMAP scatterplot of the Conditional VAE latent space for protein target 1SYH (GluR2 ligand-binding core); molecules color-coded from dark blue (docking score near 0, weak binding) to yellow (most negative docking score, strong binding), forming distinct spatial clusters by score level.}, width=\linewidth]{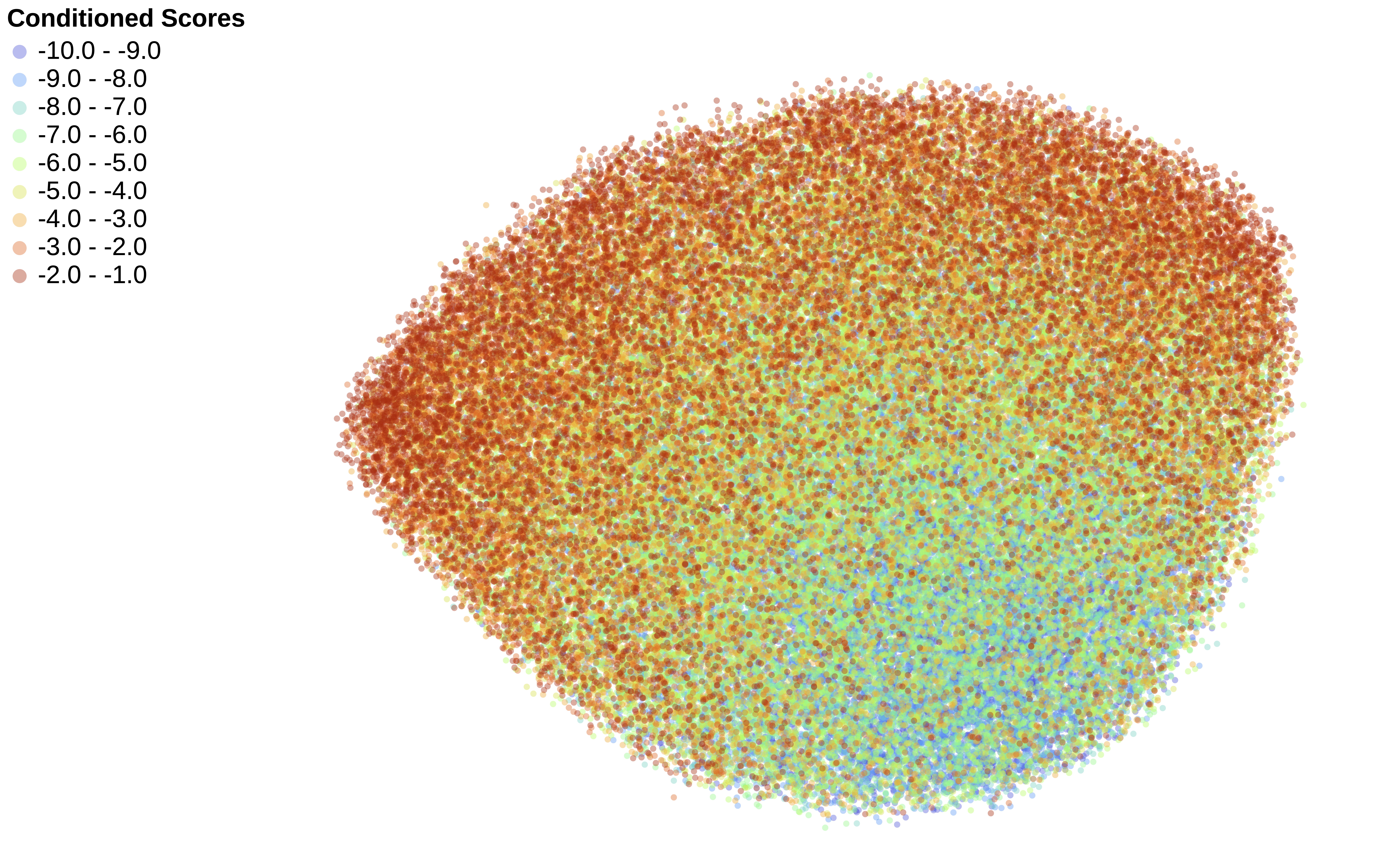}
        \caption{1SYH}
        \label{fig:latent-umap-1syh}
    \end{subfigure}

    \vspace{0.5em}

    \begin{subfigure}[t]{0.48\linewidth}
        \centering
        \includegraphics[alt={UMAP scatterplot of the Conditional VAE latent space for protein target 4LDE (beta-2 adrenoceptor); molecules color-coded from dark blue (weak binding) to yellow (strong binding), showing property-score-driven spatial clustering with smooth transitions at cluster boundaries.}, width=\linewidth]{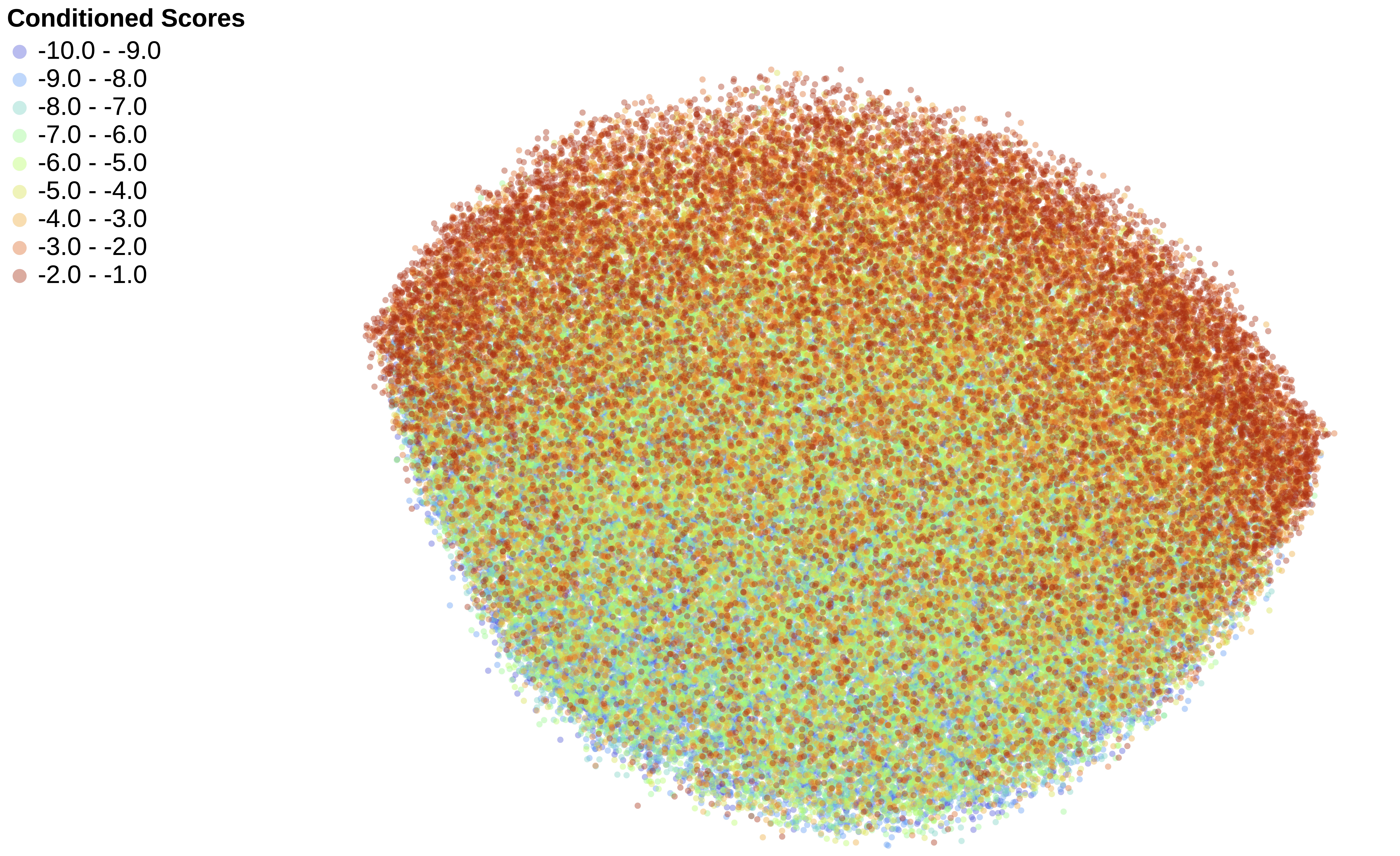}
        \caption{4LDE}
        \label{fig:latent-umap-4lde}
    \end{subfigure}
    \hfill
    \begin{subfigure}[t]{0.48\linewidth}
        \centering
        \includegraphics[alt={UMAP scatterplot of the Conditional VAE latent space for protein target 6Y2F (SARS-CoV-2 main protease); molecules color-coded from dark blue (weak binding) to yellow (strong binding), showing property-score-driven spatial clustering consistent with 1SYH and 4LDE panels.}, width=\linewidth]{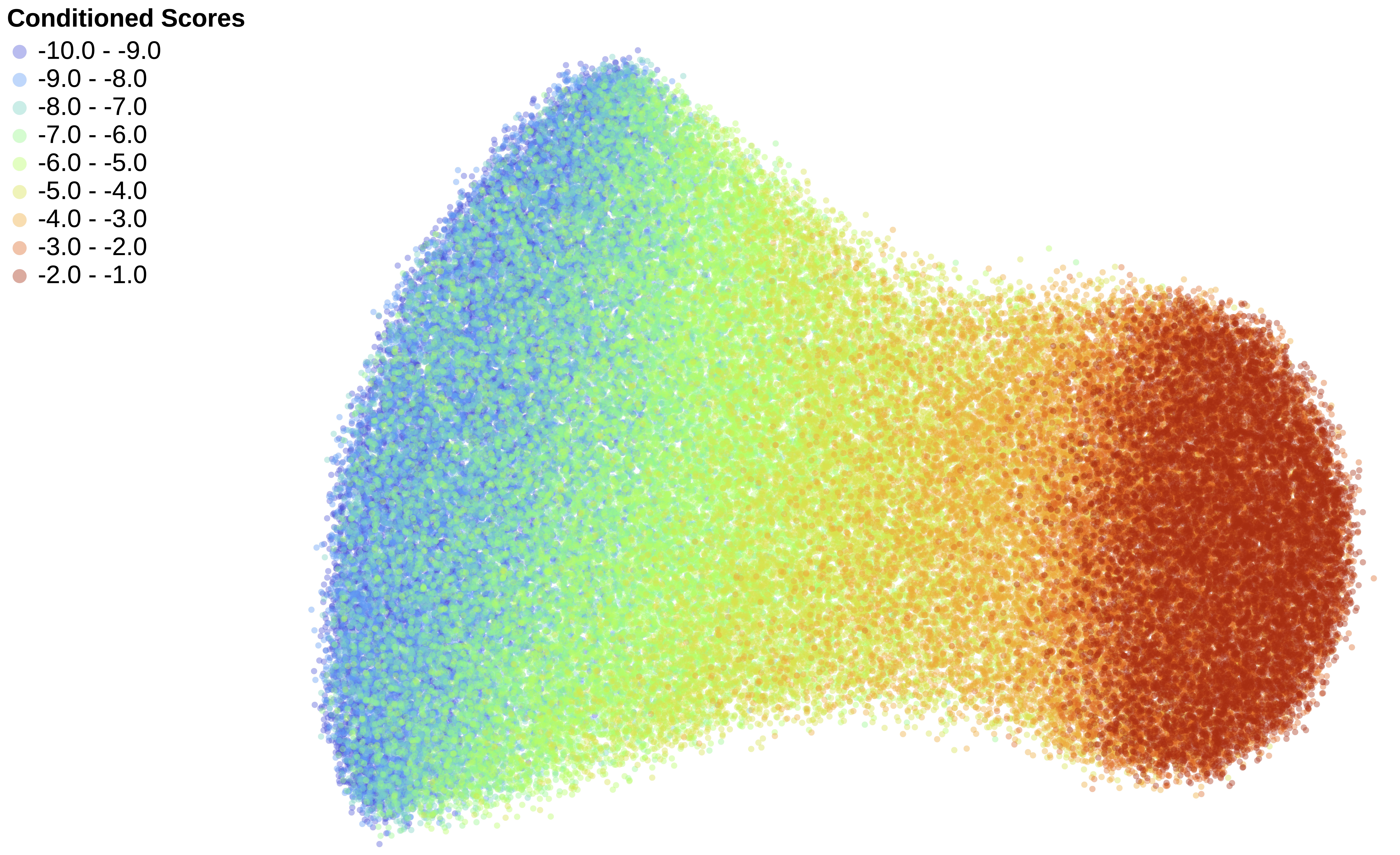}
        \caption{6Y2F}
        \label{fig:latent-umap-6y2f}
    \end{subfigure}

    \caption{UMAP projections of the CVAE conditional prior colored by property score. \textbf{(a)}~SA score: molecules colored by SA score bin, from low SA score (easier to synthesize) to high SA score (harder to synthesize); adjacent bins overlap continuously. \textbf{(b--d)}~Docking score for targets 1SYH, 4LDE, and 6Y2F: molecules colored from dark blue (docking score near 0, weaker predicted binding) to yellow (most negative docking score, strongest predicted binding). In all four panels, property-score levels occupy visibly different latent regions with smooth overlap at boundaries, showing score-associated organization in the two-dimensional projection. UMAP is a nonlinear embedding and is used here for visualization only; the quantitative evidence for conditioning is the rank correlation and calibration analysis in Table~\ref{tab:conditioning-summary}, not this projection.}
    \begin{flushleft}\small Alt text: Four Uniform Manifold Approximation and Projection scatterplots of the latent space. Panel a: synthetic accessibility score conditioning; molecules colored from low SA score (easier to synthesize) to high SA score (harder to synthesize), with smooth overlap between adjacent bins. Panel b: target 1SYH (GluR2) docking score; dark blue indicates weak binding near 0 kcal/mol, yellow indicates strong binding near -10 kcal/mol. Panel c: target 4LDE (beta-2 adrenoceptor) docking score; analogous color-coded clustering. Panel d: target 6Y2F (SARS-CoV-2 main protease) docking score; analogous clustering. All panels show property-level separation with smooth boundary overlap.\end{flushleft}
    \label{fig:latent-analysis-conditional}
\end{figure}
\newcommand{\cvaetartaruslabel}{fig:latent-analysis-conditional}
 
\old{Figure~\ref{fig:latent-analysis-conditional} visualizes the UMAP distribution of the molecules generated with CVAE conditioned with labels, (a) when guided by the synthetic accessibility (SA) score, computed using the RDKit \texttt{sascore} package~\cite{ertl2009estimation}, and (b) blood-brain barrier permeability (BBBP), computed with a predictive model trained on a benchmark dataset. The generated molecules are binned into score intervals and embedded into two dimensions using UMAP applied to samples from the conditional prior. Figure~\ref{fig:latent-analysis-conditional}~(a) shows a clear progression: low-SA molecules, which are harder to synthesize, appear in blue and form a concentrated region, while higher-SA molecules, in red, spread across the opposite side of the space. Intermediate bins transition smoothly between these extremes, indicating that the model has learned a semantically organized latent manifold aligned with synthetic accessibility. Figure~\ref{fig:latent-analysis-conditional}~(b) shows analogous separation among molecules generated under different BBBP conditions.
In both plots the conditioning score varies smoothly across the latent space, with adjacent score bins overlapping. This continuity is useful for interpolation and controllable generation.
These plots show where conditioning works well and where adjacent bins blur, which suggests where controllability could be improved.}
\rev{We fine-tuned a CVAE on 9,990 PubChem molecules labeled with RDKit SA scores~\cite{ertl2009estimation}, normalized to $[0,1]$ via min-max over the observed range $[1.44, 7.51]$ ($y\!=\!0$: easiest synthesis, $y\!=\!1$: hardest).
In this final adaptation stage, we updated only the FiLM module, conditional-prior mean and log-variance networks, latent-to-hidden projection, and label projection (726{,}016 of 91{,}047{,}752 parameters, 0.80\%). The Transformer encoder and decoder, their LoRA adapters, and the posterior heads remained frozen. LoRA adapters were trained in a preceding adaptation stage. Table~\ref{tab:experiment-configurations} gives the full configuration.
Figure~\ref{fig:latent-analysis-conditional}(a) shows a UMAP of the conditional prior colored by SA score bin: low-SA (easy) and high-SA (hard) molecules occupy distinct regions with smooth overlap at the boundary.}

\paragraph{\rev{Docking-score conditioning.}}

\old{We used the protein-ligand design portion of the Tartarus benchmark~\cite{nigam_2023_tartarus}, which reports docking scores against three targets: 1SYH (GluR2/GluA2 ligand-binding core), 4LDE ($\beta_2$ adrenoceptor), and 6Y2F (SARS-CoV-2 main protease). Docking scores (QVina) proxy binding affinity; lower is stronger.}
\rev{We used the protein-ligand design subset of the Tartarus benchmark~\cite{nigam_2023_tartarus}, which provides QVina docking scores against three targets: 1SYH (GluR2/GluA2 ligand-binding core), 4LDE ($\beta_2$ adrenoceptor), and 6Y2F (SARS-CoV-2 main protease).
Lower docking scores indicate stronger predicted binding.}

\rev{The Tartarus training corpus contains 105,338 molecules with docking scores for all three targets.
For 6Y2F all molecules dock with negative scores, so min-max normalization over the full set spans $[0,1]$ without compression.
For 1SYH and 4LDE, however, a small fraction of molecules yield large positive docking scores (up to $+255.6$ and $+18.6$\,kcal/mol respectively), compressing the labels of the binding-relevant negative-score molecules into only $[0,\,0.037]$ and $[0,\,0.346]$ of the unit interval.
We corrected this by re-normalizing within the negative-score subset for each target, mapping $[s_{\min},\,0]$ linearly to $[0,\,1]$ ($s_{\min}$: most negative docking score in training set) and discarding positive-score molecules from \emph{training-label construction}. At evaluation, by contrast, a generated molecule is treated as a completed docking calculation whenever QVina returns a non-sentinel score. Real non-negative scores are retained as unfavorable outcomes. This distinction avoids labeling an unfavorable but completed docking score as a computational failure.
This yielded 89,086 training molecules for 1SYH, 104,988 for 4LDE, and 105,338 for 6Y2F.
The label convention is $y\!=\!0$ for the strongest predicted binder ($s\!=\!s_{\min}$) and $y\!=\!1$ for the weakest ($s\!\to\!0$).
As for SA score, only the conditioning pathway (FiLM, conditional-prior networks, latent-to-hidden projection, and label projection) was updated in the final adaptation stage, with the Transformer weights, LoRA adapters, and posterior heads frozen (Table~\ref{tab:experiment-configurations}).}

\old{Three CVAEs were fine-tuned on the Tartarus training set (100,000 molecules), one per target; UMAP projections show clear latent separation by docking-score range (Figure~\ref{fig:latent-analysis-conditional}(b--d)). We generated 1,000 molecules per target per approach (VAE and CVAE), using $y^{*} = -10.0$ as the target condition, and docked all against their respective targets with \texttt{QVina}. A two-sample $t$-test found significant differences for two of three targets: 1SYH ($\mu_{\text{CVAE}} = -6.3$ vs.\ $\mu_{\text{VAE}} = -5.0$; $t = 21.5$, $p < 0.0001$) and 6Y2F ($\mu_{\text{CVAE}} = -5.4$ vs.\ $\mu_{\text{VAE}} = -5.2$; $t = 4.8$, $p < 0.0001$). For 1SYH and 6Y2F, the CVAE shifts docking scores toward stronger predicted binding than the VAE baseline. For the 4LDE target, the CVAE also generated molecules with strong docking scores, as reflected in the left-hand tail of the distribution. Instead of using the Tartarus single-best-molecule objective, we aim to shift the entire docking score distribution toward the conditioned value, using the VAE as a baseline. Shifting the full distribution produces many high-scoring, diverse candidates rather than a single optimized outlier.}
\rev{We generated 3,003 molecules per target per guidance scale, sweeping the requested label from 0 to 1 in steps of 0.001 and docking all valid molecules with QVina. The label direction is defined below. Note that the network-side label $y_{\text{model}}$ and the reported sweep coordinate $y^{*}$ run in opposite directions.
RDKit SA scores were computed for the valid set at each $\alpha$: mean SA ranged from 2.9 to 3.6 across all three targets and all guidance scales. SA score is a heuristic estimate rather than a measurement of synthesizability, and a mean can conceal a problematic tail, so we report these as favorable heuristic SA scores rather than as confirmation of synthetic accessibility. They do indicate that docking-score conditioning did not shift the bulk of the distribution toward the high-SA range~\cite{ertl2009estimation}.}

\begin{revblock}
\begin{figure}[H]
    \centering
    \includegraphics[alt={Grid of 28 scatter plots (4 rows by 7 columns). Rows show docking scores for protein targets 1SYH, 4LDE, and 6Y2F, and synthetic accessibility score. Columns show label-reference guidance strengths alpha from 1.0 to 12.0. Each panel plots conditioning label (x-axis, 0 to 1) versus measured score with a regression line and annotated Spearman correlation and slope. The pattern is target-dependent: synthetic accessibility and 4LDE strengthen with increasing alpha, whereas 6Y2F peaks at moderate alpha and then weakens, and 1SYH is approximately flat before rising at the highest alpha.}, width=\linewidth]{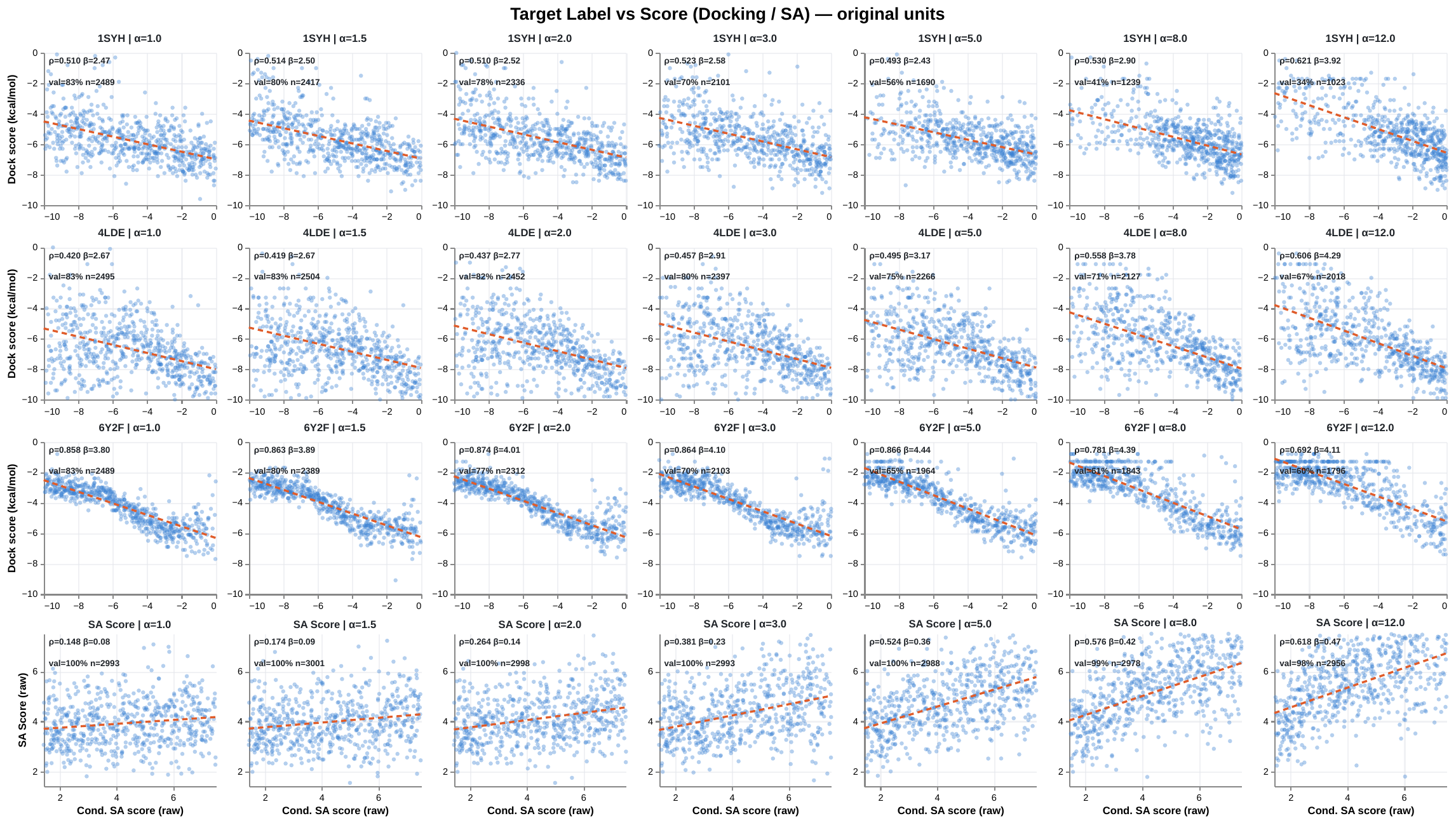}
    \caption{\revfloat{Conditioning label versus measured score for all four property targets across seven label-reference guidance strengths ($\alpha$). Rows~1--3 show docking score (kcal/mol) against conditioning label for targets 1SYH, 4LDE, and 6Y2F. The label axis is the network-side label $y_{\text{model}}$, for which $y_{\text{model}} = 0$ targets strong binding (most negative). Row~4 shows SA score (raw units, 1--7) against conditioning label. The value $y^{*} = 0$ targets easy synthesis (low SA score). Each panel shows a random subsample of up to 600 of the scored molecules for legibility (fixed seed), while the least-squares regression line and the annotated Spearman $\rho$ and calibration slope $\beta$ are computed on all scored molecules for that condition. The dependence on $\alpha$ is target-specific rather than uniformly increasing. SA score and 4LDE strengthen with $\alpha$, 6Y2F is strongest at moderate $\alpha$ and declines thereafter, and 1SYH is nearly flat until the largest $\alpha$. Validity also falls as $\alpha$ increases for the docking targets, so panels at high $\alpha$ summarize fewer molecules (Table~\ref{tab:conditioning-summary}).}}
    \begin{flushleft}\small Alt text: Grid of 28 scatter plots (4 rows, 7 columns). Rows correspond to docking score targets 1SYH (GluR2), 4LDE (beta-2 adrenoceptor), and 6Y2F (SARS-CoV-2 protease), plus synthetic accessibility score. Columns span label-reference guidance strength alpha from 1.0 to 12.0. Each panel shows conditioning label versus measured score with a regression line. The association is target-dependent: it strengthens with alpha for synthetic accessibility and 4LDE, peaks at moderate alpha for 6Y2F, and is flat then rising for 1SYH.\end{flushleft}
    \label{fig:combined-correlation}
\end{figure}
 \end{revblock}

\paragraph{\rev{Target Condition Alignment.}}

\rev{Figure~\ref{fig:combined-correlation} shows $y^{*}$ against the measured score for all four properties across seven guidance scales ($\alpha \in \{1.0,\allowbreak\ 1.5,\allowbreak\ 2.0,\allowbreak\ 3.0,\allowbreak\ 5.0,\allowbreak\ 8.0,\allowbreak\ 12.0\}$), with regression slope and Spearman $\rho$ annotated per panel. Table~\ref{tab:conditioning-summary} reports $\rho$, calibration slope $\beta$, validity, SR, and IntDiv at all tested $\alpha$ for each target.}

\rev{For SA score, $\rho$ increases from 0.15 at $\alpha\!=\!1$ to 0.62 at $\alpha\!=\!12$, with molecular validity remaining above 98\% across all tested $\alpha$ (Table~\ref{tab:conditioning-summary}, rows 1--7).
At $\alpha\!=\!12$, molecules conditioned on $y^{*}\!\in\![0,\,0.33]$ have a mean normalized SA score of 0.53 versus 0.84 for $y^{*}\!\in\![0.67,\,1]$, a separation of $\Delta\!=\!0.31$ (${\approx}1.9$ raw SA-score units).}

\rev{The docking labels require care, because two label coordinates appear in the pipeline. The label supplied to the network is $y_{\text{model}}$, for which $y_{\text{model}}\!=\!0$ denotes the strongest (most negative) training score. The saved generation files also contain sweep coordinate $y^{*}=1-y_{\text{model}}$. To remove the prior sign ambiguity, Table~\ref{tab:conditioning-summary} and Figure~\ref{fig:combined-correlation} report $y_{\text{model}}$ throughout. Rank correlations and calibration slopes are computed between $y_{\text{model}}$ and raw docking score. Positive values therefore denote increasing score as the requested condition moves from strong toward weak binding.
The strength of this association is target-dependent and not monotone in $\alpha$. Under the standardized sentinel-only population, 6Y2F reaches $\rho\!=\!0.813$ at $\alpha\!=\!1$, peaks at $0.826$ at $\alpha\!=\!2$, and declines to $0.647$ at $\alpha\!=\!12$. 1SYH remains near $0.52$ from $\alpha\!=\!1$ to $\alpha\!=\!5$ before rising to $0.630$ at $\alpha\!=\!12$. 4LDE increases from $0.482$ at $\alpha\!=\!1$ to $0.635$ at $\alpha\!=\!12$. Increasing $\alpha$ also reduces generation validity and docking completion for every protein target (Table~\ref{tab:conditioning-summary}), so larger $\alpha$ is not uniformly preferable. UMAP projections show latent-space organization by docking-score range (Figure~\ref{fig:latent-analysis-conditional}(b--d)).}

\rev{Success rate (SR) is the fraction of successfully scored molecules whose externally recomputed score falls within $\varepsilon$ of the target score implied by the reported conditioning coordinate, at $\varepsilon_{10}\!=\!0.1$ and $\varepsilon_{20}\!=\!0.2$ times the property range. The score is recomputed with the same deterministic procedure that produced the training labels (RDKit SA score, or QVina against the same receptor), so it is independent of the model's parameters but not an orthogonal predictor of the same property. For docking, one operational rule is now used throughout: only the explicit QVina failure sentinel denotes a failed docking job. Real non-negative scores are retained. Gen.Valid is the RDKit-parseable fraction of all 3,003 attempts, Dock is the fraction receiving a non-sentinel QVina score, and $\rho$, $\beta$, SR, and IntDiv are computed over that docked population. Table~\ref{tab:docking-pareto} separately reports attempt-level SR, where every invalid or failed-docking attempt counts as failure. For docking, the target score is $\hat{s}(y_{\text{model}})=s_{\min}(1-y_{\text{model}})$, so $\rho$, $\beta$, and SR use the same coordinate and population. For SA score at $\alpha\!=\!12$, SR$_{10}$/SR$_{20}$ are 0.214/0.411. At the rank-correlation-maximizing docking settings, docked-only SR$_{10}$/SR$_{20}$ are 0.449/0.725 (6Y2F, $\alpha=2$), 0.189/0.463 (1SYH, $\alpha=12$), and 0.366/0.657 (4LDE, $\alpha=12$). The corresponding attempt-level SR$_{20}$ values are 0.563, 0.161, and 0.404. Internal diversity (IntDiv) computed as $1 - \bar{\tau}$ over Morgan fingerprints (radius\,=\,2, 1024\,bits) is reported for both the full scored set and the SR$_{20}$ subset. Calibration curves (Figure~\ref{fig:controllability-calibration}) visualize the bin-level trend.}

\begingroup\footnotesize\setlength{\tabcolsep}{3pt}%
\begin{longtable}{llrrrrrrrr}
\caption{\revfloat{Conditioning performance across all label-reference guidance scales $\alpha$ for each property target. Bold rows mark the $\alpha$ with the highest $\rho$ per target. Because $\rho$ is not the only relevant objective, the full sweep is reported and Table~\ref{tab:docking-pareto} gives an explicit trade-off view. $\rho$: Spearman rank correlation between the reported conditioning coordinate and the recomputed score. Docking rows use the network-side coordinate $y_{\text{model}}=1-y^{*}$. $\beta$: calibration slope. Gen.Valid: RDKit-parseable molecules divided by 3,003 attempts. Dock: molecules with a non-sentinel QVina score divided by 3,003. Not applicable to SA score. SR$_{10}$/SR$_{20}$: valid-/docked-only success at $\varepsilon = 0.1/0.2 \times$ property range. IntDiv: internal diversity of the full scored set and the SR$_{20}$ subset. Docking rows use the sentinel-only failure rule.}}
\label{tab:conditioning-summary} \\
\toprule
\textbf{Target} & \textbf{$\alpha$} & \textbf{$\rho$} & \textbf{$\beta$} & \textbf{Gen.Valid} & \textbf{Dock} & \textbf{SR$_{10}$} & \textbf{SR$_{20}$} & \textbf{IntDiv} & \textbf{IntDiv\,SR$_{20}$} \\
\midrule
\endfirsthead
\multicolumn{10}{l}{\small\textit{Table~\ref{tab:conditioning-summary} continued from previous page}}\\[2pt]
\toprule
\textbf{Target} & \textbf{$\alpha$} & \textbf{$\rho$} & \textbf{$\beta$} & \textbf{Gen.Valid} & \textbf{Dock} & \textbf{SR$_{10}$} & \textbf{SR$_{20}$} & \textbf{IntDiv} & \textbf{IntDiv\,SR$_{20}$} \\
\midrule
\endhead
\multicolumn{10}{r}{\small\textit{Continued on next page}}\\
\endfoot
\bottomrule
\endlastfoot
 & 1.0 & 0.148 & 0.077 & 99.7\% & -- & 0.206 & 0.423 & 0.901 & 0.898 \\
 & 1.5 & 0.174 & 0.095 & 99.9\% & -- & 0.219 & 0.435 & 0.902 & 0.901 \\
 & 2.0 & 0.264 & 0.145 & 99.8\% & -- & 0.230 & 0.458 & 0.903 & 0.902 \\
SA score & 3.0 & 0.381 & 0.230 & 99.7\% & -- & 0.247 & 0.498 & 0.907 & 0.905 \\
 & 5.0 & 0.524 & 0.360 & 99.5\% & -- & 0.288 & 0.552 & 0.915 & 0.914 \\
 & 8.0 & 0.576 & 0.423 & 99.2\% & -- & 0.245 & 0.482 & 0.925 & 0.927 \\
 & \textbf{12.0} & \textbf{0.618} & \textbf{0.473} & \textbf{98.4\%} & -- & \textbf{0.214} & \textbf{0.411} & \textbf{0.934} & \textbf{0.939} \\
\midrule
 & 1.0 & 0.813 & 2.611 & 85.8\% & 83.6\% & 0.449 & 0.708 & 0.919 & 0.919 \\
 & 1.5 & 0.802 & 1.771 & 83.6\% & 80.6\% & 0.448 & 0.711 & 0.919 & 0.919 \\
 & \textbf{2.0} & \textbf{0.826} & \textbf{1.929} & \textbf{81.9\%} & \textbf{77.7\%} & \textbf{0.449} & \textbf{0.725} & \textbf{0.920} & \textbf{0.921} \\
6Y2F & 3.0 & 0.800 & 2.092 & 77.7\% & 71.0\% & 0.449 & 0.727 & 0.921 & 0.921 \\
 & 5.0 & 0.821 & 2.940 & 75.2\% & 66.2\% & 0.416 & 0.730 & 0.926 & 0.924 \\
 & 8.0 & 0.726 & 2.513 & 72.9\% & 62.1\% & 0.340 & 0.665 & 0.931 & 0.921 \\
 & 12.0 & 0.647 & 2.516 & 72.9\% & 60.4\% & 0.309 & 0.596 & 0.931 & 0.911 \\
\midrule
 & 1.0 & 0.523 & 2.972 & 86.2\% & 84.2\% & 0.270 & 0.522 & 0.902 & 0.905 \\
 & 1.5 & 0.528 & 3.097 & 84.0\% & 81.9\% & 0.278 & 0.536 & 0.901 & 0.905 \\
 & 2.0 & 0.526 & 3.122 & 81.2\% & 79.2\% & 0.270 & 0.538 & 0.901 & 0.903 \\
1SYH & 3.0 & 0.540 & 3.398 & 74.7\% & 71.4\% & 0.280 & 0.562 & 0.902 & 0.904 \\
 & 5.0 & 0.516 & 3.501 & 63.4\% & 57.5\% & 0.284 & 0.574 & 0.904 & 0.905 \\
 & 8.0 & 0.544 & 3.962 & 54.2\% & 41.9\% & 0.253 & 0.553 & 0.912 & 0.913 \\
 & \textbf{12.0} & \textbf{0.630} & \textbf{5.203} & \textbf{54.8\%} & \textbf{34.7\%} & \textbf{0.189} & \textbf{0.463} & \textbf{0.927} & \textbf{0.932} \\
\midrule
 & 1.0 & 0.482 & 9.927 & 89.2\% & 87.4\% & 0.322 & 0.559 & 0.915 & 0.917 \\
 & 1.5 & 0.482 & 7.507 & 89.6\% & 87.9\% & 0.321 & 0.557 & 0.914 & 0.917 \\
 & 2.0 & 0.498 & 11.226 & 88.6\% & 86.1\% & 0.331 & 0.567 & 0.914 & 0.917 \\
4LDE & 3.0 & 0.525 & 14.356 & 87.7\% & 85.0\% & 0.335 & 0.573 & 0.914 & 0.919 \\
 & 5.0 & 0.563 & 12.848 & 85.0\% & 81.2\% & 0.350 & 0.590 & 0.914 & 0.919 \\
 & 8.0 & 0.616 & 11.188 & 82.0\% & 75.8\% & 0.358 & 0.622 & 0.917 & 0.922 \\
 & \textbf{12.0} & \textbf{0.635} & \textbf{8.744} & \textbf{80.6\%} & \textbf{70.2\%} & \textbf{0.366} & \textbf{0.657} & \textbf{0.925} & \textbf{0.928} \\
 \end{longtable}
\endgroup

\rev{Because rank correlation alone does not capture the cost of stronger guidance,
Table~\ref{tab:docking-pareto} reports the same sentinel-only sweep at attempt level, adding the number of
molecules that were actually generated and docked, the attempt-level success rate, and the
fraction of attempts reaching a target-specific strong-binding cutoff. Two patterns are visible
there that the valid-only view obscures. First, the number of successfully docked molecules falls
substantially with $\alpha$: for 1SYH, from 2{,}530 of 3{,}003 attempts at $\alpha\!=\!1$ to 1{,}041 at
$\alpha\!=\!12$. Second, and more consequentially for practical use, the fraction of attempts
producing a strong binder \emph{decreases} monotonically with $\alpha$ for all three targets
(1SYH: 9.0\% to 2.7\%, 4LDE: 9.7\% to 3.7\%, and 6Y2F: 9.7\% to 3.0\%). Increasing guidance therefore
improves the rank ordering of scores while reducing the absolute yield of strong-binding molecules
per attempt. We consequently do not designate a single recommended $\alpha$ as uniformly best, and
report the full sweep instead.}

\begin{table}[H]
\centering
\caption{\revfloat{\rev{Attempt-level accounting and trade-off analysis for docking-score conditional generation. Every condition begins from the same 3,003 generation attempts (1,001 requested labels $\times$ 3 samples). $n_{\text{gen}}$: RDKit-parseable molecules produced. $n_{\text{dock}}$: molecules receiving a usable QVina score. Only the explicit failure sentinel is excluded, and real non-negative scores are retained. Valid.: $n_{\text{gen}}/3003$. Dock: $n_{\text{dock}}/3003$. $\rho$: Spearman correlation between network-side conditioning coordinate $y_{\text{model}}=1-y^{*}$ and raw docking score over the same docked population used in Table~\ref{tab:conditioning-summary}. SR$^{\text{att}}_{20}$: success rate at $\varepsilon=0.2\times$ property range with all 3,003 attempts in the denominator. Strong: fraction of \emph{attempts} yielding a docking score at or below the target-specific strong-binding cutoff (the 10th percentile of the $\alpha\!=\!1$ docked distribution: $-7.4$, $-8.9$, and $-6.0$\,kcal/mol for 1SYH, 4LDE, and 6Y2F). Increasing $\alpha$ raises $\rho$ for 1SYH and 4LDE and reduces strong-binder yield for all three targets. Attempt-level success also declines overall but is not strictly monotone. No single $\alpha$ is optimal across these measures.}}}
\label{tab:docking-pareto}
\smallskip
\footnotesize
\setlength{\tabcolsep}{4pt}
\begin{tabular}{llrrrrrrr}
\toprule
\textbf{Target} & \textbf{$\alpha$} & \textbf{$n_{\text{gen}}$} & \textbf{$n_{\text{dock}}$} & \textbf{Valid.} & \textbf{Dock} & \textbf{$\rho$} & \textbf{SR$^{\text{att}}_{20}$} & \textbf{Strong} \\
\midrule
1SYH & 1.0 & 2588 & 2530 & 0.862 & 0.842 & 0.523 & 0.440 & 0.0896 \\
 & 1.5 & 2522 & 2460 & 0.840 & 0.819 & 0.528 & 0.439 & 0.0886 \\
 & 2.0 & 2439 & 2377 & 0.812 & 0.792 & 0.526 & 0.426 & 0.0759 \\
 & 3.0 & 2243 & 2145 & 0.747 & 0.714 & 0.540 & 0.402 & 0.0703 \\
 & 5.0 & 1904 & 1727 & 0.634 & 0.575 & 0.516 & 0.330 & 0.0430 \\
 & 8.0 & 1627 & 1258 & 0.542 & 0.419 & 0.544 & 0.232 & 0.0383 \\
 & 12.0 & 1646 & 1041 & 0.548 & 0.347 & 0.630 & 0.161 & 0.0270 \\
\midrule
4LDE & 1.0 & 2680 & 2625 & 0.892 & 0.874 & 0.482 & 0.489 & 0.0966 \\
 & 1.5 & 2692 & 2639 & 0.896 & 0.879 & 0.482 & 0.489 & 0.0876 \\
 & 2.0 & 2660 & 2586 & 0.886 & 0.861 & 0.498 & 0.488 & 0.0842 \\
 & 3.0 & 2635 & 2552 & 0.877 & 0.850 & 0.525 & 0.487 & 0.0733 \\
 & 5.0 & 2554 & 2437 & 0.850 & 0.812 & 0.563 & 0.479 & 0.0586 \\
 & 8.0 & 2461 & 2277 & 0.820 & 0.758 & 0.616 & 0.472 & 0.0476 \\
 & 12.0 & 2420 & 2109 & 0.806 & 0.702 & 0.635 & 0.461 & 0.0370 \\
\midrule
6Y2F & 1.0 & 2577 & 2512 & 0.858 & 0.836 & 0.813 & 0.592 & 0.0966 \\
 & 1.5 & 2512 & 2419 & 0.836 & 0.806 & 0.802 & 0.572 & 0.0836 \\
 & 2.0 & 2460 & 2334 & 0.819 & 0.777 & 0.826 & 0.563 & 0.0806 \\
 & 3.0 & 2333 & 2133 & 0.777 & 0.710 & 0.800 & 0.516 & 0.0703 \\
 & 5.0 & 2259 & 1988 & 0.752 & 0.662 & 0.821 & 0.484 & 0.0606 \\
 & 8.0 & 2189 & 1866 & 0.729 & 0.621 & 0.726 & 0.413 & 0.0423 \\
 & 12.0 & 2188 & 1815 & 0.729 & 0.604 & 0.647 & 0.360 & 0.0303 \\
\bottomrule
\end{tabular}
\end{table}
 
\begin{revblock}
\begin{figure}[H]
    \centering
    \includegraphics[alt={Four calibration curve panels showing conditioning label (x-axis, 0 to 1) versus mean normalized measured property score (y-axis, 0 to 1) across 10 equal bins. A dashed diagonal marks perfect calibration; blue shaded bands show plus-or-minus one standard deviation of the per-molecule scores within each bin. The SA panel uses the reported requested label; docking panels use the network-side label, where zero denotes stronger binding. All panels show a systematic monotone trend with partial departure from the ideal diagonal at extremes.}, width=\linewidth]{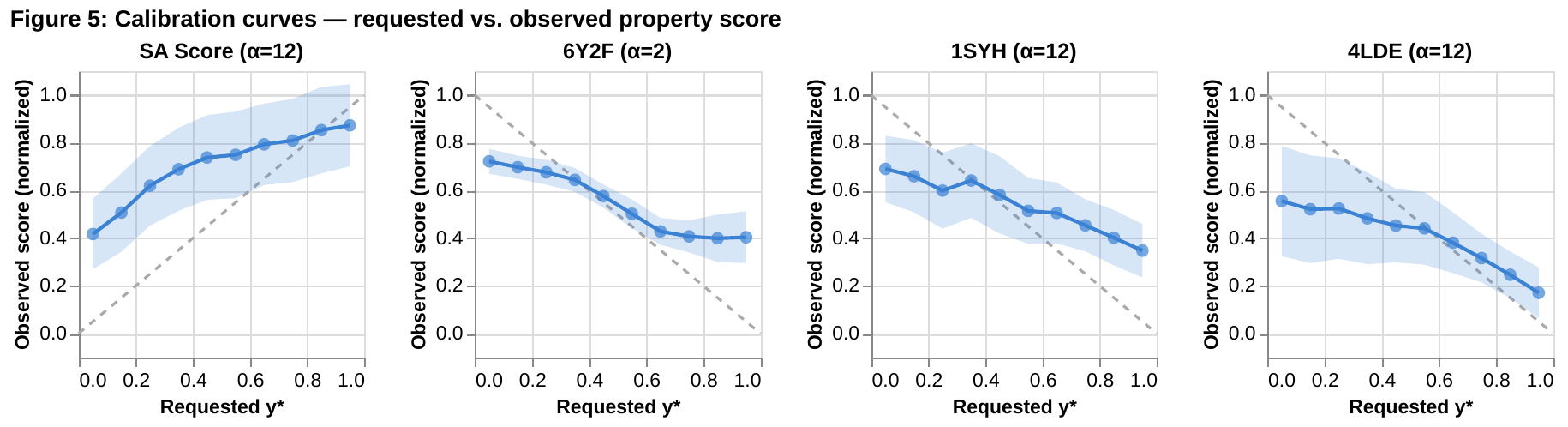}
    \caption{\revfloat{Calibration curves for all four conditioning targets at the $\alpha$ with the highest reported rank correlation.
    Each panel bins the successfully scored molecules from the 3,003 generation attempts into ten equal-width intervals and plots the mean $\pm$ 1 SD of the externally recomputed normalized property score in each bin (blue line and shaded band) against the target-specific ideal line (dashed grey). The SA panel uses requested label $y^{*}$. Docking panels use network-side label $y_{\text{model}}=1-y^{*}$. The shaded band shows the molecule-to-molecule spread within a bin, not the uncertainty of the binned mean, and the number of molecules per bin varies with validity.
    For docking targets (plotted against the network-side label $y_{\text{model}}$, for which $y_{\text{model}}\!=\!0$ is the strongest binder) the observed score is normalized to $[0,1]$ via $\hat{s} = (s - s_{\min})/(0 - s_{\min})$ so all four panels share the same axis.
    Deviation from the ideal line indicates residual miscalibration. Each panel indicates a monotone association between the requested label and the recomputed property, with the largest departures from the diagonal at the extremes of the label range.}}
    \begin{flushleft}\small Alt text: Four calibration curve panels. Each plots conditioning label (0 to 1) against mean normalized observed property score across 10 bins, with a dashed target-specific ideal line and shaded standard deviation band. Panels cover synthetic accessibility score and docking scores for targets 1SYH (GluR2), 4LDE (beta-2 adrenoceptor), and 6Y2F (SARS-CoV-2 main protease). All panels show monotone trends with the largest miscalibration at the extremes of the label range. Shaded bands show per-molecule spread within each bin, not uncertainty of the mean.\end{flushleft}
    \label{fig:controllability-calibration}
\end{figure}
 \end{revblock}

\paragraph{\rev{Scope and limitations.}}
\rev{
Each CVAE model reported above is trained independently for a single fixed protein target using docking scores as the scalar conditioning signal.
No cross-target transfer is performed or claimed: adapting to a new target uses the two-stage adaptation procedure, effective LoRA and FiLM adaptation followed by conditioning-pathway refinement with inherited LoRA frozen, on per-target docking data. Both stages together complete in under two hours on a single GPU using the pipeline described in the Model Architecture subsection.
This per-target training requirement is a current limitation that distinguishes \ModelName\ from approaches that explicitly represent target identity (such as protein-sequence or pocket-structure embeddings) as a conditioning variable, and is an open direction for future work.}

\paragraph{\rev{Baseline Comparison for SA score Conditioning.}}
\label{sec:sa-score-baseline-comparison-section}

\begin{revblock}

\paragraph{Baseline selection.}
We compare \ModelName\ SA score conditioning against three baselines representing distinct conditioning paradigms: reinforcement learning (REINVENT~4~\cite{loeffler2024reinvent4}), autoregressive prefix-token conditioning (MolGPT~\cite{bagal2021molgpt}), and latent diffusion with predictor guidance (DrugDiff~\cite{oestreich2025drugdiff}).
Together with \ModelName\ (latent-variable conditioning), these four models span the principal approaches to property-conditioned molecular generation.
MolGuidance~\cite{jin2025molguidance} and PropMolFlow~\cite{zeng2026propmolflow} were excluded because both require 3D SDF input and condition on quantum-chemistry properties not supported by the 2D SA score setting.

\paragraph{Baseline configurations.}
Each baseline is evaluated in its standard deployment configuration, which is the comparison relevant to practitioners choosing among frameworks.
\textbf{REINVENT~4} is first transfer-learned for five epochs on the same 9,990 PubChem SMILES to align its generation distribution with the PubChem-10K chemical space. It is then optimized through staged reinforcement learning with the RDKit SA score as the reward signal. This workflow is the standard deployment procedure for goal-directed generation with REINVENT~4.
\textbf{MolGPT} is trained from scratch on the same 9,990 molecules with SA score as the conditioning property; it encodes the target SA score value as a prefix token prepended to the SMILES sequence and has no pretrained backbone.
\textbf{DrugDiff} retains its published ZINC-250K pretrained VAE and latent diffusion model; an SA score prediction head (multi-layer perceptron, 3 hidden layers) is trained on ZINC-250K molecules scored with RDKit and used to steer reverse diffusion via predictor gradient guidance.

Performance differences therefore reflect both conditioning-mechanism design and initialization scale. Controlling for initialization by re-pretraining all models on PubChem-80M would require weeks of computation per baseline and produce non-standard configurations outside each method's intended use. The comparisons below are deployment-realistic, not architecturally controlled.

\paragraph{Evaluation protocol.}
All models generate exactly 3,003 molecules by sweeping the conditioning target $y^{*}$ uniformly from 0 to 1 in steps of 0.001 ($\times 3$ samples per level), matching the protocol used for \ModelName\ exactly.
Generated SMILES are rescored with the same deterministic RDKit SA-score procedure and normalized using the same scaler ($s_{\min}\!=\!1.437$, $s_{\max}\!=\!7.513$).
We report Spearman $\rho$ between $y^{*}$ and realized score, calibration slope $\beta$, MAE, success rate (SR$_{10}$/SR$_{20}$), molecular validity, and internal diversity (IntDiv). \rev{The $\rho$, $\beta$, and validity values are computed as in Table~\ref{tab:conditioning-summary}. IntDiv, however, was computed by three different pipelines that differ in fingerprint width and in which molecules enter the pairwise average, and we report the values as originally computed rather than silently re-deriving them. For \ModelName\ in this table the estimate uses Morgan radius 2 with 2048 bits over the first 2{,}000 molecules. For the three external baselines it uses Morgan radius 2 with 2048 bits over a random 1{,}000-molecule subsample. And in Table~\ref{tab:conditioning-summary} it uses Morgan radius 2 with 1024 bits over all valid molecules. The same \ModelName\ molecules at $\alpha\!=\!12$ therefore yield IntDiv $=0.928$ here and $0.934$ in Table~\ref{tab:conditioning-summary}, and would yield $0.946$ under the subsampling convention used for the baselines. IntDiv differences of this size ($\lesssim 0.02$) are consequently not meaningful in this table, and we do not draw conclusions from the IntDiv ranking.}
Because REINVENT~4 and DrugDiff do not natively support a fine-grained 1001-level $y^{*}$ sweep, we assign bucket midpoints ($y^{*}\!\in\!\{0.167, 0.500, 0.833\}$) to the low, mid, and high generation campaigns and sample 1,001 molecules per bucket; Spearman $\rho$ is computed over the resulting three-level ranking, which is structurally different from the 1,001-point sweep used for \ModelName\ and MolGPT (a distinction we address below).

\begin{table}[H]
\centering
\caption{\revfloat{SA score conditional generation benchmark. All models evaluated on 3,003 generation attempts using the same deterministic RDKit SA-score procedure and min--max scaler ($s_{\min}\!=\!1.437$, $s_{\max}\!=\!7.513$). Bold indicates the best value \emph{within each paradigm group}. The two groups use fundamentally different conditioning protocols and are not directly rank-comparable (see text). $\rho$: Spearman rank correlation between $y^{*}$ and realized score. $\beta$: calibration slope (OLS). MAE: mean $|y^{*}-\hat{y}|$. SR$_{10}$/SR$_{20}$: success rate at $\varepsilon\!=\!0.10/0.20$. Valid.: $n_{\text{valid}}/3003$, i.e.\ the denominator is the number of generation attempts for every model. IntDiv: internal diversity. Computed by three different pipelines that differ in fingerprint width and evaluated subset, so it is not bolded and is not comparable across rows or against Table~\ref{tab:conditioning-summary} (see text). Scaffolds: unique Murcko frameworks. Scaf/mol = unique scaffolds / $n_{\text{valid}}$. $|\Delta|$: $|$mean(high bucket) $-$ mean(low bucket)$|$ (larger = better separation). $^\ddagger$Spearman $\rho$ is a 3-point test. $^{*}$Metrics use the 50.4\% valid survivors only.}}
\label{tab:sa-score-baseline-comparison}
\scriptsize
\setlength{\tabcolsep}{2pt}
\begin{adjustbox}{max width=\linewidth}
\begin{tabular}{lrrrrrrrrrr}
\toprule
\textbf{Model} & \textbf{$\rho$} & \textbf{$\beta$} & \textbf{MAE} & \textbf{SR$_{10}$} & \textbf{SR$_{20}$} & \textbf{Valid.} & \textbf{IntDiv} & \textbf{Scaffolds} & \textbf{Scaf/mol} & \textbf{$|\Delta|$} \\
\midrule
\multicolumn{11}{l}{\textit{Paradigm A: oracle-free continuous conditioning (1,001 levels)}} \\
\textbf{\ModelName} (ours, $\alpha\!=\!12$) & 0.618 & 0.473 & 0.261 & 0.215 & 0.411 & \textbf{0.984} & 0.928 & \textbf{1,974} & 0.668 & 0.314 \\
MolGPT$^{*}$ & \textbf{0.895} & \textbf{0.703} & \textbf{0.092} & \textbf{0.685} & \textbf{0.890} & 0.505 & 0.907 & 1,088 & \textbf{0.718} & \textbf{0.428} \\
\midrule
\multicolumn{11}{l}{\textit{Paradigm B: oracle-in-the-loop / 3-bucket conditioning}} \\
REINVENT~4$^\ddagger$ & \textbf{0.941} & \textbf{1.482} & \textbf{0.145} & \textbf{0.228} & \textbf{0.910} & 0.875 & 0.918 & 1,347 & \textbf{0.513} & \textbf{0.971} \\
DrugDiff$^\ddagger$ & $-$0.026 & $-$0.013 & 0.273 & 0.207 & 0.390 & \textbf{1.000} & 0.922 & \textbf{1,373} & 0.457 & 0.008 \\
\bottomrule
\end{tabular}
\end{adjustbox}
\end{table}

\paragraph{Results.}
Table~\ref{tab:sa-score-baseline-comparison} summarizes all metrics. Because the two paradigms use different evaluation protocols, Spearman $\rho$ is only meaningful when compared within the same paradigm group.

\emph{Paradigm~A: oracle-free continuous conditioning} (\ModelName, MolGPT).
Both models are trained on static (label, SELFIES/SMILES) pairs, with no oracle access during training, and are evaluated across 1,001 uniformly-spaced conditioning targets $y^{*} \in [0,1]$.
In this setting, Spearman $\rho$ measures how well each model tracks the requested property value across the full conditioning range.
\ModelName\ generates valid molecules at $98.4\%$ of all attempts versus $50.4\%$ for MolGPT, and produces 1,974 distinct Murcko scaffolds~\cite{bemis1996murcko} from 2,956 scored molecules (scaf/mol $= 0.668$).
MolGPT's $\rho\!=\!0.895$ is based only on its 1,515 valid outputs and does not account for the 1,488 failed attempts ($49.6\%$ of all tries). Validity drops sharply as the conditioning target moves toward harder synthesis: $75.7\%$ at low $y^{*}$ ($y^{*}\!\leq\!0.33$, raw SA$\,\leq\,3.5$, easy-to-synthesize) versus $24.4\%$ at high $y^{*}$ ($y^{*}\!\geq\!0.67$, raw SA$\,\geq\,5.5$, hard-to-synthesize); Table~S2 shows the full per-bucket breakdown. This reflects a known failure mode of autoregressive SMILES decoders, which struggle to produce valid outputs for structurally complex molecules. The valid survivors skew toward easy-to-synthesize structures (mean SA$_\text{norm}\!=\!0.35$ vs.\ $\approx\!0.60$ for the training set), so the effective SR$_{10}$ across all 3,003 attempts falls from the reported $0.685$ to approximately $0.346$, close to \ModelName's $0.215$. On the same attempt basis, MolGPT's scaffold coverage rate is $1{,}088/3{,}003 = 0.362$, versus \ModelName's $0.668$.

\emph{Paradigm~B: oracle-in-the-loop / 3-bucket conditioning} (REINVENT~4, DrugDiff).
REINVENT~4 optimizes three separate RL policies, one per conditioning bucket ($y^{*}\!\in\!\{0.167, 0.500, 0.833\}$), using approximately 384,000 SA score oracle calls in total.
Its $\rho\!=\!0.941$ is a 3-point rank test: it simply checks whether the three bucket-level means are correctly ordered, which is expected when each policy was directly trained to hit its bucket. SR$_{10}\!=\!0.228$ shows that pointwise accuracy remains modest despite the high rank correlation. \rev{Its aggregate SR values are also computed on retained molecules only: REINVENT~4 retained 2,628 unique valid molecules from 3,003 requested ($87.5\%$), and retention is strongly bucket-dependent ($63.9\%$ at $y^{*}\!=\!0.167$ versus $99.1\%$ at $y^{*}\!=\!0.833$). Because its sampler discards duplicates as well as unparseable strings, the low-$y^{*}$ shortfall reflects repeated structures rather than invalid chemistry. On a per-requested-molecule basis its SR$_{20}$ in that campaign falls from $1.000$ to $0.639$ (Table~S2).} The calibration slope $\beta\!=\!1.482$ indicates that the model over-shoots the requested targets: the easy-synthesis campaign ($y^{*}\!=\!0.167$) produces molecules with a mean realized SA score of $0.003$ (target: $0.167$), and the hard-synthesis campaign ($y^{*}\!=\!0.833$) reaches a mean of $0.974$ (target: $0.833$); both are farther from their targets than their nominal distance suggests (Table~S2). Scaffold diversity is also highly uneven: the easy-synthesis campaign converges to only 61 unique Murcko scaffolds from the 640 unique molecules it retained out of 1,001 requested (scaf/mol $= 0.095$), while the hard-synthesis campaign yields near-unique scaffolds per molecule (979 unique from 992 valid; scaf/mol $= 0.987$), because the RL agent saturates the accessible scaffold pool when the reward target is easy (Table~S2).
DrugDiff generates valid molecules at $100\%$ but produces no conditioning effect on the PubChem distribution ($\rho\!\approx\!0$, $|\Delta|\!=\!0.008$) and the fewest distinct scaffolds (scaf/mol $= 0.457$). The SA score predictor trained on ZINC-250K molecules does not transfer to the PubChem chemical space without retraining.

\paragraph{Scope and limitations.}
The four baselines differ in pretraining scale and initialization, so observed performance gaps reflect both the conditioning mechanism and the starting distribution. A controlled comparison would require re-pretraining all models on PubChem-80M, which is outside the scope of this study.
 \end{revblock}
\subsection{Rediscovery Evaluation}

\rev{Rediscovery evaluation was performed across four ChEMBL target datasets. Each dataset contains fewer than 11,000 labeled compounds after curation, which is insufficient for direct generative fine-tuning. We therefore adopted a silver-label strategy. MMELON~\cite{suryanarayanan2024multi} (Multi-view Molecular Encoder with Long-context and Omics Networks), a multi-view molecular foundation model, was fine-tuned on each target dataset and applied to 100,000 randomly selected PubChem molecules to produce pseudo-labels for \ModelName\ fine-tuning. Pseudo-label quality was validated against held-out experimental ChEMBL values (Table~\ref{tab:results}). RMSE values of 0.095--0.114 log-units across targets indicate that the pseudo-labels are sufficiently accurate for use as conditioning signals. All independent property predictors used in controllability validation were trained exclusively on experimental labels, without access to the pseudo-labeled PubChem set.}

\begin{sloppypar}
Four targets were selected from ChEMBL for molecular rediscovery evaluation:
ABL1 kinase (CHEMBL1862), Bromodomain-containing protein~4 (BRD4, CHEMBL1163125),
Glucocorticoid receptor (GRL, CHEMBL2034), and HIV-1 protease (CHEMBL243).
Standard curation was applied: IC50-only assays, canonical SMILES, charge removal, duplicate averaging, and Murcko scaffold computation. After refinement, the ABL1 dataset was reduced from 18,830 to 2,064 samples, BRD4 from 58,432 to 10,800, GRL from 15,052 to 2,744, and HIV-1 protease from 9,226 to 2,843 samples.
\end{sloppypar}

MMELON was fine-tuned on each target dataset using an 80/10/10 train/validation/test split. Fine-tuning used the AdamW optimizer with a learning rate of $10^{-4}$, batch size 32, and gradient clipping at 1.0. Models were trained for up to 200 epochs with early stopping (patience 20) based on validation RMSE, and the best checkpoint was selected by minimum validation RMSE. Predictions on 100,000 randomly selected PubChem molecules provided silver labels for \ModelName\ fine-tuning. All models converged without overfitting (Table~\ref{tab:results}).

\begin{table}[H]
\centering
\caption{Evaluation metrics for ChEMBL molecular property prediction across four target datasets. Training completed successfully for all datasets. The evaluation metrics (MSE, RMSE, MAE) are reported. Sample counts reflect the 80/10/10 train/validation/test split.}
\label{tab:results}
\begin{tabular}{@{}lrrrcccccc@{}} \\
\toprule
\textbf{Dataset} & \multicolumn{3}{c}{\textbf{Samples}} & \textbf{MSE} & \textbf{RMSE} & \textbf{MAE} \\
\cmidrule(lr){2-4}
& \textbf{Train} & \textbf{Val} & \textbf{Test} & & & & & \\
\midrule
ABL1 & 1650 & 207 & 207 & 0.0090 & 0.0951 & 0.0695 \\
BRD4 & 8640 & 1080 & 1080 & 0.0050 & 0.0704 & 0.0442 \\
GRL & 2194 & 275 & 275 & 0.0120 & 0.1093 & 0.0707 \\
HIV-1 Protease & 2273 & 285 & 285 & 0.0130 & 0.1140 & 0.0845 \\
\bottomrule
\end{tabular}
\end{table}

Three generation approaches were compared: PubChem sampling (baseline), VAE (temperature $T=1.1$), and CVAE (LoRA with label conditioning). Each was run on the four ChEMBL targets at five sample sizes (10K to 50K), for 60 conditions total. \rev{%
\textbf{PubChem sampling} (baseline): molecules were drawn uniformly at random from the 100,000-molecule PubChem set used for silver-label generation, with no model involvement.
\textbf{VAE}: the pretrained VAE encodes each training-set seed molecule into its posterior mean $\mu$, samples latent vectors in the neighbourhood of $\mu$ (perturbation scale $\sigma$, decoder temperature $T=1.1$), and decodes to SMILES.
Seeds were drawn from the training set by cluster-proportional sampling to reflect the full scaffold distribution.
\textbf{CVAE}: the fine-tuned CVAE generates from a label alone, with no seed molecule. \rev{In this configuration the label-conditioned prior was disabled, so latent vectors are drawn from the isotropic Gaussian $\mathcal{N}(\mathbf{0},\mathbf{I})$ and modulated by FiLM with the requested label before decoding (Table~\ref{tab:experiment-configurations}). Conditioning therefore acts on the decoder rather than on the latent distribution.}
Conditioning targets $y^{*}$ were drawn proportionally from all ten label bins of the training distribution, covering the full range of the min-max-normalized $\log_{10}$(IC$_{50}$/nM) scale (label\,=\,0: strongest binder. Label\,=\,1: weakest binder).
This broad sampling was chosen to evaluate generation coverage. Conditioning on a targeted label tier is addressed in the Scope and Limitations paragraph below.} Metrics cover Exact Rediscovery (exact SMILES matches), Nearest Neighbor Similarity (Tanimoto, Morgan fingerprints radius\,=\,2, 2048\,bits), Scaffold Rediscovery (Bemis--Murcko), and Clustering Rediscovery (Butina, threshold 0.6).

The held-out split was constructed by Butina clustering with Morgan fingerprints (radius = 2, 2048 bits) and a similarity threshold of 0.6. Approximately 20\% of clusters ($\sim$20\% of molecules) were held out using stratified sampling across 10 score bins, so that entire clusters, rather than individual molecules, were reserved for evaluation. \rev{This removes close analogues from the target-specific fine-tuning set. It does not remove the held-out compounds from the pretraining corpus (Appendix~F).} The clustering proportion metric measures the fraction of held-out clusters reached by at least one generated molecule, evaluating whether models generate molecules in chemically feasible regions that were excluded from training. The number of molecules in the held-out sets is 364, 2028, 615, and 492 for ABL1, BRD4, GRL, and HIV-1 protease, respectively.

\newcommand{\grouprule}{\noalign{\vskip1pt}%
  \specialrule{0.4pt}{0pt}{0pt}%
  \noalign{\vskip1pt}%
}

\begingroup\scriptsize\setlength{\tabcolsep}{2pt}%
\begin{longtable}{@{}llrrrrrrrrrr@{}}
\caption{Comprehensive rediscovery metrics for all sample sizes (10K, 20K, 30K, 40K, 50K) across four datasets and three approaches. Within each 3-row group (CVAE, PUBCHEM, VAE at the same dataset and sample size), the highest value is bolded for Exact P$^{*}$, Exact R$^{*}$, NN Sim, Clust 0.6, Scaff Redisc, \old{CrossSim Ratio}\rev{CrossSim}, Div, and Uniq; for InterSim the lowest value is bolded. Exact P$^{*}$ and Exact R$^{*}$ are reported in $\times 10^{-4}$. \rev{CrossSim is the mean Tanimoto similarity over all generated--held-out scaffold pairs. It is a raw similarity in $[0,1]$, not a ratio. Appendix~A gives the definitions.}}
\label{tab:rediscovery_results} \\
\toprule
Dataset & Model & N & Exact P$^{*}$ & Exact R$^{*}$ & NN Sim & Clust 0.6 & Scaff Redisc & \old{CrossSim Ratio}\rev{CrossSim} & InterSim & Div & Uniq \\
\midrule
\endfirsthead
\multicolumn{12}{l}{\small\textit{Table~\ref{tab:rediscovery_results} continued from previous page}}\\[2pt]
\toprule
Dataset & Model & N & Exact P$^{*}$ & Exact R$^{*}$ & NN Sim & Clust 0.6 & Scaff Redisc & \old{CrossSim Ratio}\rev{CrossSim} & InterSim & Div & Uniq \\
\midrule
\endhead
\multicolumn{12}{r}{\small\textit{Continued on next page}}\\
\endfoot
\bottomrule
\endlastfoot
ABL1 & CVAE & 10000 & 0 & 0 & 0.180 & 0.000 & 0.032 & 0.110 & \textbf{0.104} & \textbf{1.096} & \textbf{1.014} \\
ABL1 & PUBCHEM & 10000 & 0 & 0 & 0.210 & 0.000 & 0.059 & 0.129 & 0.127 & 1.086 & \textbf{1.014} \\
ABL1 & VAE & 10000 & \textbf{1.16} & \textbf{27.47} & \textbf{0.246} & \textbf{0.073} & \textbf{0.100} & \textbf{0.133} & 0.120 & 1.078 & 0.954 \\
\grouprule
ABL1 & CVAE & 20000 & 0 & 0 & 0.181 & 0.000 & 0.046 & 0.110 & \textbf{0.104} & \textbf{1.096} & \textbf{1.014} \\
ABL1 & PUBCHEM & 20000 & 0 & 0 & 0.210 & 0.018 & 0.078 & 0.130 & 0.127 & 1.086 & \textbf{1.014} \\
ABL1 & VAE & 20000 & \textbf{1.77} & \textbf{82.42} & \textbf{0.246} & \textbf{0.101} & \textbf{0.110} & \textbf{0.133} & 0.120 & 1.078 & 0.937 \\
\grouprule
ABL1 & CVAE & 30000 & 0 & 0 & 0.181 & 0.000 & 0.046 & 0.110 & \textbf{0.104} & \textbf{1.096} & \textbf{1.014} \\
ABL1 & PUBCHEM & 30000 & 0 & 0 & 0.210 & 0.037 & 0.082 & 0.129 & 0.127 & 1.086 & \textbf{1.014} \\
ABL1 & VAE & 30000 & \textbf{1.98} & \textbf{137.36} & \textbf{0.245} & \textbf{0.138} & \textbf{0.128} & \textbf{0.132} & 0.120 & 1.078 & 0.928 \\
\grouprule
ABL1 & CVAE & 40000 & 0 & 0 & 0.181 & 0.000 & 0.050 & 0.110 & \textbf{0.104} & \textbf{1.096} & \textbf{1.014} \\
ABL1 & PUBCHEM & 40000 & 0 & 0 & 0.210 & 0.037 & 0.087 & 0.130 & 0.127 & 1.086 & \textbf{1.014} \\
ABL1 & VAE & 40000 & \textbf{1.8} & \textbf{164.84} & \textbf{0.246} & \textbf{0.128} & \textbf{0.160} & \textbf{0.132} & 0.120 & 1.078 & 0.917 \\
\grouprule
ABL1 & CVAE & 50000 & 0 & 0 & 0.181 & 0.000 & 0.055 & 0.110 & \textbf{0.104} & \textbf{1.096} & \textbf{1.014} \\
ABL1 & PUBCHEM & 50000 & 0 & 0 & 0.210 & 0.037 & 0.087 & 0.130 & 0.127 & 1.086 & \textbf{1.014} \\
ABL1 & VAE & 50000 & \textbf{1.93} & \textbf{219.78} & \textbf{0.246} & \textbf{0.138} & \textbf{0.174} & \textbf{0.132} & 0.120 & 1.078 & 0.912 \\
\midrule
BRD4 & CVAE & 10000 & 0 & 0 & 0.207 & 0.000 & 0.009 & 0.105 & \textbf{0.103} & \textbf{1.066} & 1.043 \\
BRD4 & PUBCHEM & 10000 & 0 & 0 & \textbf{0.240} & 0.009 & 0.015 & \textbf{0.119} & 0.127 & 1.056 & \textbf{1.043} \\
BRD4 & VAE & 10000 & \textbf{1.09} & \textbf{4.93} & 0.240 & \textbf{0.025} & \textbf{0.049} & 0.114 & 0.107 & 1.053 & 1.034 \\
\grouprule
BRD4 & CVAE & 20000 & 0 & 0 & 0.207 & 0.000 & 0.012 & 0.106 & \textbf{0.104} & \textbf{1.066} & 1.043 \\
BRD4 & PUBCHEM & 20000 & 0 & 0 & 0.240 & 0.012 & 0.020 & \textbf{0.119} & 0.127 & 1.056 & \textbf{1.043} \\
BRD4 & VAE & 20000 & \textbf{1.1} & \textbf{9.86} & \textbf{0.241} & \textbf{0.056} & \textbf{0.063} & 0.114 & 0.107 & 1.053 & 1.024 \\
\grouprule
BRD4 & CVAE & 30000 & 0 & 0 & 0.207 & 0.000 & 0.013 & 0.106 & \textbf{0.104} & \textbf{1.066} & 1.043 \\
BRD4 & PUBCHEM & 30000 & 0 & 0 & 0.240 & 0.012 & 0.027 & \textbf{0.119} & 0.127 & 1.056 & \textbf{1.043} \\
BRD4 & VAE & 30000 & \textbf{0.74} & \textbf{9.86} & \textbf{0.241} & \textbf{0.053} & \textbf{0.077} & 0.114 & 0.107 & 1.053 & 1.018 \\
\grouprule
BRD4 & CVAE & 40000 & 0 & 0 & 0.207 & 0.000 & 0.013 & 0.106 & \textbf{0.104} & \textbf{1.066} & 1.043 \\
BRD4 & PUBCHEM & 40000 & 0 & 0 & \textbf{0.240} & 0.012 & 0.031 & \textbf{0.120} & 0.127 & 1.056 & \textbf{1.043} \\
BRD4 & VAE & 40000 & \textbf{1.67} & \textbf{29.59} & 0.240 & \textbf{0.062} & \textbf{0.070} & 0.114 & 0.107 & 1.053 & 1.014 \\
\grouprule
BRD4 & CVAE & 50000 & 0 & 0 & 0.207 & 0.000 & 0.014 & 0.106 & \textbf{0.104} & \textbf{1.066} & 1.043 \\
BRD4 & PUBCHEM & 50000 & 0 & 0 & 0.240 & 0.019 & 0.033 & \textbf{0.120} & 0.127 & 1.056 & \textbf{1.043} \\
BRD4 & VAE & 50000 & \textbf{0.45} & \textbf{9.86} & \textbf{0.240} & \textbf{0.062} & \textbf{0.088} & 0.114 & 0.107 & 1.053 & 1.011 \\
\midrule
GRL & CVAE & 10000 & \textbf{0} & \textbf{0} & 0.203 & 0.000 & 0.064 & 0.106 & \textbf{0.104} & \textbf{1.036} & \textbf{1.025} \\
GRL & PUBCHEM & 10000 & \textbf{0} & \textbf{0} & 0.231 & 0.020 & 0.120 & \textbf{0.118} & 0.127 & 1.027 & \textbf{1.025} \\
GRL & VAE & 10000 & \textbf{0} & \textbf{0} & \textbf{0.242} & \textbf{0.030} & \textbf{0.127} & 0.110 & 0.111 & 1.036 & 0.906 \\
\grouprule
GRL & CVAE & 20000 & 0 & 0 & 0.203 & 0.000 & 0.090 & 0.107 & \textbf{0.105} & \textbf{1.036} & \textbf{1.025} \\
GRL & PUBCHEM & 20000 & 0 & 0 & 0.231 & \textbf{0.035} & \textbf{0.142} & \textbf{0.118} & 0.127 & 1.028 & \textbf{1.025} \\
GRL & VAE & 20000 & \textbf{2.38} & \textbf{48.78} & \textbf{0.241} & 0.025 & 0.139 & 0.110 & 0.110 & 1.034 & 0.872 \\
\grouprule
GRL & CVAE & 30000 & 0 & 0 & 0.203 & 0.000 & 0.105 & 0.107 & \textbf{0.105} & \textbf{1.036} & \textbf{1.025} \\
GRL & PUBCHEM & 30000 & 0 & 0 & 0.231 & \textbf{0.066} & \textbf{0.187} & \textbf{0.118} & 0.127 & 1.028 & \textbf{1.025} \\
GRL & VAE & 30000 & \textbf{1.63} & \textbf{48.78} & \textbf{0.242} & 0.051 & 0.157 & 0.110 & 0.111 & 1.035 & 0.851 \\
\grouprule
GRL & CVAE & 40000 & 0 & 0 & 0.203 & 0.000 & 0.109 & 0.107 & \textbf{0.105} & \textbf{1.036} & \textbf{1.025} \\
GRL & PUBCHEM & 40000 & 0 & 0 & 0.231 & \textbf{0.076} & \textbf{0.217} & \textbf{0.118} & 0.127 & 1.027 & \textbf{1.025} \\
GRL & VAE & 40000 & \textbf{1.65} & \textbf{65.04} & \textbf{0.241} & 0.045 & 0.184 & 0.110 & 0.111 & 1.034 & 0.840 \\
\grouprule
GRL & CVAE & 50000 & 0 & 0 & 0.203 & 0.000 & 0.112 & 0.107 & \textbf{0.105} & \textbf{1.036} & \textbf{1.025} \\
GRL & PUBCHEM & 50000 & 0 & 0 & 0.231 & \textbf{0.086} & \textbf{0.228} & \textbf{0.118} & 0.127 & 1.027 & \textbf{1.025} \\
GRL & VAE & 50000 & \textbf{1.34} & \textbf{65.04} & \textbf{0.242} & 0.051 & 0.180 & 0.110 & 0.110 & 1.034 & 0.828 \\
\midrule
HIV-1 Protease & CVAE & 10000 & \textbf{0} & \textbf{0} & 0.184 & \textbf{0.000} & 0.017 & 0.112 & \textbf{0.104} & \textbf{1.173} & \textbf{1.070} \\
HIV-1 Protease & PUBCHEM & 10000 & \textbf{0} & \textbf{0} & \textbf{0.208} & \textbf{0.000} & \textbf{0.021} & \textbf{0.133} & 0.127 & 1.163 & \textbf{1.070} \\
HIV-1 Protease & VAE & 10000 & \textbf{0} & \textbf{0} & 0.207 & \textbf{0.000} & \textbf{0.021} & 0.128 & 0.117 & 1.152 & 1.047 \\
\grouprule
HIV-1 Protease & CVAE & 20000 & \textbf{0} & \textbf{0} & 0.184 & 0.000 & 0.017 & 0.113 & \textbf{0.104} & \textbf{1.173} & \textbf{1.070} \\
HIV-1 Protease & PUBCHEM & 20000 & \textbf{0} & \textbf{0} & \textbf{0.208} & 0.000 & 0.026 & \textbf{0.133} & 0.127 & 1.163 & \textbf{1.070} \\
HIV-1 Protease & VAE & 20000 & \textbf{0} & \textbf{0} & 0.206 & \textbf{0.012} & \textbf{0.030} & 0.127 & 0.116 & 1.153 & 1.044 \\
\grouprule
HIV-1 Protease & CVAE & 30000 & \textbf{0} & \textbf{0} & 0.184 & 0.000 & 0.017 & 0.113 & \textbf{0.105} & \textbf{1.173} & \textbf{1.070} \\
HIV-1 Protease & PUBCHEM & 30000 & \textbf{0} & \textbf{0} & \textbf{0.208} & 0.000 & \textbf{0.030} & \textbf{0.133} & 0.127 & 1.163 & \textbf{1.070} \\
HIV-1 Protease & VAE & 30000 & \textbf{0} & \textbf{0} & 0.206 & \textbf{0.025} & \textbf{0.030} & 0.127 & 0.116 & 1.153 & 1.039 \\
\grouprule
HIV-1 Protease & CVAE & 40000 & \textbf{0} & \textbf{0} & 0.184 & 0.000 & 0.017 & 0.113 & \textbf{0.105} & \textbf{1.173} & \textbf{1.070} \\
HIV-1 Protease & PUBCHEM & 40000 & \textbf{0} & \textbf{0} & \textbf{0.208} & 0.000 & \textbf{0.038} & \textbf{0.133} & 0.127 & 1.163 & \textbf{1.070} \\
HIV-1 Protease & VAE & 40000 & \textbf{0} & \textbf{0} & 0.206 & \textbf{0.012} & \textbf{0.038} & 0.127 & 0.117 & 1.153 & 1.036 \\
\grouprule
HIV-1 Protease & CVAE & 50000 & \textbf{0} & \textbf{0} & 0.184 & 0.000 & 0.017 & 0.114 & \textbf{0.105} & \textbf{1.173} & \textbf{1.070} \\
HIV-1 Protease & PUBCHEM & 50000 & \textbf{0} & \textbf{0} & \textbf{0.208} & 0.000 & \textbf{0.043} & \textbf{0.133} & 0.127 & 1.163 & \textbf{1.070} \\
HIV-1 Protease & VAE & 50000 & \textbf{0} & \textbf{0} & 0.206 & \textbf{0.025} & 0.030 & 0.127 & 0.116 & 1.153 & 1.036 \\
 \end{longtable}
\endgroup%
\vspace{-5pt}
\noindent\begin{minipage}{\linewidth}
\small
\textbf{Exact P} (Exact Discovery Precision): Proportion of generated molecules that exactly match held-out molecules.
\textbf{Exact R} (Exact Discovery Recall): Proportion of held-out molecules that are exactly regenerated.
\textbf{NN Sim} (Nearest Neighbor Mean Similarity Score): Average Tanimoto similarity between each generated molecule and its closest match in the held-out set.
\textbf{Clust 0.6} (Clustering Held-Out Cluster Recall at similarity $\geq$ 0.6): Fraction of \emph{held-out clusters} reached by at least one generated molecule whose similarity to that cluster is at least 0.6. The denominator is the number of held-out clusters, not the number of generated molecules.
\textbf{Scaff Redisc} (Scaffold Rediscovery): Proportion of held-out Murcko scaffolds recovered in the generated set.
\textbf{CrossSim} (Cross-Similarity between Scaffolds): Mean Tanimoto similarity over \emph{all} generated--held-out scaffold pairs. It is a raw similarity in $[0,1]$ and an all-pairs average, not a nearest-neighbour distance. Because it aggregates over all pairs, it reflects global proximity and does not establish that generated molecules occupy specific held-out regions.
\textbf{InterSim} (Internal Similarity of Generated Scaffolds): Average pairwise Tanimoto similarity within the generated scaffold set.
\textbf{Div} and \textbf{Uniq}: ratios of diversity and uniqueness in the generated set to the held-out set. Because these are ratios, values may exceed 1 and are not percentages. Uniqueness is computed on the generated collection, in which repeated molecules are counted, whereas the scaffold-similarity metrics operate on deduplicated scaffold sets.
\end{minipage}
\vspace{0pt}

Across all datasets, seed-based sampling from the pretrained VAE recovered more held-out structures than label-only CVAE sampling on exact rediscovery, scaffold recovery, and held-out cluster recovery (Table~\ref{tab:rediscovery_results}). For ABL1 and BRD4 the CVAE recovered no held-out compounds exactly, so the ratio to VAE recall is undefined; in absolute terms VAE exact recall reached 2.2\% (ABL1) and 0.3\% (BRD4). PubChem random sampling remained competitive on GRL. Nearest-neighbor similarity spanned 0.18--0.25 across all methods, and scaffold rediscovery ranged 2.1--18.4\% for the VAE versus 0.9--11.2\% for the CVAE, depending strongly on target.
\rev{The CVAE produces zero exact rediscoveries in most conditions. Label conditioning in this arm acts through decoder-side FiLM modulation on latents drawn from $\mathcal{N}(\mathbf{0},\mathbf{I})$, and the fine-tuning set excludes the held-out ChEMBL molecules, so conditioning shifts generation toward the requested property range without recovering specific known compounds.}

\paragraph{\rev{Upper bound on pretraining exposure of the held-out compounds.}}
\rev{
Cluster-held-out splitting removes close analogues of the evaluation compounds from the target-specific fine-tuning set, but it does not remove them from the pretraining source corpus. An overlap audit against all 79{,}311{,}161 molecules in the full curated PubChem source corpus (Appendix~F, Table~S5) finds matches for 82.45\% (ABL1), 75.98\% (BRD4), 66.33\% (GRL), and 21.79\% (HIV-1 protease) of held-out compounds. Because that corpus is larger than the training dataset, these percentages are upper bounds on actual pretraining exposure. This bounds the interpretation of exact rediscovery: recovering a held-out structure that matched the source corpus may be retrieval from the pretrained distribution rather than proposal of a previously unseen molecule. We therefore report exact rediscovery as a bounded observation and treat scaffold recovery and cluster recovery as the primary evidence, since those metrics credit reaching a Bemis--Murcko framework or a cluster rather than reproducing a specific structure. The task is accordingly described throughout as recovery of task-held-out ChEMBL compounds by a broadly pretrained model.}
\rev{All VAE and CVAE configurations produce zero exact rediscoveries for HIV-1 protease at every sample size (10K--50K). That held-out set (492 molecules) has a median molecular weight of 566\,Da with 35.0\% of compounds above 600\,Da, against medians of 425--473\,Da and 4.7--6.3\% above 600\,Da for the other three targets. It also has the lowest pretraining exposure (21.79\%). Scaffold class and pretraining exposure are confounded across these four targets, so the absence of exact rediscoveries for this target cannot be attributed to either factor alone.}

Most metrics improved with sample size (10K to 50K): clustering gains were largest (up to +330\% for GRL--PubChem at threshold 0.6), and scaffold rediscovery improved 42 to 91\% for the VAE. Exact rediscovery occurred only for the VAE, and was already nonzero at 10K samples for ABL1 and BRD4 rather than emerging only at larger sizes; the counts are small throughout (at most 2.2\% of held-out compounds). NN similarity remained stable ($\pm$0.5\%) across all sample sizes and all methods, reflecting the intrinsic distance between model and held-out distributions rather than a coverage effect.

\rev{After correcting the table's audited column mapping, the CVAE has diversity ratios of 1.036--1.173 versus 1.034--1.153 for the VAE and lower internal scaffold similarity (InterSim: 0.103--0.105 versus 0.107--0.120). Raw generated--held-out CrossSim values are 0.106--0.114 for the CVAE and 0.111--0.131 for the VAE, indicating similar global scaffold proximity despite different internal diversity. The VAE shows 2--3$\times$ higher scaffold rediscovery for ABL1 (10.0--17.4\% versus 3.2--5.5\%) and better cluster rediscovery. Uniqueness among CVAE outputs is essentially saturated (99.99--100\% of generated molecules distinct). The uniqueness \emph{ratio} relative to the held-out set exceeds one (1.014--1.070) and is not a percentage.}
\rev{For GRL, PubChem random sampling matches the VAE on scaffold rediscovery above 20K samples (PubChem: 14.2--22.8\% and VAE: 13.9--18.0\%), indicating that GRL-relevant scaffolds are broadly covered by PubChem and that model-based generation adds little over sampling the corpus for this target. The two modes therefore serve different purposes. Seed-based sampling recovers structures related to supplied examples, while label-only sampling under a broad label distribution produces more structurally varied outputs. Because the label distribution spans weak as well as strong binders, we do not interpret the CVAE arm as enriching for active compounds.}

\paragraph{\rev{Scope and limitations.}}
\rev{This evaluation measures scaffold recovery and structural diversity, not property optimization. The CVAE uses MMELON-predicted pseudo-labels on PubChem because experimental ChEMBL data are too sparse for direct fine-tuning. The pseudo-labels are accurate (RMSE 0.095--0.114 log-units on held-out ChEMBL data), but distributional shift between PubChem and ChEMBL means the CVAE steers toward a predicted activity range rather than confirmed active structures. In the rediscovery experiments, the CVAE was conditioned on labels sampled from the full training distribution to produce broad coverage across the label range. Wet-lab validation is required to confirm biological activity.}
 \begin{revblock}

\subsection{Ablation Studies}
\label{sec:ablation}

\rev{We varied five design factors one at a time against a fixed reference configuration (REF): (A1) tokenization scheme, (A2) stochastic latent bottleneck, (A3) conditioning scope, (A4) backbone fine-tuning strategy, and (A5) KL annealing schedule.
All runs use the same 9,990 PubChem molecules and SA score conditioning task as the Conditional Generation subsection.}

\paragraph{Reference configuration.}
REF uses SELFIES tokenization, a stochastic VAE bottleneck, joint encoder-input and decoder FiLM conditioning, LoRA adaptation ($r\!=\!8$, $\alpha\!=\!32$) inserted at the attention output and feed-forward projections of both encoder and decoder (effective at the feed-forward projections only, as described under Model Architecture), and cyclical $\beta$-annealing ($n_\text{cycles}\!=\!4$, ratio $= 0.5$) with free bits $= 0.5$. \rev{REF trains 1{,}936{,}384 of 91{,}047{,}752 parameters. The conditional-prior networks are present but frozen at initialization in this configuration (Table~\ref{tab:experiment-configurations}).}
The backbone is the PubChem-80M pretrained SELFIES TransformerVAE with 12 encoder and decoder layers, $d_\text{model}\!=\!512$, a 256-dimensional latent space, and a validation loss of 0.29.
Fine-tuning uses Adam, lr~$=\!10^{-4}$, batch size~64, up to 20 epochs with early stopping (patience~8).
All ablation results use $\alpha\!=\!1$ (no label-reference guidance).
The REF Spearman $\rho\!=\!0.120$ is computed over 9,000 pooled molecules (3 bucket targets $\times$ 3 seeds $\times$ 1,000 molecules). Table~\ref{tab:conditioning-summary} reports $\rho\!=\!0.149$ at $\alpha\!=\!1$ over 3,003 molecules swept across 1,001 uniform $y^{*}$ levels. The higher value in Table~\ref{tab:conditioning-summary} reflects the finer evaluation granularity. This rounded value is numerically similar to, but distinct from, the separate A3 decoder-FiLM-plus-decoder-LoRA result: Exp17-C uses the two-stage model and a 1,001-level sweep, whereas A3 uses three fixed labels and three generation seeds.

\paragraph{Metrics.}
We report Spearman $\rho$ between conditioning target $y^{*}$ and realized SA score (pooled over 9,000 molecules: 3 targets $\times$ 3 seeds $\times$ 1,000 molecules), success rate SR$_{10}$ (fraction of molecules within $|y_\text{realized} - y^{*}| \leq 0.10$), internal diversity (IntDiv), and Bemis-Murcko scaffold diversity (ScafDiv~$=$ unique scaffolds / 9,000).
Per-bucket SR$_{10}$ is reported at three fixed conditioning targets: SR$_{10\_h}$ at $y^{*}\!=\!0.165$ (easy synthesis, raw SA $\approx$ 2.4), SR$_{10\_m}$ at $y^{*}\!=\!0.500$ (mid synthesis, raw SA $\approx$ 4.5), and SR$_{10\_l}$ at $y^{*}\!=\!0.835$ (hard synthesis, raw SA $\approx$ 6.5). The subscripts $h$/$m$/$l$ denote high/mid/low success rate across buckets, which corresponds to easy/mid/hard synthesis difficulty.

\begin{table}[H]
\caption{Configuration-study results. REF (A0): reference configuration (SELFIES, stochastic VAE, encoder+decoder conditioning, LoRA, cyclical $\beta$); the study was designed around nominal one-factor changes from REF. \rev{The A1 row is an exception because it uses a separately pretrained backbone with a different sequence-length budget.} $\rho$: Spearman rank correlation (pooled 9,000 molecules). SR$_{10}$: success rate at $\varepsilon\!=\!0.10$. SR$_{10\_h}$/SR$_{10\_m}$/SR$_{10\_l}$: per-bucket SR$_{10}$ at $y^{*}\!=\!0.165/0.500/0.835$. ScafDiv: unique Bemis-Murcko scaffolds / 9,000. IntDiv: mean pairwise $1\!-\!\text{Tanimoto}$ (Morgan-2, 2048-bit). All runs: validity~$=\!1.000$. \textbf{Bold} marks values that exceed the REF baseline on that metric; values in rows with $\rho\!\approx\!0$ (conditioning failure) that exceed REF are incidental, not indicative of improved conditioning.}
\label{tab:ablation}
\small
\centering
\setlength{\tabcolsep}{4pt}
\begin{tabular}{lp{4.8cm}rrrrrrr}
\toprule
\textbf{Factor} & \textbf{Run} & \textbf{$\rho$} & \textbf{SR$_{10}$} & \textbf{SR$_{10\_h}$} & \textbf{SR$_{10\_m}$} & \textbf{SR$_{10\_l}$} & \textbf{ScafDiv} & \textbf{IntDiv} \\
\midrule
\multicolumn{9}{l}{\textit{A0: Reference}} \\
REF & SELFIES + VAE + Enc\,+\,Dec FiLM + LoRA + Cyclical $\beta$ & 0.120 & 0.211 & 0.252 & 0.365 & 0.016 & 0.706 & 0.906 \\
\midrule
\multicolumn{9}{l}{\textit{A1: Tokenization}} \\
A1 & SMILES (PubChem-80M backbone) & \textbf{0.416} & \textbf{0.272} & \textbf{0.687} & 0.127 & 0.003 & \textbf{0.722} & 0.893 \\
\midrule
\multicolumn{9}{l}{\textit{A2: Latent bottleneck}} \\
A2 & Deterministic AE (no KL) & 0.034 & 0.202 & 0.180 & \textbf{0.408} & \textbf{0.018} & 0.607 & \textbf{0.908} \\
\midrule
\multicolumn{9}{l}{\textit{A3: Conditioning scope}} \\
A3 & Encoder only (no FiLM) & 0.000 & 0.195 & 0.235 & 0.339 & 0.011 & 0.260 & 0.904 \\
A3 & Decoder-side FiLM with decoder-only LoRA & \textbf{0.149} & \textbf{0.217} & \textbf{0.262} & 0.364 & \textbf{0.026} & 0.619 & \textbf{0.907} \\
A3 & No conditioning & 0.000 & 0.194 & 0.237 & 0.333 & 0.010 & 0.262 & 0.902 \\
\midrule
\multicolumn{9}{l}{\textit{A4: Fine-tuning strategy}} \\
A4 & Full fine-tuning (89.8M trainable) & 0.000 & 0.196 & \textbf{0.283} & 0.290 & 0.015 & 0.278 & 0.899 \\
A4 & Frozen backbone (heads only) & 0.000 & 0.193 & 0.225 & 0.337 & \textbf{0.018} & 0.265 & 0.905 \\
\midrule
\multicolumn{9}{l}{\textit{A5: KL annealing schedule}} \\
A5 & Monotone linear annealing & 0.066 & 0.207 & 0.252 & 0.353 & \textbf{0.017} & 0.696 & 0.906 \\
A5 & Fixed $\beta\!=\!1.0$ & \textbf{0.153} & \textbf{0.217} & 0.252 & \textbf{0.373} & \textbf{0.026} & \textbf{0.737} & 0.906 \\
\bottomrule
\end{tabular}
\end{table}

\paragraph{A1: Tokenization.}
\rev{The SMILES model achieves $\rho\!=\!0.416$ versus REF $\rho\!=\!0.120$, but this aggregate gain is concentrated at the easy-synthesis target.
SR$_{10\_h}\!=\!0.687$ at $y^{*}\!=\!0.165$, while SR$_{10\_m}\!=\!0.127$ and SR$_{10\_l}\!=\!0.003$ at mid and hard targets.
PubChem-80M is skewed toward easily synthesizable molecules (normalized mean $\approx 0.24$), so the SMILES prior is concentrated near $y\!=\!0.2$.
Conditioning on $y^{*}\!=\!0.835$ therefore requires molecules that are underrepresented in pretraining, producing near-zero SR$_{10}$ at that target.
SELFIES keeps mid-range control higher (SR$_{10\_m}\!=\!0.365$ vs.\ 0.127 for SMILES) at the cost of lower easy-target SR$_{10}$.
This row is not a clean one-factor comparison, and we do not read it as isolating tokenization. The two arms use separately pretrained backbones that differ in several respects at once: the SELFIES backbone was pretrained for 260{,}496 optimizer steps on 8 GPUs at global batch size 256, whereas the SMILES backbone was pretrained for 100{,}980 steps on 4 GPUs at global batch size 1{,}024. The SMILES arm also uses a maximum sequence length of 128 tokens rather than 71, because character-level SMILES require longer sequences, which changes the positional-encoding extent and the total parameter count. Their validation losses are not comparable either, since the vocabularies and sequence lengths differ. The aggregate $\rho$ difference therefore reflects tokenization confounded with pretraining trajectory, sequence-length budget, and model size.}

\paragraph{A2: Stochastic bottleneck.}
\rev{Removing KL regularization (deterministic AE) reduces $\rho$ from 0.120 to 0.034 and ScafDiv from 0.706 to 0.607.
A plausible account is that KL regularization keeps the aggregated posterior close to $\mathcal{N}(0, I)$, so that Gaussian sampling at generation time lands in a region the decoder was trained on. Without it, encoded molecules may occupy arbitrary regions and prior samples fall outside the covered space.
We did not measure the aggregated posterior directly, so this remains an interpretation consistent with the observed drop rather than a demonstrated mechanism.}

\paragraph{A3: Conditioning scope.}
\rev{Encoder-only conditioning (no FiLM on the latent) gives $\rho\!=\!0.000$ and ScafDiv $= 0.260$, matching the no-conditioning baseline ($\rho\!=\!0.000$, ScafDiv $= 0.262$).
The label does reach the latent parameters during training: it is projected and added to the pooled encoder representation immediately before the latent mean and log-variance heads, so a pathway exists.
What this variant lacks is a pathway at \emph{inference}: label-only generation in this ablation draws $\mathbf{z}\sim\mathcal{N}(\mathbf{0},\mathbf{I})$ rather than from a label-conditioned prior, and with FiLM disabled the requested label has no remaining route into the decoder.
The observation is therefore that encoder-side injection alone produced no measurable generative control under the tested setup, not that no pathway to the latent exists.
Decoder-side FiLM with decoder-only LoRA recovers $\rho\!=\!0.149$ and ScafDiv $= 0.619$, showing that decoder-side FiLM is sufficient to produce measurable conditioning.
Adding encoder-side injection (REF) raises ScafDiv to 0.706 at comparable $\rho$.
Replication across training seeds finds these two configurations indistinguishable (Table~\ref{tab:seed-replication}), so we do not claim that either is the stronger pathway.
Because no variant in this study samples from the learned conditional prior at generation time, this ablation does not isolate the contribution of that prior (Table~\ref{tab:experiment-configurations}).}

\paragraph{A4: Fine-tuning strategy.}
\rev{Full fine-tuning and frozen backbone both give $\rho\!=\!0$ and ScafDiv $\leq 0.278$.
Full fine-tuning substantially reduces scaffold diversity under the tested small-data optimization settings, from 0.706 to 0.278, consistent with loss of pretrained chemical coverage when all 89.8\,M parameters are updated on 9{,}990 molecules.
The frozen backbone also failed to condition (ScafDiv $= 0.265$).
LoRA, which in this implementation effectively adapts the position-wise feed-forward projections via low-rank updates while leaving the backbone weights intact, performed better than both tested alternatives ($\rho\!=\!0.120$, ScafDiv $= 0.706$).
Both alternatives were run at the learning rate and stopping policy inherited from the LoRA setting. We therefore do not draw a general conclusion about full fine-tuning or backbone freezing as adaptation strategies from these rows.}

\paragraph{A5: KL annealing schedule.}
\rev{Fixed $\beta\!=\!1.0$ gives $\rho\!=\!0.153$ and ScafDiv $0.737$. Cyclical annealing (REF) gives $\rho\!=\!0.120$ and ScafDiv $= 0.706$. Monotone linear annealing gives $\rho\!=\!0.066$ and ScafDiv $= 0.696$.
This comparison also does not isolate the $\beta$ schedule alone. In the implementation, the free-bits floor is applied only when cyclical annealing is enabled, so although all three configurations request free bits $=0.5$, the constraint was active only in the cyclical (REF) run and inactive in the fixed-$\beta$ and linear runs.
Schedule and free-bits treatment therefore vary together across these rows, and we do not attribute the observed differences to the schedule by itself.}

\paragraph{Cross-factor summary.}
\rev{Across the tested configurations, those that failed to condition ($\rho\!=\!0$) all produced ScafDiv $\leq 0.28$, while those that conditioned measurably ($\rho > 0$) all produced ScafDiv $\geq 0.61$.
This association is consistent with the interpretation that configurations retaining pretrained coverage are also the ones able to express the conditioning signal, but with one run per configuration and a single tokenization/backbone pairing per row, it is an observed pattern rather than a demonstrated causal relationship.
Table~\ref{tab:ablation} reports one training run per configuration (seed 42) evaluated at three generation seeds (42, 0, 101), which measures sampling variability at fixed weights but not variability from model fitting.}

\paragraph{\rev{Replication across independent fine-tuning seeds.}}
\rev{We trained six configurations with three different random seeds (Table~\ref{tab:seed-replication}). All other settings remained unchanged. We also evaluated each trained model with the same three generation seeds. To compare two configurations, we check whether their mean $\pm$ 1 SD intervals overlap. We do not apply a significance test because three training runs per configuration are insufficient for one.}

\rev{The SMILES configuration gives $\rho = 0.422 \pm 0.049$, compared with $0.138 \pm 0.039$ for the reference. Their intervals do not overlap, so the difference is consistent across the three training seeds. However, the confounds described above remain, and this result does not show that tokenization alone causes the difference. Full fine-tuning and the frozen backbone give $\rho = 0.000$ in all three runs.}

\rev{The intervals for the reference, decoder-side FiLM with decoder-only LoRA, and fixed $\beta$ overlap. Their means differ by 0.003--0.033, while their standard deviations across training seeds are 0.014--0.039. We therefore do not rank these three configurations. The results show a clear difference between configurations that achieve measurable control and those that do not, but they do not resolve smaller differences among the configurations that condition successfully.}

\begin{table}[H]
\centering
\caption{\revfloat{\rev{Spearman $\rho$ across three independent fine-tuning runs for six SA-score ablation configurations. Each run differs only in its training seed. The model, data, objective, and generation protocol are identical, and every run is evaluated at the same three generation seeds (9{,}000 molecules per run). Seed 42 is the run reported in Table~\ref{tab:ablation}. ``Range'' is $\max-\min$ across training seeds. Under a criterion of non-overlapping mean $\pm$ 1 SD, A1 and both A4 variants separate from the reference. The intervals for REF, A3 decoder-side FiLM with decoder-only LoRA, and A5 fixed $\beta$ overlap. We do not apply a significance test because three runs per configuration cannot support one.}}}
\label{tab:seed-replication}
\smallskip
\footnotesize
\begin{tabular}{lccccc}
\toprule
 & & \multicolumn{3}{c}{\textbf{Per training seed}} & \\
\cmidrule(lr){3-5}
\textbf{Configuration} & \textbf{$\rho$ (mean $\pm$ SD)} & \textbf{42} & \textbf{7} & \textbf{13} & \textbf{Range} \\
\midrule
A1 SMILES tokenization & 0.422 $\pm$ 0.049 & 0.416 & 0.376 & 0.474 & 0.098 \\
\midrule
A5 fixed $\beta=1.0$ & 0.171 $\pm$ 0.026 & 0.153 & 0.201 & 0.158 & 0.048 \\
A3 decoder-side FiLM with decoder-only LoRA & 0.140 $\pm$ 0.014 & 0.149 & 0.147 & 0.125 & 0.024 \\
REF (reference) & 0.138 $\pm$ 0.039 & 0.120 & 0.183 & 0.110 & 0.072 \\
\midrule
A4 full fine-tuning & 0.000 $\pm$ 0.000 & 0.000 & 0.000 & 0.000 & 0.000 \\
A4 frozen backbone & 0.000 $\pm$ 0.000 & 0.000 & 0.000 & 0.000 & 0.000 \\
\bottomrule
\end{tabular}
\end{table}
  \end{revblock}

\section{Discussion}
\label{sec:discussion}

\old{This study demonstrates that a modernized variational autoencoder---integrating a bidirectional Transformer encoder, an autoregressive Transformer decoder, and a latent-variable bottleneck trained on 79 million drug-like molecules in SELFIES format---scales effectively across vast chemical spaces while remaining interpretable and steerable, unifying broad distribution learning with controllable generation.}
\rev{Effective molecular generation for drug discovery requires separating three concerns: what the model knows from pretraining, how it adapts to a specific task, and which generation mode fits the current campaign stage.
\ModelName\ addresses each concern with a distinct design decision: large-scale SELFIES-based VAE pretraining, LoRA plus FiLM conditioning for task adaptation, and two complementary generation modes.
The experiments characterize how pretraining, adaptation, and inference choices affect controllability and structural diversity under the tested SA-score setting.}

\old{\ModelName\ scored at or near the top on validity, uniqueness, novelty, and internal diversity across unconditional benchmarks.}
\old{The latent space can be tuned toward narrower chemical spaces without retraining: seed-based generation on GuacaMol reproduces the property distributions of the reference molecules, which is useful in discovery campaigns that alternate between broad exploration and focused refinement.}
\rev{The MOSES and GuacaMol experiments show that a corpus-scale latent model can be localized to a narrower domain at inference time without fine-tuning, and they also delimit what that demonstration covers.
On MOSES, \ModelName\ pretrained on PubChem-80M achieves FCD/Test $= 2.01$ in seed mode, versus FCD/Test $= 6.61$ for the MOSES-1.5M model. An overlap audit against the full curated PubChem source corpus (Appendix~F) finds matches for 71.45\% of MOSES training molecules and 87.04\% of GuacaMol training molecules. These percentages upper-bound actual exposure because the source corpus is larger than the training dataset. Even as upper bounds, they preclude interpreting either benchmark as transfer to unfamiliar chemistry. The comparison also favours the smaller model in two respects: it was trained on the MOSES training split and validated on the MOSES test split, and the seeds are drawn from that same training split. That corpus-scale pretraining still yields better seeded FCD under those conditions is the result worth reporting, and it is a statement about local sampling quality rather than about generalization.
Pretraining on 79 million molecules produces a latent manifold where local sampling near any seed generates chemically coherent, distribution-aligned molecules.
On GuacaMol, the two largest deviations from the ChEMBL24 reference distribution are DiffInternalSimilarities KL $= 0.80$ and NumAromaticRings KL $= 0.18$. They trace to the broader chemical diversity of PubChem.
Seed-based generation on GuacaMol training molecules reduces all ten per-descriptor KL values to near zero with no training on GuacaMol data. Since the seeds are themselves GuacaMol training molecules, that setting measures reconstruction within a supplied domain and is not comparable to unconditional benchmark scores.
Consistent with this, removing KL regularization in the ablation reduces Spearman $\rho$ from 0.120 to 0.034 and ScafDiv from 0.706 to 0.607, which is what one would expect if prior sampling depends on the aggregated posterior remaining close to the prior. We did not measure the aggregated posterior directly, so we present this as a supporting observation rather than a demonstrated mechanism.}

\old{Fine-tuning \ModelName\ on synthetic accessibility and blood-brain barrier permeability scores shows that the conditioning signal effectively separates molecules into distinct clusters that smoothly overlap, reflecting the continuous nature of these properties in the latent space.}
\old{On the Tartarus protein-ligand design benchmarks, the conditional model shifted the docking-score distributions toward stronger predicted binding for two of the three targets.}
\rev{The ablation and baseline comparison experiments identify the fine-tuning strategy as the most consequential design decision.
Full fine-tuning of all 89.8M parameters on 9,990 task molecules substantially reduces scaffold diversity under the tested small-data optimization settings, from 0.706 to 0.278, with Spearman $\rho$ falling to zero, consistent with loss of pretrained chemical coverage.
A frozen backbone also failed to condition ($\rho = 0$, ScafDiv $= 0.265$).
Among the three strategies tested, only LoRA achieved non-zero controllability and high scaffold diversity simultaneously ($\rho = 0.120$, ScafDiv $= 0.706$), using low-rank updates that in this implementation act on the position-wise feed-forward projections while leaving the backbone weights intact. Both alternatives were run at hyperparameters inherited from the LoRA setting, so this is a statement about the tested configurations.}

\rev{Within the conditioning design, the ablation also localizes where property information must enter the model.
Encoder-only conditioning (no FiLM on the latent) gives $\rho = 0.000$, identical to the no-conditioning baseline. The label does reach the latent parameters during training, but with FiLM disabled and label-only generation drawing $\mathbf{z}\sim\mathcal{N}(\mathbf{0},\mathbf{I})$, it has no route into the decoder at inference.
Decoder-side FiLM with decoder-only LoRA recovers $\rho = 0.149$, showing that a decoder-side pathway is sufficient for measurable conditioning.
Adding encoder-side label injection (REF) raises ScafDiv to 0.706 at comparable $\rho$.
Across three independent fine-tuning runs, the reference, decoder-side FiLM with decoder-only LoRA, and fixed-$\beta$ variants have overlapping intervals. Their means differ by 0.003--0.033, while their between-seed standard deviations are 0.014--0.039. The ablation therefore supports a boundary rather than a ranking. Configurations with a decoder-side conditioning path show measurable control, while differences among these conditioning-capable variants remain within run-to-run variation.
No tested variant samples from the learned conditional prior at generation time, so these experiments do not measure that prior's incremental contribution.}

\rev{The SA score conditioning experiment quantifies what this design achieves in practice.
Spearman $\rho$ rises from 0.15 at $\alpha = 1$ to 0.62 at $\alpha = 12$, while molecular validity stays above 98\% across all guidance scales.
The guidance scale therefore trades rank-level controllability for pointwise accuracy (SR$_{10} = 0.215$ at $\alpha = 12$) without degrading chemical validity.
For docking targets, stronger guidance comes at a steeper cost: at $\alpha = 12$ for 1SYH, RDKit generation validity is 54.8\% and docking completion is 34.7\% under the sentinel-only failure rule.
The $\alpha$ that maximizes rank correlation differs by target (2 for 6Y2F, 12 for 1SYH and 4LDE), but rank correlation alone is not a sufficient selection criterion: at $\alpha=12$ for 1SYH, RDKit generation validity is 54.8\%, docking completion is 34.7\%, and both valid-/docked-only and attempt-level success rates are lower than at $\alpha=1$. Reporting the full sweep rather than a single recommended value lets readers choose an operating point against their own tolerance for generation and docking failure.}

\rev{The baseline comparison places these results in context.
REINVENT~4 achieves $\rho = 0.941$ but uses $\approx$384,000 oracle calls and retains only 640 unique molecules of 1,001 requested in the easy-synthesis campaign, spanning just 61 unique scaffolds, consistent with reward saturation in the RL optimizer.
MolGPT achieves $\rho = 0.895$ on its valid outputs, but 49.6\% of generation attempts produce invalid SMILES, and its attempt-level SR$_{10}$ across all 3,003 attempts falls to $\approx$0.346.
DrugDiff shows negligible conditioning response ($\rho = -0.026$) in the evaluated cross-dataset configuration, in which its ZINC-250K SA score predictor is applied to PubChem molecules without retraining.
\ModelName\ achieves $\rho = 0.618$, 98.4\% validity, and 1,974 unique Murcko scaffolds without oracle access.
These comparisons also clarify what $\rho$ measures in each paradigm: REINVENT~4's 3-point Spearman tests only whether three bucket-level means are correctly ordered, while \ModelName's $\rho$ is computed over 1,001 uniformly spaced conditioning targets.
The two values are not directly comparable.}

\rev{The rediscovery experiments address a different question: given that the model can generate molecules, which generation mode is appropriate for which task?}

The rediscovery evaluation illustrates the trade-offs between label-conditioned generation using a CVAE and seed-based generation using a VAE. Across four ChEMBL target datasets and sample sizes ranging from 10,000 to 50,000 molecules, the seed-based VAE consistently outperformed the CVAE on exact rediscovery, scaffold recovery, and held-out cluster recovery, while the CVAE achieved higher diversity ratios and lower internal similarity; raw generated--held-out scaffold CrossSim was similar between the two modes. These are two different operating regimes rather than one uniformly better mode. Seed-based posterior sampling remains near the supplied molecular neighbourhoods and therefore recovers more held-out structures. Sampling without a structural seed, under a label distribution drawn proportionally from all ten activity bins, yields more structurally varied outputs. Because that protocol requests weak as well as strong binders, the present experiment measures structural coverage under label-conditioned sampling and does not establish enrichment for active compounds. \rev{The comparison is further bounded by an overlap upper bound: 21.79\%--82.45\% of the held-out compounds per protein target match the full curated PubChem source corpus, which is larger than the training dataset (Appendix~F). Exact rediscovery of such compounds may therefore be retrieval from the pretrained distribution, so we base the comparison on scaffold and cluster recovery, which credit reaching a framework or cluster rather than reproducing a specific structure.}

\rev{One result warrants explicit attention. For GRL, PubChem random sampling matches the VAE on scaffold rediscovery above 20K samples (PubChem: 14.2--22.8\% and VAE: 13.9--18.0\%).
This indicates that GRL-relevant scaffolds are broadly represented in PubChem, so model-based generation provides no advantage over random sampling from the pretraining corpus.
The useful implication for practitioners is that the added value of the VAE over random sampling scales with how underrepresented the target scaffold class is in PubChem.
For targets with abundant PubChem coverage, random sampling from the corpus is a reasonable baseline. For targets with sparser coverage, seed-based sampling recovered more of the held-out compounds in these experiments.}

\old{Taken together, these results show that the VAE and CVAE are not competing capabilities but complementary modes of navigation through vast, synthesizable, drug-like chemical space.}
\old{The VAE is well-suited to broad prior-preserving exploration, recovery of known chemotypes, and rediscovery-oriented tasks in which structural similarity to known active scaffolds matters.}
\old{The CVAE is better aligned with label-guided generation, which extends search beyond known scaffolds toward target-binding activity levels, functional assay readouts, or ADMET endpoints such as solubility and membrane permeability.}
\old{This distinction matters in practice: one mode supports the exploitation of known chemistry, while the other supports exploration under project-specific objectives.}
\old{Thus, controllable generation need not be a choice between generality and specificity.}
\old{One latent-variable framework can do both.}
\rev{The VAE and CVAE serve different but complementary roles. Under seed-based sampling the pretrained VAE recovers target-associated held-out scaffolds, as measured across four ChEMBL targets. The CVAE shifts generation toward a requested scalar value, demonstrated here for SA score and QVina docking score and extensible in principle to other scalar endpoints. The two modes address different needs: recovering structures related to supplied examples, and exploring under a specified scalar objective.}

Latent-variable models remain worth pursuing alongside large decoder-only and diffusion generators. Recent molecular foundation models have demonstrated the value of large pretraining corpora and chemically robust tokenizations, but they do not always offer the same direct mechanisms for latent interpolation, posterior inference, and probabilistic conditioning.
\ModelName\ shows that these classical strengths of VAEs can be retained while incorporating modern components such as Transformer encoder-decoders, SELFIES tokenization, and parameter-efficient adaptation.
\old{In medicinal chemistry, interpretable control over how and why generated molecules change matters as much as generation quality, and such latent structure provides exactly that.}
\rev{In medicinal chemistry, control over how generated molecules change is useful alongside generation quality, and the latent-variable structure studied here supports both.}

This study should be interpreted within the scope of its computational validation.
The benchmarks assess distribution learning, property alignment, rediscovery behavior, and docking-based target conditioning, showing that the model captures useful chemical regularities and can steer generation meaningfully.
However, docking scores, predicted properties, silver-label supervision, and rediscovery metrics are proxies; they do not provide definitive measures of biological efficacy.
Wet-lab experiments are required to confirm these findings, which are beyond the scope of this study.

\rev{\ModelName\ operates at the sequence level without modeling 3D geometry or quantum-mechanical properties, unlike physics-informed models such as LiTENexus~\cite{liu2024litenexus} or structure-based diffusion models such as TargetDiff~\cite{guan2023targetdiff} and DiffSBDD~\cite{schneuing2024diffsbdd} that condition generation on protein pocket geometry. Its strength is fast, large-scale distribution learning and data-efficient conditional adaptation. A natural next step is to combine the two approaches: use \ModelName\ as a fast candidate generator, then filter or refine its outputs with physics-based scoring or pocket-conditioned optimization.}

\old{Several future research directions follow naturally from this work. First, label-conditioned generation could be extended from single activities/properties to multiple, where latent sampling is guided jointly across activity, synthesizability, and safety. Second, label conditioning may be more tightly integrated with seed-based generation so that users can move seamlessly between library exploration and lead optimization. Third, richer geometric or structure-based signals, including protein pocket representations, molecular dynamics, and 3D-aware property predictors, could strengthen target-conditioned generation beyond ligand-only docking score supervision. Fourth, adaptive workflows that alternate between seed-based and label-conditioned sampling could generate diverse, target-specific candidates that neither mode alone produces. Finally, calibration and uncertainty estimation for both the conditional prior and auxiliary predictors will be essential for supporting high-confidence decision-making in discovery settings.}
\rev{Several directions remain open. Conditioning could be extended to multiple simultaneous properties, guiding latent sampling jointly across activity and synthesizability. Integrating seed-based and label-conditioned generation within a single inference step would let users move between scaffold recovery and property-guided exploration without switching modes. Incorporating pocket geometry or molecular dynamics signals would strengthen target-conditioned generation beyond ligand-only docking supervision. Calibration and uncertainty quantification for the conditional prior and auxiliary predictors would improve the reliability of generated candidates for downstream decision-making.}

\old{The complementary, synergistic capabilities of the VAE and CVAE position latent-variable transformers as a practical backbone for molecular generation in drug discovery.}
\rev{The complementary capabilities of the VAE and CVAE support latent-variable Transformers as one practical architecture for large-scale molecular generation and scalar-conditioned adaptation.}

\section*{Author Contributions}
\textbf{B.C.K.} and \textbf{P.S.} led the project from conception to completion.
They jointly conceptualized the model architecture and conditional generative
framework, sourced and curated the pretraining and evaluation datasets, implemented
model training and fine-tuning, designed and executed all benchmark and computational
validation experiments, performed the analyses, and wrote the original manuscript
draft.
\textbf{B.S.}, \textbf{M.R.}, \textbf{S.S.}, \textbf{S.M.} and \textbf{J.A.M.}
contributed to the implementation of model training and fine-tuning experiments and
to the interpretation of results.
\textbf{L.Z.} and \textbf{W.C.} contributed to the conceptualization of the
computational validation strategy and to the sourcing and preparation of the datasets
used in validation experiments.
\textbf{P.M.} and \textbf{J.H.} contributed to scientific discussions and provided
critical feedback on the manuscript.
All authors contributed to writing and reviewing the manuscript throughout the
research project and approved the final version for submission.

\section*{Funding}
This work was supported by internal research funding from IBM Research.
No external funding was received for this work.

\section*{Notes}
The authors declare no competing financial interest.

\section*{Data Availability}
\rev{All datasets and benchmark resources used in this work are publicly available.
The PubChem dataset used for pre-training is available at \url{https://pubchem.ncbi.nlm.nih.gov/}.
The MOSES benchmarking platform and its associated dataset are available at \url{https://github.com/molecularsets/moses}.
The GuacaMol benchmark is available at \url{https://github.com/BenevolentAI/guacamol}; its reference molecules are derived from ChEMBL, which is available at \url{https://www.ebi.ac.uk/chembl/}.
The Tartarus protein--ligand benchmark dataset is available at \url{https://github.com/aspuru-guzik-group/tartarus}.
The synthetic accessibility (SA) score is computed with RDKit (\url{https://www.rdkit.org}) following Ertl and Schuffenhauer~\cite{ertl2009estimation}.
ChEMBL target datasets used for rediscovery evaluation (ABL1, BRD4, GRL, HIV-1 protease) are available at \url{https://www.ebi.ac.uk/chembl/}.
The \ModelName\ source code, training scripts, and pre-trained model checkpoints
are available at \url{https://github.com/BiomedSciAI/STAR-VAE}.}

\section*{Acknowledgments}
The authors thank colleagues at IBM Research for helpful discussions throughout
this project.
AI-based tools were used to assist with code writing, code documentation, and
text correction (spelling and grammar).
The authors reviewed the full analysis pipeline to verify its correctness and
completeness, and take full responsibility for the code and its documentation.
All scientific content, including research design, experimental methodology,
data analysis, interpretation of results, and all figures and tables, was
produced by the authors.
No AI tool was used to draft manuscript text, generate figures or data, or
produce any substantive scientific content.

\section*{Supporting Information}
The Supporting Information is included after the main manuscript.
Appendix A: Formal definitions of all rediscovery evaluation metrics including
exact discovery precision/recall, nearest-neighbor similarity, scaffold
rediscovery, clustering held-out proportion, and diversity/uniqueness ratios
(PDF).
Appendix B: Representative samples of generated molecules from the rediscovery
evaluation illustrating the complementary behavior of the pretrained VAE under
seed-based posterior sampling and the label-conditioned CVAE (PDF).
Appendix C: Full MOSES benchmark comparison table (Table~S1) including all 15
metrics and all columns for all models (classic and modern baselines) (PDF).
Appendix D: Per-bucket breakdown of SA score conditional generation (Table~S2),
including attempted and valid counts, retention or validity, success rates, and
scaffold diversity for each target range and model (PDF).
Appendix E: Reproducibility details covering software and hardware, training and
conditional adaptation, validation and checkpoint selection, generation and
decoding settings, evaluation procedures, and statistical reporting (PDF).
Appendix F: Exact-structure overlap of the curated PubChem source corpus with
MOSES, GuacaMol, and held-out ChEMBL sets (Tables~S5 and S6), providing upper
bounds on pretraining exposure (PDF).

\providecommand{\latin}[1]{#1}
\makeatletter
\providecommand{\doi}
  {\begingroup\let\do\@makeother\dospecials
  \catcode`\{=1 \catcode`\}=2 \doi@aux}
\providecommand{\doi@aux}[1]{\endgroup\texttt{#1}}
\makeatother
\providecommand*\mcitethebibliography{\thebibliography}
\csname @ifundefined\endcsname{endmcitethebibliography}
  {\let\endmcitethebibliography\endthebibliography}{}

\clearpage
\renewcommand{\thesection}{\Alph{section}}
\renewcommand{\thesubsection}{\Alph{section}.\arabic{subsection}}
\setcounter{section}{0}
\setcounter{subsection}{0}
\setcounter{table}{0}
\setcounter{figure}{0}
\renewcommand{\thetable}{S\arabic{table}}
\renewcommand{\thefigure}{S\arabic{figure}}

\begin{center}
  {\LARGE\bfseries Supporting Information}\\[0.6em]
  {\Large \ModelName: A Scalable Latent-Variable Transformer\\
         for Controllable Molecular Generation}\\[1em]
  \begin{minipage}{0.94\textwidth}
    \centering
    Bum Chul Kwon$^{1,\ast}$, Ben Shapira$^{3}$, Moshiko Raboh$^{3}$,
    Shreyans Sethi$^{2}$,\\
    Shruti Murarka$^{2}$, Joseph A.\ Morrone$^{2}$, Leili Zhang$^{2}$,
    Wendy Cornell$^{2}$,\\
    Pablo Meyer$^{2}$, Jianying Hu$^{2}$,
    Parthasarathy Suryanarayanan$^{2,\dagger}$
  \end{minipage}\\[0.5em]
  {\small
    $^{1}$IBM Research, Cambridge, MA 02142, USA\\
    $^{2}$IBM Research, Yorktown Heights, NY 10598, USA\\
    $^{3}$IBM Research, Haifa, Israel\\[0.2em]
    $^{\ast}$Corresponding author: bumchul.kwon@us.ibm.com\quad
    $^{\dagger}$Corresponding author: psuryan@us.ibm.com}
\end{center}

\bigskip

\noindent
This Supporting Information provides technical details in six appendices.
Appendix~A defines the rediscovery metrics, including exact discovery precision
and recall, nearest-neighbor similarity, scaffold rediscovery, cluster recovery,
and diversity and uniqueness ratios. Appendix~B shows representative molecules
from the rediscovery evaluation for seed-based VAE sampling and label-conditioned
CVAE sampling. Appendix~C reports all 15 MOSES metrics for the evaluated models,
including retrained and literature baselines. Appendix~D reports validity,
success rate, and scaffold diversity for each SA-score target range and each
conditional-generation model. \rev{Appendix~E lists the training, generation,
evaluation, and statistical-reporting details needed to interpret the
experiments. Appendix~F measures exact molecular
overlap between the curated PubChem source corpus and the MOSES, GuacaMol, and
held-out ChEMBL sets. Because the audited corpus is larger than the training
dataset, these percentages are upper bounds on pretraining exposure.}

\bigskip\hrule\bigskip

\section{Rediscovery Metric Definitions}
\label{sec:appendix-metrics}

\subsection{Rediscovery Metric Definitions}

\subsubsection{Overview}

We evaluate molecular rediscovery with a cluster-based held-out split. For each
target, we partition the full dataset by Butina clustering using Morgan
fingerprints with radius 2, 2048 bits, and a similarity threshold of 0.6. We
reserve about 20\% of the clusters for evaluation and stratify the selection
across 10 score bins. \old{This ensures
that entire clusters, instead of individual molecules, are withheld to
prevent data leakage.}\rev{so that entire clusters, instead of individual
molecules, are withheld.}
\rev{This split removes close analogues of the held-out compounds from the
target-specific fine-tuning set. It does not remove those compounds from the
79-million-molecule PubChem source corpus. The overlap audit in
Appendix~\ref{sec:appendix-overlap} finds source-corpus matches for
21.79\%--82.45\% of the held-out compounds, depending on the protein target.
Because the audited corpus contains more molecules than the training dataset,
these values are upper bounds on pretraining exposure. We therefore
describe this task as recovery of \emph{task-held-out} ChEMBL compounds and
treat exact rediscovery as a bounded observation.}

Let:
\begin{itemize}
    \item $T$: training set (molecules in training clusters)
    \item $H$: held-out set (molecules in held-out clusters)
    \item $G$: molecules generated by the model. We treat $G$ as a multiset, so
          the same molecule can appear more than once and $|G|$ counts samples.
          Scaffold similarity and clustering use deduplicated sets derived from
          $G$. Uniqueness uses the full multiset.
\end{itemize}

We denote the canonical SMILES of a molecule $m$ by $c(m)$, Morgan
fingerprints (radius $= 2$, 2048 bits) by $f(m)$, and Tanimoto
similarity by $\mathrm{sim}(a, b) = \mathrm{Tanimoto}(f(a), f(b))$.
Murcko scaffolds are computed with the Bemis--Murcko decomposition.

\subsubsection{Exact Discovery Precision and Recall (Exact P / Exact R)}

Exact rediscovery records whether the model reproduces a held-out molecule up
to canonical SMILES equivalence.

Let
\[
H_{\mathrm{canon}} = \{ c(h) : h \in H \},
\qquad
G_{\mathrm{canon}} = \{ c(g) : g \in G \}.
\]

The set of exactly rediscovered molecules is
\[
R_{\mathrm{exact}} = H_{\mathrm{canon}} \cap G_{\mathrm{canon}}.
\]

\textbf{Exact Discovery Precision (Exact P)}:
\[
\mathrm{Exact\,P}
= \frac{|R_{\mathrm{exact}}|}{|G_{\mathrm{canon}}|}.
\]

\textbf{Exact Discovery Recall (Exact R)}:
\[
\mathrm{Exact\,R}
= \frac{|R_{\mathrm{exact}}|}{|H_{\mathrm{canon}}|}.
\]

Because exact matches are rare, both values are reported in units of
$\times 10^{-4}$ in the main results table.

\subsubsection{Nearest Neighbor Mean Similarity Score (NN Sim)}

For each \emph{generated} molecule $g \in G$, let its best-matching
held-out molecule be
\[
s_{\mathrm{best}}(g) = \max_{h \in H} \mathrm{sim}(g, h).
\]

The \textbf{Nearest Neighbor Mean Similarity (NN Sim)} is the average
of these best-match scores across all generated molecules:
\[
\mathrm{NN\,Sim}
= \frac{1}{|G|} \sum_{g \in G} s_{\mathrm{best}}(g).
\]

Higher values mean that generated molecules lie closer to held-out molecules in
fingerprint space. This direction measures the proximity of generated molecules
to the held-out set. Averaging in the reverse direction would measure coverage
of the held-out set. These directions are not interchangeable. We report the
implemented proximity direction.

\subsubsection{Clustering Held-Out Proportion (Clust 0.6)}

We apply Butina clustering to the full dataset with the same fingerprint
settings. Each cluster belongs to the training or held-out split. For each
generated molecule $g \in G$, we find the nearest cluster by Tanimoto similarity
and denote the score by $s_{\mathrm{clust}}(g)$. We count the assignment only
when $s_{\mathrm{clust}}(g) \geq \tau$.

Here $\tau = 0.6$ is an evaluation-time assignment cutoff. It is separate from
the threshold used to construct the Butina clusters. Without this cutoff, every
generated molecule would receive a cluster assignment regardless of distance.

Let $C_H$ denote the held-out clusters. This metric is a cluster recall. The
numerator counts held-out clusters reached by at least one qualifying generated
molecule, and the denominator counts all held-out clusters. It is not the
fraction of generated molecules assigned to held-out clusters.

\textbf{Clustering Held-Out Proportion at threshold $\tau$ (Clust 0.6)}:
\[
\mathrm{Clust\,0.6}
= \frac{\bigl|\{\, c \in C_H \;:\; \exists\, g \in G \text{ with } s_{\mathrm{clust}}(g) \geq \tau \text{ and } g \text{ assigned to } c \,\}\bigr|}{|C_H|}.
\]

Generating multiple molecules in the same cluster does not increase the score.
The value is bounded by 1 for any sample size. It measures coverage of held-out
cluster space at the stated cutoff.

\subsubsection{Scaffold Rediscovery (Scaff Redisc)}

Let $\mathrm{scaff}(m)$ denote the Bemis--Murcko scaffold of molecule
$m$.  Define the scaffold sets
\[
S_{H} = \{ \mathrm{scaff}(h) : h \in H \},
\qquad
S_{G} = \{ \mathrm{scaff}(g) : g \in G \}.
\]

\textbf{Scaffold Rediscovery (Scaff Redisc)}:
\[
\mathrm{Scaff\,Redisc}
= \frac{|S_H \cap S_G|}{|S_H|}.
\]

This metric gives the fraction of held-out scaffolds recovered by the generated
set. It evaluates core chemical frameworks instead of exact structures.

\subsubsection{Cross-Similarity between Scaffold Sets (CrossSim)}

\textbf{Cross-Similarity (CrossSim)} is the mean Tanimoto similarity over
all generated--held-out scaffold pairs:
\[
\mathrm{CrossSim}
= \frac{1}{|S_G|\cdot|S_H|}
  \sum_{s \in S_G}\sum_{h \in S_H} \mathrm{sim}(s,h).
\]

$S_G$ and $S_H$ are deduplicated scaffold sets, so repeated scaffolds do not add
repeated pairs. CrossSim lies in $[0,1]$ and summarizes average proximity between
the two scaffold sets. It is not a nearest-neighbor distance. Because it averages
all pairs, a high value does not show that generated molecules occupy a specific
held-out region. The released code includes a worked example and unit tests.

\subsubsection{Internal Similarity of Generated Scaffolds (InterSim)}

\textbf{Internal Similarity (InterSim)} measures the average pairwise
Tanimoto similarity among scaffolds within the generated set:
\[
\mathrm{InterSim}
= \frac{1}{|S_G|(|S_G|-1)}
  \sum_{\substack{s, s' \in S_G \\ s \neq s'}} \mathrm{sim}(s, s').
\]

Lower InterSim indicates greater scaffold diversity in the generated output.
Higher values indicate that generated molecules share similar core frameworks.

\subsubsection{Diversity Ratio (Div)}

Let $D(X)$ denote the internal diversity of a set $X$, defined as the
average pairwise Tanimoto \emph{dissimilarity} ($1 - \mathrm{sim}$)
among molecules in $X$.

\textbf{Diversity Ratio (Div)}:
\[
\mathrm{Div} = \frac{D(G)}{D(H)}.
\]

Values greater than 1 indicate that the generated set is more diverse than the
held-out set. Values below 1 indicate less diversity.

\subsubsection{Uniqueness Ratio (Uniq)}

Let $U(X) = |X_{\mathrm{canon}}| / |X|$ denote the uniqueness of a collection
$X$. It divides the number of distinct canonical SMILES by the number of samples.
This definition applies because $X$ is a multiset. For a set, it would always
equal 1.

\textbf{Uniqueness Ratio (Uniq)}:
\[
\mathrm{Uniq} = \frac{U(G)}{U(H)}.
\]

Values near 1 mean that the generated and held-out collections have similar
uniqueness. The ratio can exceed 1 because both its numerator and denominator are
fractions. We therefore do not report it as a percentage.

\section{Generated Molecules}
\label{sec:appendix-generated-molecules}
\begin{figure}[t]
  \centering
  \includegraphics[alt={A grid of selected chemical structure diagrams across five columns. Column 1 shows five task-held-out molecules exactly reproduced by the VAE and labeled by ChEMBL identifier. Columns 2 and 3 show VAE scaffold and cluster recovery examples. A reference molecule is followed by four generated molecules with Tanimoto similarity above 0.6. Columns 4 and 5 show analogous CVAE examples selected by the same nearest-neighbor display procedure.}, width=\linewidth]{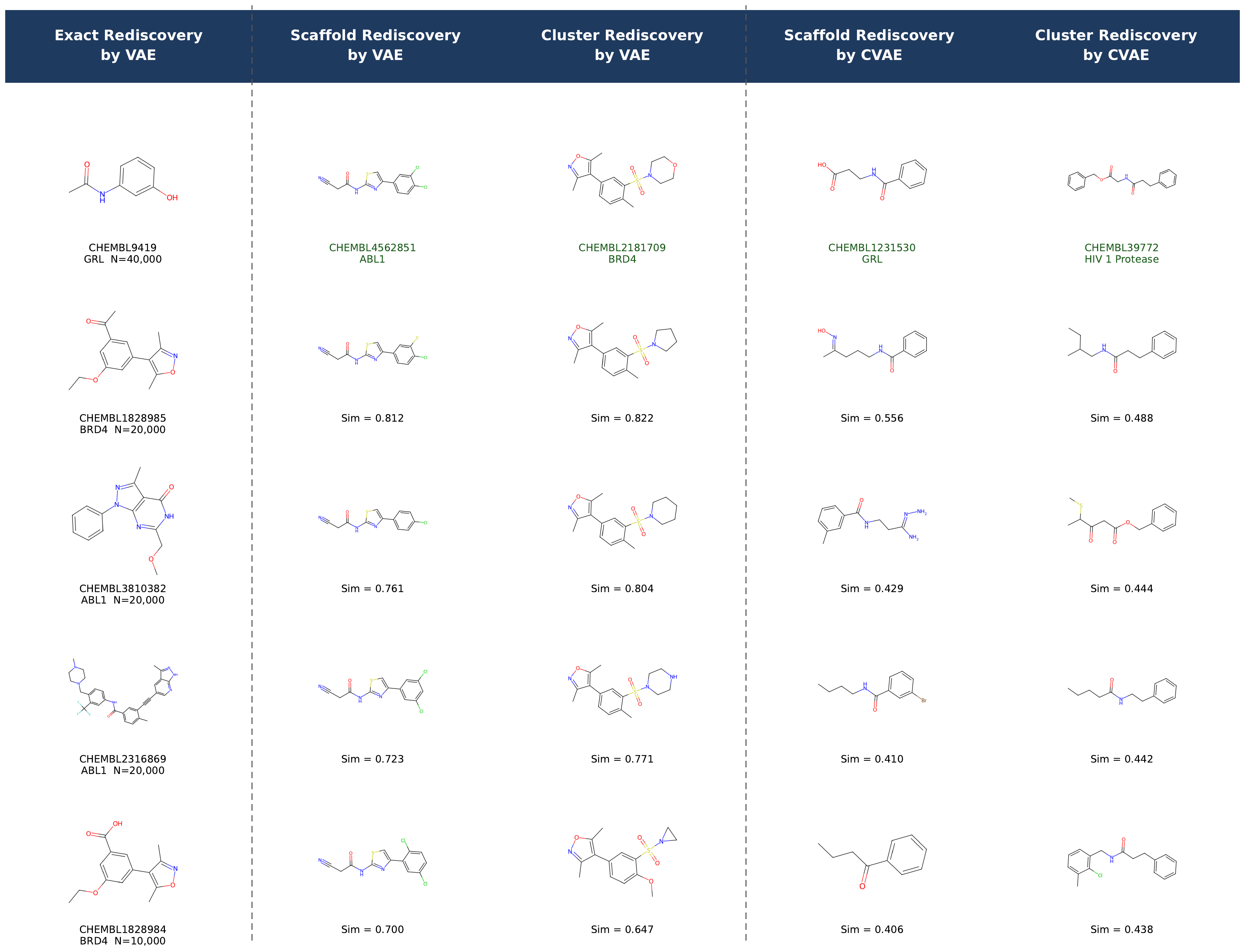}
  \caption{Samples of generated molecules from the rediscovery evaluation.
    Column~1 lists five held-out molecules exactly reproduced by the VAE (ChEMBL ID, dataset, $N$).
    Columns~2--3 show scaffold and cluster rediscovery by the VAE. Columns~4--5 present the same for the CVAE.
    In Columns~2--3 and 4--5, a reference molecule in the held-out set is shown in the first row,
    followed by its four most similar generated molecules by Tanimoto similarity.
    The VAE outputs share nearly identical scaffolds or cluster centroids (Tanimoto $> 0.6$),
    whereas the selected CVAE outputs are more varied. These examples are illustrative and do not establish a causal effect of conditioning.}
  \par\noindent\small Alt text: A grid of selected chemical structure diagrams across five columns. Column 1 shows five task-held-out molecules exactly reproduced by the VAE and labeled by ChEMBL identifier. Columns 2 and 3 show VAE scaffold and cluster recovery. A reference molecule is followed by four generated molecules with Tanimoto similarity above 0.6. Columns 4 and 5 show analogous CVAE examples selected by the same nearest-neighbor display procedure.
  \label{fig:generated-molecules}
\end{figure}

Figure~\ref{fig:generated-molecules} shows representative examples of the two
generation modes evaluated in the main manuscript. Column~1 contains five
task-held-out molecules reproduced exactly by the VAE. The overlap audit only
bounds exposure to the source corpus, so these examples do not show that the
structures were absent from pretraining. In Columns~2--3, VAE outputs remain
close to the reference scaffold, with Tanimoto similarities above 0.6. This
pattern agrees with the scaffold and cluster recovery results in Table~8 of the
main manuscript. The selected CVAE outputs in Columns~4--5 are more varied, in
line with their higher aggregate diversity ratios and lower internal similarity
across the four ChEMBL targets. These examples were selected for illustration.
They are not a random sample and do not establish that conditioning caused the
difference in diversity.
 
\section{Full MOSES Benchmark Comparison (All 15 Metrics)}
\label{sec:appendix-moses-full}

Table~\ref{tab:model_comparison_full} reports all 15 MOSES metrics for every
model evaluated in the main text. We obtained values for the eight classic
baselines (HMM, NGram, Combinatorial, CharRNN, AAE, VAE, JTN-VAE, and LatentGAN)
from the official molecularsets repository~\cite{polykovskiy2020molecular}. We
obtained values for the five modern baselines (SAFE-GPT, DiGress, FragFM, MolHIT,
and GVT) from their publications. These papers report only Validity, Unique@10k,
FCD/Test, Filters, and Novelty. We mark the other ten columns with \textemdash.
For \ModelName, values are means $\pm$ standard deviations over three evaluation
seeds.

\begin{landscape}
\begin{table}[H]
\caption{Full MOSES benchmark comparison across all 15 metrics. Baseline values (HMM, NGram, Combinatorial, CharRNN~\cite{segler2018generating}, AAE~\cite{DBLP:journals/corr/MakhzaniSJG15}, VAE~\cite{DBLP:journals/corr/KingmaW13}, JTN-VAE~\cite{jin2018junction}, LatentGAN~\cite{priyadarsini2024self}) are retrieved directly from the official molecularsets (MOSES) repository~\cite{polykovskiy2020molecular}. Our \ModelName\ runs are reported as means $\pm$ standard deviations computed over three independent evaluation seeds. Modern baseline rows (SAFE-GPT, DiGress, FragFM, MolHIT, GVT) are literature values ($^\dagger$) taken from the cited publications. Columns not reported in the original papers are indicated by \textemdash.}
\label{tab:model_comparison_full}
\centering
\scriptsize
\setlength{\tabcolsep}{3pt}
\begin{adjustbox}{max width=\linewidth}
\begin{tabular}{lccccccccccccccc}
\toprule
\textbf{Models} & \textbf{Valid (↑)} & \textbf{Unique@1k (↑)} & \textbf{Unique@10k (↑)} & \textbf{FCD/Test (↓)} & \textbf{FCD/TestSF (↓)} & \textbf{SNN/Test (↑)} & \textbf{SNN/TestSF (↑)} & \textbf{Frag/Test (↑)} & \textbf{Frag/TestSF (↑)} & \textbf{Scaf/Test (↑)} & \textbf{Scaf/TestSF (↑)} & \textbf{IntDiv1 (↑)} & \textbf{IntDiv2 (↑)} & \textbf{Filters (↑)} & \textbf{Novelty (↑)} \\
\midrule
\textit{Reference Split} & & & & & & & & & & & & & & \\
\textit{Train} & \textit{1.0} & \textit{1.0} & \textit{1.0} & \textit{0.008} & \textit{0.4755} & \textit{0.6419} & \textit{0.5859} & \textit{1.0} & \textit{0.9986} & \textit{0.9907} & \textit{0.0} & \textit{0.8567} & \textit{0.8508} & \textit{1.0} & \textit{1.0} \\
\midrule
\textit{Classic baselines} & & & & & & & & & & & & & & \\
HMM & 0.076$\pm$0.0322 & 0.623$\pm$0.1224 & 0.5671$\pm$0.1424 & 24.4661$\pm$2.5251 & 25.4312$\pm$2.5599 & 0.3876$\pm$0.0107 & 0.3795$\pm$0.0107 & 0.5754$\pm$0.1224 & 0.5681$\pm$0.1218 & 0.2065$\pm$0.0481 & 0.049$\pm$0.018 & 0.8466$\pm$0.0403 & 0.8104$\pm$0.0507 & 0.9024$\pm$0.0489 & 0.9994$\pm$0.001 \\
NGram & 0.2376$\pm$0.0025 & 0.974$\pm$0.0108 & 0.9217$\pm$0.0019 & 5.5069$\pm$0.1027 & 6.2306$\pm$0.0966 & 0.5209$\pm$0.001 & 0.4997$\pm$0.0005 & 0.9846$\pm$0.0012 & 0.9815$\pm$0.0012 & 0.5302$\pm$0.0163 & 0.0977$\pm$0.0142 & 0.8738$\pm$0.0002 & 0.8644$\pm$0.0002 & 0.9582$\pm$0.001 & 0.9694$\pm$0.001 \\
Combinatorial & 1.0$\pm$0.0 & 0.9983$\pm$0.0015 & 0.9909$\pm$0.0009 & 4.2375$\pm$0.037 & 4.5113$\pm$0.0274 & 0.4514$\pm$0.0003 & 0.4388$\pm$0.0002 & 0.9912$\pm$0.0004 & 0.9904$\pm$0.0003 & 0.4445$\pm$0.0056 & 0.0865$\pm$0.0027 & 0.8732$\pm$0.0002 & 0.8666$\pm$0.0002 & 0.9557$\pm$0.0018 & 0.9878$\pm$0.0008 \\
CharRNN & 0.9748$\pm$0.0264 & 1.0$\pm$0.0 & 0.9994$\pm$0.0003 & \textbf{0.0732}$\pm$0.0247 & \textbf{0.5204}$\pm$0.0379 & 0.6015$\pm$0.0206 & 0.5649$\pm$0.0142 & \textbf{0.9998}$\pm$0.0002 & \textbf{0.9983}$\pm$0.0003 & 0.9242$\pm$0.0058 & \textbf{0.1101}$\pm$0.0081 & 0.8562$\pm$0.0005 & 0.8503$\pm$0.0005 & \textbf{0.9943}$\pm$0.0034 & 0.8419$\pm$0.0509 \\
AAE & 0.9368$\pm$0.0341 & 1.0$\pm$0.0 & 0.9973$\pm$0.002 & 0.5555$\pm$0.2033 & 1.0572$\pm$0.2375 & 0.6081$\pm$0.0043 & 0.5677$\pm$0.0045 & 0.991$\pm$0.0051 & 0.9905$\pm$0.0039 & 0.9022$\pm$0.0375 & 0.0789$\pm$0.009 & 0.8557$\pm$0.0031 & 0.8499$\pm$0.003 & 0.996$\pm$0.0006 & 0.7931$\pm$0.0285 \\
VAE & 0.9767$\pm$0.0012 & 1.0$\pm$0.0 & 0.9984$\pm$0.0005 & 0.099$\pm$0.0125 & 0.567$\pm$0.0338 & 0.6257$\pm$0.0005 & 0.5783$\pm$0.0008 & 0.9994$\pm$0.0001 & 0.9984$\pm$0.0003 & 0.9386$\pm$0.0021 & 0.0588$\pm$0.0095 & 0.8558$\pm$0.0004 & 0.8498$\pm$0.0004 & 0.997$\pm$0.0002 & 0.6949$\pm$0.0069 \\
JTN-VAE & 1.0$\pm$0.0 & 1.0$\pm$0.0 & 0.9996$\pm$0.0003 & 0.3954$\pm$0.0234 & 0.9382$\pm$0.0531 & 0.5477$\pm$0.0076 & 0.5194$\pm$0.007 & 0.9965$\pm$0.0003 & 0.9947$\pm$0.0002 & 0.8964$\pm$0.0039 & 0.1009$\pm$0.0105 & 0.8551$\pm$0.0034 & 0.8493$\pm$0.0035 & 0.976$\pm$0.0016 & 0.9143$\pm$0.0058 \\
LatentGAN & 0.8966$\pm$0.0029 & 1.0$\pm$0.0 & 0.9968$\pm$0.0002 & 0.2968$\pm$0.0087 & 0.8281$\pm$0.0117 & 0.5371$\pm$0.0004 & 0.5132$\pm$0.0002 & 0.9986$\pm$0.0004 & 0.9972$\pm$0.0007 & 0.8867$\pm$0.0009 & 0.1072$\pm$0.0098 & 0.8565$\pm$0.0007 & 0.8505$\pm$0.0006 & 0.9735$\pm$0.0006 & 0.9498$\pm$0.0006 \\
\midrule
\textit{Modern baselines ($^\dagger$literature values)} & & & & & & & & & & & & & & \\
SAFE-GPT$^\dagger$~\cite{noutahi2024safe} & 0.998 & \textemdash & 0.989 & 0.72 & \textemdash & 0.57 & \textemdash & \textemdash & \textemdash & 0.063 & \textemdash & \textemdash & \textemdash & 0.977 & 0.437 \\
DiGress$^\dagger$~\cite{vignac2023digress} & 0.857 & \textemdash & 1.000 & 1.19 & \textemdash & 0.52 & \textemdash & \textemdash & \textemdash & 0.148 & \textemdash & \textemdash & \textemdash & 0.971 & 0.950 \\
FragFM$^\dagger$~\cite{kim2025fragfm} & 0.998 & \textemdash & 1.000 & 0.58 & \textemdash & 0.56 & \textemdash & \textemdash & \textemdash & 0.109 & \textemdash & \textemdash & \textemdash & 0.991 & 0.871 \\
MolHIT$^\dagger$~\cite{jung2026molhit} & 0.991 & \textemdash & 0.998 & 1.03 & \textemdash & 0.55 & \textemdash & \textemdash & \textemdash & 0.144 & \textemdash & \textemdash & \textemdash & 0.980 & 0.914 \\
GVT$^\dagger$~\cite{zheng2025gvt} & 0.994 & \textemdash & 1.000 & 0.16 & \textemdash & \textbf{0.64} & \textemdash & \textemdash & \textemdash & 0.034 & \textemdash & \textemdash & \textemdash & 0.996 & 0.614 \\
\midrule
\textit{Ours (\ModelName)} & & & & & & & & & & & & & & \\
\ModelName~PubChem 80M (Random) & \textbf{1.0000}$\pm$0.0000 & \textbf{1.0000}$\pm$0.0000 & \textbf{1.0000}$\pm$0.0000 & 14.1140$\pm$0.0607 & 15.0960$\pm$0.0637 & 0.3140$\pm$0.0002 & 0.3059$\pm$0.0003 & 0.9116$\pm$0.0012 & 0.9058$\pm$0.0013 & 0.2831$\pm$0.0095 & 0.1303$\pm$0.0041 & \textbf{0.8990}$\pm$0.0001 & \textbf{0.8939}$\pm$0.0001 & 0.4766$\pm$0.0014 & \textbf{0.9999}$\pm$0.0000 \\
\ModelName~PubChem 80M (Seed) & \textbf{1.0000}$\pm$0.0000 & \textbf{1.0000}$\pm$0.0000 & 0.9998$\pm$0.0000 & 2.0062$\pm$0.0084 & 2.7635$\pm$0.0122 & 0.5090$\pm$0.0007 & 0.4798$\pm$0.0005 & 0.9964$\pm$0.0001 & 0.9931$\pm$0.0001 & \textbf{0.8987}$\pm$0.0029 & 0.0708$\pm$0.0055 & 0.8683$\pm$0.0002 & 0.8626$\pm$0.0002 & 0.8313$\pm$0.0013 & 0.7583$\pm$0.0010 \\
\ModelName~MOSES 1.5M (Random) & \textbf{1.0000}$\pm$0.0000 & \textbf{1.0000}$\pm$0.0000 & 0.9999$\pm$0.0001 & 9.1489$\pm$0.0291 & 10.0725$\pm$0.0265 & 0.3842$\pm$0.0003 & 0.3722$\pm$0.0004 & 0.9676$\pm$0.0011 & 0.9598$\pm$0.0012 & 0.5622$\pm$0.0120 & 0.1123$\pm$0.0112 & 0.8809$\pm$0.0001 & 0.8755$\pm$0.0001 & 0.6398$\pm$0.0029 & 0.9939$\pm$0.0002 \\
\ModelName~MOSES 1.5M (Seed) & \textbf{1.0000}$\pm$0.0000 & 0.9997$\pm$0.0005 & 0.9996$\pm$0.0002 & 6.6121$\pm$0.0335 & 7.5617$\pm$0.0315 & 0.4263$\pm$0.0003 & 0.4080$\pm$0.0003 & 0.9795$\pm$0.0001 & 0.9731$\pm$0.0002 & 0.7583$\pm$0.0057 & 0.0880$\pm$0.0124 & 0.8806$\pm$0.0001 & 0.8745$\pm$0.0001 & 0.6931$\pm$0.0018 & 0.9567$\pm$0.0004 \\
\bottomrule
\end{tabular}
\end{adjustbox}
\end{table}
\end{landscape}

\section{Per-Bucket Breakdown: SA score Conditional Generation}
\label{sec:appendix-sa-bucket}

Table~\ref{tab:sa-bucket-breakdown} separates the aggregate SA-score benchmark
results into three target regions: low $y^{*}$ (easy synthesis,
$y^{*}\!\approx\!0.167$), middle $y^{*}$ ($y^{*}\!\approx\!0.500$), and high
$y^{*}$ (hard synthesis, $y^{*}\!\approx\!0.833$). All rows use the same RDKit
SA scorer and min--max scaler as the main experiment.

\paragraph{Denominator convention.}
We assign each attempt to a bucket from its \emph{requested} label $y^{*}$ before
validity filtering. Every attempt therefore belongs to one bucket, including
attempts that do not produce a parseable molecule. \ModelName\ and MolGPT sample
three molecules at each of 1,001 $y^{*}$ values. This gives
$n_{\text{att}} = 999/1002/1002$ in the low, middle, and high regions, for 3,003
attempts in total. REINVENT~4 ran three campaigns of 1,001 requests
(\texttt{num\_smiles}~$=1001$, \texttt{unique\_molecules}~$=$~true) and retained
640, 996, and 992 unique valid molecules. Its sampler removes duplicate and
unparseable outputs. The low-$y^{*}$ campaign recorded no sanitization failures,
so nearly all 361 removed outputs were duplicates. We therefore describe
$n_{\text{val}}$ for REINVENT~4 as the number of \emph{unique valid} molecules
retained from 1,001 requests and label the ratio ``Retain.'' DrugDiff generates
1,001 molecules per bucket and is valid by construction. We report two success
rates because output retention differs across models. SR$^{\text{val}}$ divides
by $n_{\text{val}}$. SR$^{\text{att}}$ divides by $n_{\text{att}}$ and counts
invalid or unscored attempts as failures.
The table reports both $n_{\text{att}}$ and $n_{\text{val}}$ so that validity
and retention are calculated from the appropriate denominators.

\paragraph{Bucket-level observations.}
\textbf{MolGPT.} Validity decreases most strongly in the high-$y^{*}$ region.
Only 244 of 1,002 attempts ($24.4\%$) produce valid SMILES, compared with 756 of
999 attempts ($75.7\%$) in the low-$y^{*}$ region.

\textbf{REINVENT~4.} The low-$y^{*}$ campaign produces 61 unique Murcko
scaffolds from 640 retained molecules (Scaf/mol~$=0.095$). The high-$y^{*}$
campaign produces 979 scaffolds from 992 molecules (Scaf/mol~$=0.987$). The
reward used in each campaign may contribute to this difference. \rev{The
low-$y^{*}$ campaign has the lowest retention outside MolGPT
($640/1001 = 63.9\%$). Because REINVENT~4 removes duplicates, this result reflects
repeated generation as well as chemical validity. Its
SR$^{\text{att}}_{20}=0.639$ measures success per requested molecule.}

\textbf{\ModelName.} \rev{Validity remains between $97\%$ and $100\%$ across
the three regions ($995/999$, $985/1002$, and $976/1002$).} The model produces
1,974 distinct scaffolds, the highest total in this comparison. Scaf/mol declines
from 0.774 at low $y^{*}$ to 0.578 at high $y^{*}$. Hard-synthesis structures are
less common in the training corpus, which may contribute to this trend.

\textbf{DrugDiff.} All generated molecules are valid, but the mean SA scores do
not move toward the requested values. The observed means are 0.521, 0.541, and
0.513 for targets of 0.167, 0.500, and 0.833. Under this test, the predictor
trained on ZINC-250K does not transfer to the PubChem distribution.

\begin{table}[H]
\centering
\caption{Per-bucket breakdown of SA score conditional generation metrics for all four models. Each row reports results for one model--bucket combination. \textbf{Bucket}: low $y^{*}$ = easy synthesis ($y^{*}\!\approx\!0.167$, SA$_{\text{raw}}\!\approx\!2.45$); mid $y^{*}$ ($y^{*}\!\approx\!0.500$, SA$_{\text{raw}}\!\approx\!4.48$); high $y^{*}$ = hard synthesis ($y^{*}\!\approx\!0.833$, SA$_{\text{raw}}\!\approx\!6.49$). \rev{$n_{\text{att}}$: generation attempts assigned to the bucket by requested label, \emph{before} validity filtering. $n_{\text{val}}$: valid SMILES retained. Ret.: $n_{\text{val}}/n_{\text{att}}$; this is RDKit validity for \ModelName, MolGPT, and DrugDiff, but for REINVENT~4 it is unique-valid retention, because that sampler also discards duplicates (see above). $\hat{y}$: mean realized normalized SA score $\pm\,\sigma$ over valid molecules. MAE: mean $|y^{*}-\hat{y}|$ over valid molecules. SR$^{\text{val}}_{10}$/SR$^{\text{val}}_{20}$: fraction of \emph{valid} molecules within $\varepsilon\!=\!0.10/0.20$ of $y^{*}$. SR$^{\text{att}}_{10}$/SR$^{\text{att}}_{20}$: the same numerator divided by $n_{\text{att}}$, so invalid attempts count as failures. The two coincide only at 100\% validity.} Scaf: unique Bemis--Murcko scaffolds; Scaf/mol $= \text{Scaf}/n_{\text{val}}$. Bold: best value per metric within each paradigm group (Paradigm A: oracle-free continuous conditioning; Paradigm B: oracle-in-the-loop / 3-bucket conditioning). $^{\ddagger}$REINVENT~4 and DrugDiff use a single fixed $y^{*}$ per bucket.}
\label{tab:sa-bucket-breakdown}
\scriptsize
\setlength{\tabcolsep}{2.5pt}
\begin{adjustbox}{max width=\linewidth}
\begin{tabular}{llrrrrrrrrrrrr}
\toprule
\textbf{Model} & \textbf{Bucket} & \textbf{$y^{*}$} & \textbf{$n_{\text{att}}$} & \textbf{$n_{\text{val}}$} & \textbf{Ret.} & \textbf{$\hat{y}$} & \textbf{MAE} & \textbf{SR$^{\text{val}}_{10}$} & \textbf{SR$^{\text{att}}_{10}$} & \textbf{SR$^{\text{val}}_{20}$} & \textbf{SR$^{\text{att}}_{20}$} & \textbf{Scaf} & \textbf{Scaf/mol} \\
\midrule
\multicolumn{14}{l}{\textit{Paradigm A: oracle-free continuous conditioning (1,001 levels $\times$ 3 samples; buckets are $y^{*}$ bins)}} \\
\textbf{\ModelName}\ (ours, $\alpha\!=\!12$) & low $y^{*}$  & 0.167 & 999  & 995  & 0.996 & 0.531 & 0.366 & 0.044 & 0.044 & 0.143 & 0.142 & \textbf{770} & \textbf{0.774} \\
\textbf{\ModelName}\ (ours, $\alpha\!=\!12$) & mid $y^{*}$  & 0.500 & 1002 & 985  & 0.983 & 0.739 & 0.265 & 0.179 & 0.176 & 0.374 & 0.367 & \textbf{691} & 0.702 \\
\textbf{\ModelName}\ (ours, $\alpha\!=\!12$) & high $y^{*}$ & 0.833 & 1002 & 976  & \textbf{0.974} & \textbf{0.845} & \textbf{0.148} & \textbf{0.424} & \textbf{0.413} & \textbf{0.722} & \textbf{0.704} & \textbf{564} & 0.578 \\
\cmidrule(lr){1-14}
MolGPT                                      & low $y^{*}$  & 0.167 & 999  & 756  & 0.757 & \textbf{0.189} & \textbf{0.056} & \textbf{0.843} & \textbf{0.638} & \textbf{0.978} & \textbf{0.740} & 561  & 0.742 \\
MolGPT                                      & mid $y^{*}$  & 0.500 & 1002 & 515  & 0.514 & \textbf{0.469} & \textbf{0.090} & \textbf{0.656} & 0.337 & \textbf{0.915} & 0.470 & 386  & \textbf{0.750} \\
MolGPT                                      & high $y^{*}$ & 0.833 & 1002 & 244  & 0.244 & 0.616 & 0.204 & 0.258 & 0.063 & 0.566 & 0.138 & 183  & \textbf{0.750} \\
\midrule
\multicolumn{14}{l}{\textit{Paradigm B: oracle-in-the-loop / 3-bucket conditioning (fixed $y^{*}$ per campaign$^{\ddagger}$)}} \\
REINVENT~4$^{\ddagger}$                     & low $y^{*}$  & 0.167 & 1001 & 640  & 0.639 & \textbf{0.003} & 0.164 & 0.000 & 0.000 & \textbf{1.000} & 0.639 & 61   & 0.095 \\
REINVENT~4$^{\ddagger}$                     & mid $y^{*}$  & 0.500 & 1001 & 996  & \textbf{0.995} & \textbf{0.395} & \textbf{0.132} & \textbf{0.473} & \textbf{0.471} & 0.764 & 0.760 & 312  & 0.313 \\
REINVENT~4$^{\ddagger}$                     & high $y^{*}$ & 0.833 & 1001 & 992  & \textbf{0.991} & \textbf{0.974} & 0.144 & 0.129 & 0.128 & \textbf{0.998} & \textbf{0.989} & \textbf{979} & \textbf{0.987} \\
\cmidrule(lr){1-14}
DrugDiff$^{\ddagger}$                       & low $y^{*}$  & 0.167 & 1001 & \textbf{1001} & \textbf{1.000} & 0.521 & 0.356 & 0.053 & 0.053 & 0.191 & 0.191 & 509  & \textbf{0.508} \\
DrugDiff$^{\ddagger}$                       & mid $y^{*}$  & 0.500 & 1001 & \textbf{1001} & \textbf{1.000} & 0.541 & 0.136 & 0.465 & 0.465 & 0.756 & 0.756 & \textbf{551} & 0.550 \\
DrugDiff$^{\ddagger}$                       & high $y^{*}$ & 0.833 & 1001 & \textbf{1001} & \textbf{1.000} & 0.513 & \textbf{0.327} & 0.103 & 0.103 & 0.223 & 0.223 & 440  & 0.440 \\
\bottomrule
\end{tabular}
\end{adjustbox}
\end{table}

\section{Reproducibility Details}
\label{sec:appendix-reproducibility}

This appendix lists the training, generation, and evaluation settings recovered
from the released code, configuration files, checkpoints, and run logs. We mark
settings that could not be recovered instead of estimating them.

\subsection{Pretraining}

Training used Adam, a constant learning rate of $10^{-5}$, and a batch size of
256. No learning-rate schedule was used. The KL coefficient increased
linearly from 0 to $\beta = 1.0$ during the first 10{,}000 optimizer steps and
remained at 1.0 thereafter. The checkpoint records no changes to these defaults.
We exclude fixed sinusoidal positional encodings from parameter counts. The
pretrained checkpoint contains 89{,}175{,}368 parameters.

Tokenization allows at most 71 SELFIES tokens, or 72 positions including the
beginning-of-sequence token. Longer sequences are \emph{truncated}. Truncation
can remove the end-of-sequence token and cause the decoded molecule to differ
from the input. We did not measure how many training molecules were affected.
Future preprocessing should exclude overlength sequences instead.

\subsection{Conditional adaptation}

We adapted each conditional model in two stages. The reported SA-score and
docking results use the second-stage checkpoints.

\begin{enumerate}
  \item \textbf{LoRA adaptation.} We train the LoRA adapters, FiLM network, and
  latent-to-hidden projection. We freeze all other parameters, including the
  conditional-prior networks and posterior label projection. The fused attention
  kernel bypasses adapters at the attention output projections, so those adapters
  receive no gradient. This stage therefore trains the feed-forward adapters,
  FiLM, and latent-to-hidden projection. About 1.64\,M of the 1.94\,M parameters
  reported as trainable receive updates.
  \item \textbf{Conditioning-pathway refinement.} We freeze all Transformer
  weights and LoRA adapters. We update the FiLM network, the conditional-prior
  mean and log-variance networks, the latent-to-hidden projection, and the label
  projection. This gives 726{,}016 trainable parameters out of 91{,}047{,}752.
  The stage uses a fixed KL weight of 1.0 and a per-dimension free-bits floor of
  0.3 from its first step.
\end{enumerate}

Per-experiment trainable and total parameter counts, conditioning pathways, and
prior status are given in the main text.

\subsection{Generation and decoding}

\begin{table}[H]
\centering
\caption{Generation settings by experiment. ``Greedy'' indicates that the next
token is the argmax of the (possibly temperature-scaled) logits, so temperature
has no effect on the selected token.}
\label{tab:repro-generation}
\smallskip
\footnotesize
\begin{tabular}{llll}
\toprule
\textbf{Experiment} & \textbf{Latent source} & \textbf{Token selection} & \textbf{Seeds} \\
\midrule
MOSES random & $\mathcal{N}(\mathbf{0},\mathbf{I})$ & multinomial, $T{=}1.1$, top-$k$ 20 & 3 evaluation seeds \\
MOSES seed-based & $\mu + \varepsilon\sigma$, scale 1.0 & multinomial, $T{=}1.1$, top-$k$ 20 & 3 evaluation seeds \\
SA / docking sweeps & $p_\theta(\mathbf{z}|y^{*})$ & greedy & 42 \\
Ablations & $\mathcal{N}(\mathbf{0},\mathbf{I})$ & greedy & 42, 0, 101 \\
Rediscovery (VAE) & posterior-local, scale 1.0 & multinomial, $T{=}1.1$ & not recorded \\
Rediscovery (CVAE) & $\mathcal{N}(\mathbf{0},\mathbf{I})$ + FiLM & greedy & not recorded \\
\bottomrule
\end{tabular}
\end{table}

The MOSES seed protocol encodes 30{,}000 molecules from the official MOSES
training split and draws one latent per seed as
$\mathbf{z} = \mu + \varepsilon\,\sigma_{\text{posterior}}$ with $\varepsilon \sim
\mathcal{N}(\mathbf{0},\mathbf{I})$ with a multiplier of 1.0. This samples at the
full learned posterior scale, not as a small perturbation around $\mu$. The
released artifacts do not record the exact perturbation scale used for the
rediscovery runs. The code default is 1.0.

The primary ablation and refinement runs used random seed 42. Six SA-score
ablation configurations were also trained with seeds 7 and 13, giving three
independent fine-tuning runs per configuration. Deterministic cuDNN kernels were
not enabled, so a fixed seed does not guarantee bitwise-identical results.

\paragraph{Docking evaluation populations.}
Each guidance condition starts with 3{,}003 generation attempts. We count an
output as generation-valid when it decodes from SELFIES and RDKit can parse it.
We count docking as complete when QVina returns a value other than the sentinel
10{,}000. We retain real non-negative QVina scores as unfavorable outcomes.
Table~4 calculates correlation, calibration slope, valid- or docked-only SR, and
IntDiv over the completed-docking population. It separately reports generation
validity and docking completion. Table~5 reports attempt-level SR and counts every
invalid or failed-docking attempt as a failure. The analysis uses fixed
normalization ranges: $s_{\min}=-10.0$ for 1SYH and 4LDE, and $-9.4$ for 6Y2F.
It does not estimate a new range from each generated sample.

\paragraph{Rediscovery table provenance.}
The 60-row rediscovery table was generated from the combined results using
explicit metric names. This keeps CrossSim, InterSim, Div, and Uniq aligned with
their definitions.

\subsection{Statistical reporting}

MOSES results use three evaluation seeds. They repeat sampling from one fixed
checkpoint and are not independent training runs.

For the replicated SA-score ablations, we report the mean and standard deviation
across three independently fine-tuned models. The remaining results use one
trained model per condition. GuacaMol, SA-score guidance, docking, and
rediscovery therefore report point estimates, with bootstrap or generation-seed
uncertainty where stated. Those intervals do not include variation from model
fitting, data splitting, pseudo-label estimation, or checkpoint selection.
 
\section{Pretraining Overlap Audit}
\label{sec:appendix-overlap}

The backbone was pretrained on 79 million PubChem molecules. The downstream
reference sets come from public sources that may overlap with PubChem. We
therefore measured overlap instead of assuming that the benchmark and held-out
structures were absent from the pretraining source corpus. This appendix reports
the results for MOSES, GuacaMol, and the cluster-held-out ChEMBL sets.

\subsection{Procedure}

We parsed every molecule in each reference set with RDKit and converted it to a
standard InChIKey. We applied the same procedure to all 79{,}311{,}161 molecules
in the curated PubChem corpus and intersected the resulting key sets. We matched
molecules by the full InChIKey and by its first block. The full key includes
stereochemistry and isotopic labeling. The first block represents molecular
connectivity only. Reporting both separates exact structural overlap from
connectivity matches with different stereochemistry or isotopes. The main text
uses the full-InChIKey results. The two measures are identical for all four
held-out ChEMBL sets.

The ChEMBL reference sets are the cluster-held-out splits used for the
rediscovery evaluation, so the audited molecules are exactly those the
rediscovery metrics are computed against.

Two qualifications apply to Table~\ref{tab:pretraining-overlap}. First, the
audited corpus contains more molecules than the training dataset.
The reported percentages are therefore upper bounds on exposure. Second,
overlap describes the data, not the model's behavior. It limits how benchmark
and rediscovery results can be interpreted, but it does not show that the model
memorized a particular generated molecule.

\begin{table}[H]
\centering
\caption{\revfloat{\rev{Overlap between the PubChem pretraining corpus and the benchmark and evaluation sets. Each reference molecule was canonicalized to a standard InChIKey. ``Unique keys'' is the number of distinct full InChIKeys after deduplication. These keys were matched against all 79{,}311{,}161 curated PubChem molecules. A total of 252 rows ($3\times10^{-4}\%$) failed to parse and were excluded. ``Exact'' requires full InChIKey identity, including stereochemistry. ``Connectivity'' matches only the first InChIKey block, ignoring stereochemistry and isotopes, and is therefore an upper bound on structural exposure. The two measures agree exactly for all four held-out ChEMBL sets, which carry the leakage argument, and differ by at most 0.15 percentage points elsewhere (GRL fine-tuning), so the overlap reflects structural identity rather than stereochemical variants. Percentages quoted in the main text are the exact-InChIKey values, which are the more conservative of the two. Because the audited corpus is larger than the training dataset, these values bound exposure from above.}}}
\label{tab:pretraining-overlap}
\smallskip
\footnotesize
\begin{tabular}{lrrrr}
\toprule
\textbf{Reference set} & \textbf{Unique keys} & \textbf{Exact matches} & \textbf{Exact} & \textbf{Connectivity} \\
\midrule
MOSES train & 1,584,399 & 1,132,095 & 71.45\% & 71.45\% \\
MOSES test & 176,072 & 125,331 & 71.18\% & 71.19\% \\
MOSES test-scaffolds & 176,222 & 126,042 & 71.52\% & 71.53\% \\
GuacaMol train & 1,272,168 & 1,107,308 & 87.04\% & 87.10\% \\
\midrule
ABL1 held-out & 359 & 296 & 82.45\% & 82.45\% \\
ABL1 fine-tuning & 944 & 809 & 85.70\% & 85.70\% \\
BRD4 held-out & 1,944 & 1,477 & 75.98\% & 75.98\% \\
BRD4 fine-tuning & 4,475 & 3,572 & 79.82\% & 79.82\% \\
GRL held-out & 600 & 398 & 66.33\% & 66.33\% \\
GRL fine-tuning & 1,306 & 1,048 & 80.25\% & 80.40\% \\
HIV-1 protease held-out & 459 & 100 & 21.79\% & 21.79\% \\
HIV-1 protease fine-tuning & 1,137 & 317 & 27.88\% & 27.97\% \\
\bottomrule
\end{tabular}
\end{table}
 
\subsection{Consequences for the benchmark results}

Between 71\% and 72\% of MOSES molecules and 87\% of GuacaMol training molecules
match the curated PubChem source corpus. These values are upper bounds on
pretraining exposure. The MOSES and GuacaMol results therefore do not measure
transfer to wholly unfamiliar chemical distributions. The model received no
parameter updates on these benchmark splits and no benchmark-specific tuning,
but pretraining substantially covered their chemical space. We use this narrower
interpretation throughout the manuscript.

The overlap does not affect the within-model comparison of random-prior and
seed-based sampling because both protocols use the same weights and pretraining
corpus. The improvement from seed-based sampling reflects conditioning on a
supplied structure. It does not demonstrate generalization to unfamiliar
chemistry.

\subsection{Consequences for the rediscovery results}

The cluster-held-out split removes close analogues of the evaluation compounds
from target-specific fine-tuning. It does not remove those compounds from the
PubChem source corpus. Table~\ref{tab:pretraining-overlap} shows source-corpus
matches for 21.79\%--82.45\% of held-out compounds, depending on the protein
target. These values are upper bounds on pretraining exposure.

This result directly affects the interpretation of exact rediscovery. A recovered
held-out structure that matches the source corpus may come from the pretrained
distribution. It is not evidence of a previously unseen molecule. Exposure and
exact recall follow the same general ordering across the four targets:

\begin{table}[H]
\centering
\caption{\revfloat{\rev{Held-out pretraining exposure and the highest exact rediscovery recall observed for the pretrained VAE under seed-based sampling across sample sizes from 10K to 50K. The ordering across four protein targets is descriptive. It informs the interpretation of exact rediscovery but does not estimate a memorization effect.}}}
\label{tab:overlap-vs-rediscovery}
\smallskip
\footnotesize
\begin{tabular}{lrr}
\toprule
\textbf{Protein target} & \textbf{Held-out exposure} & \textbf{Max exact recall} \\
\midrule
ABL1 & 82.45\% & 2.198\% \\
BRD4 & 75.98\% & 0.296\% \\
GRL & 66.33\% & 0.650\% \\
HIV-1 protease & 21.79\% & 0.000\% \\
\bottomrule
\end{tabular}
\end{table}

HIV-1 protease has the lowest exposure and no exact rediscoveries at any sample
size. ABL1 has the highest exposure and the highest recall. Spearman correlation
across the four targets is 0.80. With only four points, this is a descriptive
observation and not an estimate of the relationship.

The PubChem molecular-weight filter explains much of the lower exposure for
HIV-1 protease. The corpus excludes molecules above 600\,Da, so none of those
molecules can match. Across all four targets, exposure is 0.00\% above 600\,Da
and 33.4--87.1\% at or below 600\,Da. The HIV-1 protease held-out set has a
median molecular weight of 566\,Da, and 35.0\% of its compounds exceed 600\,Da.
The other targets have median molecular weights of 425--473\,Da, with
4.7--6.3\% of compounds above 600\,Da. The curation filter and molecular-weight
distribution therefore account for much of the lower HIV-1 protease overlap.
Exposure and scaffold class are also confounded across these four targets, so
the reported correlation does not estimate a memorization effect.

Scaffold and cluster recovery are less sensitive to exact pretraining overlap.
They record whether a generated molecule reaches a Bemis--Murcko framework or a
cluster, not whether it reproduces one structure. We therefore use scaffold and
cluster recovery as the main rediscovery evidence and treat exact rediscovery as
a bounded observation.
  
\end{document}